\documentclass{article}

\usepackage{graphicx} 
\usepackage{setspace} 
\usepackage{natbib} 
\usepackage{amsmath} 
\usepackage{float} 
\usepackage{placeins} 
\usepackage{rotating} 
\usepackage{pst-all} 
\usepackage{pdftricks} 
\usepackage{lscape} 
\usepackage{bbm} 
\usepackage{comment} 
\usepackage{authblk} 
\usepackage{longtable} 
\usepackage{caption} 
\usepackage{tikz} 
\usepackage{enumitem} 
\usepackage{subcaption} 
\usepackage{booktabs} 
\usepackage{geometry} 
\usepackage{hyperref} 
\usepackage{lineno} 
\usepackage{xcolor} 
\usepackage{overcite} 
\usepackage{amsmath}
\usepackage{amssymb}
\usepackage[retainorgcmds]{IEEEtrantools}
\usepackage{titling}

\hypersetup{
    unicode=false,           
    pdftoolbar=true,         
    pdfmenubar=true,         
    pdffitwindow=true,       
    pdfstartview={FitH},     
    pdftitle={},             
    pdfauthor={},            
    pdfsubject={},           
    pdfcreator={},           
    pdfkeywords={},          
    pdfnewwindow=true,       
    colorlinks=true,         
    linkcolor=black,         
    citecolor=black,         
    filecolor=black,         
    urlcolor=blue            
}

\date{\today}

\begin{document}

\author{
    Sourabh Balgi\textsuperscript{1}, 
    Adel Daoud\textsuperscript{2}, 
    Jose M. Pe{\~n}a\textsuperscript{1}, 
    Geoffrey T. Wodtke\textsuperscript{3}\thanks{Direct correspondence to: Geoffrey T. Wodtke, University of Chicago, Department of Sociology, 1126 E. 59th St., Chicago, IL, 60637; wodtke@uchicago.edu. The authors thank Steve Raudenbush, Xiang Zhou, Bernie Koch, Kaz Yamaguchi, and participants in the conference on ``New Methods to Measure Inter-generational Mobility" at the University of Chicago for helpful comments and discussions. We used ChatGPT, version 4.0, for light copyediting, assistance with the LaTeX code for our computational and directed acyclic graphs, and aid in debugging Python scripts. Responsibility for all content and any potential errors in this manuscript rests solely with the authors. This research was supported by a grant from the U.S. National Science Foundation (No. 2015613).}, 
    Jesse Zhou\textsuperscript{3} \vspace{0.75cm}\\
    \textsuperscript{1}\small Department of Computer and Information Science, Link{\"o}ping University, Link{\"o}ping, Sweden \\
    \textsuperscript{2}\small Institute for Analytical Sociology, Link{\"o}ping University, Link{\"o}ping, Sweden \\
    \textsuperscript{3}\small Department of Sociology, University of Chicago, Chicago, IL, USA
\vspace{0.75cm}}

\begin{titlingpage}

\title{Deep Learning With DAGs \vspace{0.5cm}}

\maketitle

\begin{abstract}

Social science theories often postulate causal relationships among a set of variables or events. Although directed acyclic graphs (DAGs) are increasingly used to represent these theories, their full potential has not yet been realized in practice. As non-parametric causal models, DAGs require no assumptions about the functional form of the hypothesized relationships. Nevertheless, to simplify the task of empirical evaluation, researchers tend to invoke such assumptions anyway, even though they are typically arbitrary and do not reflect any theoretical content or prior knowledge. Moreover, functional form assumptions can engender bias, whenever they fail to accurately capture the complexity of the causal system under investigation. In this article, we introduce causal-graphical normalizing flows (cGNFs), a novel approach to causal inference that leverages deep neural networks to empirically evaluate theories represented as DAGs. Unlike conventional approaches, cGNFs model the full joint distribution of the data according to a DAG supplied by the analyst, without relying on stringent assumptions about functional form. In this way, the method allows for flexible, semi-parametric estimation of any causal estimand that can be identified from the DAG, including total effects, conditional effects, direct and indirect effects, and path-specific effects. We illustrate the method with a reanalysis of Blau and Duncan’s \citeyearpar{blau1967american} model of status attainment and Zhou's \citeyearpar{zhou2019selection} model of conditional versus controlled mobility. To facilitate adoption, we provide open-source software together with a series of online tutorials for implementing cGNFs. The article concludes with a discussion of current limitations and directions for future development.

\end{abstract}

\end{titlingpage}

\newpage

\section{Introduction}\label{sec:intro}

Theories in the social sciences often posit systems of causal relationships between variables. Early models of status attainment, for example, framed social mobility as the result of a multi-generational causal process \citep{becker1979equilibrium, blau1967american, haller1973status, loury1981intergen, sewell1970wiscmodel}. According to these models, an individual's social and economic standing is shaped by both ascribed characteristics and personal achievements. Specifically, factors like parental education and occupation are thought to influence college aspirations and attendance, which in turn shape career choices and other life outcomes.

Recent debates among mobility researchers have centered on the role of higher education in the status attainment process. Some argue that post-secondary education acts as a ``great equalizer," moderating the influence of family background on career success \citep{hout1988mobility, torche2011college}. Others maintain that education primarily perpetuates existing social hierarchies, with any moderating influence driven by confounding factors like motivation or ability \citep{karlson2019education, zhou2019selection}. Regardless of their position, all these perspectives share a focus on causal systems—sets of interconnected variables that generate, or limit, social mobility.

Traditionally, linear path analysis—a form of structural equation modeling—was widely used to study these causal systems, particularly in seminal studies on the status attainment process \citep{alwin1975decomposition, blau1967american, sewell1970wiscmodel}. This approach represents causal relations between variables through a set of linear and additive equations. It also allows for estimation of multiple causal effects simultaneously, facilitating the evaluation of both simple and more complex hypotheses about the causal system under investigation.

While linear path analysis is a powerful method, it has a key drawback: namely, the assumption of linearity. This assumption is problematic because most theories in the social sciences do not explicitly suggest, or even subtly imply, linear and additive relationships among variables. Moreover, the social phenomena under study are rarely, if ever, strictly linear in reality, as complex forms of non-linearity, interaction, and moderation are all endemic to causal systems involving humans \citep{abbott1988transcending, hedstrom1998social, lieberson1985making}. As a result, the use of linear path analysis often results in both an unfaithful translation of theory and an inaccurate approximation of reality.

Structural equation modeling has evolved significantly since the advent of linear path analysis, offering greater flexibility to accommodate non-linearities and interactions \citep{bollen1989structural, bollen2022fifty, kline2023principles, winship1983structural}, but these advances come with a caveat. Most applications still compel researchers to specify the exact functional form of the causal relationships under study. However, the true form of these relationships is typically unknown, and in many cases, a hypothesized form cannot even be derived from theory. Consequently, researchers often resort to arbitrary conventions or a flawed specification search \citep{maccallum1986specification, spirtes2000causation}, resulting in a disconnect between theory and method as well as potentially misleading inferences.

In response to these challenges, social scientists are increasingly using directed acyclic graphs (DAGs) to depict causal systems \citep{elwert2013graphical, knight2013causal, pearl2009causality}. As non-parametric structural equation models (SEMs), DAGs represent causal relationships between variables without presupposing their functional form. This shift has been transformative for causal modeling in the social sciences. Unlike traditional SEMs, DAGs faithfully capture the type of prior knowledge that is typically available to analysts, while requiring no assumptions about the form of the hypothesized causal relations \citep{pearl2010foundations}. 

Nevertheless, the full utility of DAGs has not yet been realized in empirical practice. When researchers move from theoretical representation to empirical evaluation, they often dilute the advantages of DAGs by reintroducing arbitrary assumptions about functional form in order to simplify the task of estimation or facilitate communication of results (e.g., \citealt{wodtke2011ipw, wodtke2016neighborhood, wodtke2017neighborhoods}). Alternatively, when DAGs are used to guide non- or semi-parametric estimation of causal effects, researchers tend to focus on a single or narrow set of estimands \citep{Daoud2023threecultures, koch2021deep, lundberg2021your}. This approach allows for estimation of selected causal relationships without stringent parametric assumptions, but most of the hypothesized causal system is not evaluated empirically, even when the data permit a broader analysis. In this way, DAGs only guide the evaluation of isolated aspects of a causal system, while other dimensions are overlooked or sidelined.

In this study, we present a new approach to causal inference that combines deep learning with DAGs to flexibly model entire causal systems. Our approach, which we call a causal-Graphical Normalizing Flow (cGNF; \citealt{balgi2022cgnf, javaloy2023causalflow, wehenkel2020GNF}), models the full joint distribution of the data, as factorized according to a DAG supplied by the analyst, using deep neural networks that impose minimal functional form restrictions on the hypothesized causal relations. Once the cGNF has been learned from data, it can be used to simulate any causal estimand identified under the DAG, including total effects, conditional effects, direct and indirect effects, and path-specific effects, among many others. Additionally, this approach offers a straightforward method for conducting sensitivity analyses, whenever certain estimands may not be identified due to unobserved confounding \citep{balgi2022rho}. cGNFs thus provide a highly versatile method for empirically evaluating theories about causal systems, without the need for restrictive parametric assumptions. 

In the sections that follow, we first offer an overview of DAGs before introducing a class of distribution models known as normalizing flows. We begin by introducing normalizing flows for univariate distributions to demonstrate foundational principles, and then we extend these flows to multivariate joint distributions. Next, we show how any DAG can be modeled as a normalizing flow, and we demonstrate how this flow can be flexibly parameterized using a special class of invertible neural networks. After a brief primer on deep learning and how these networks are trained, we then illustrate the method by re-analyzing two seminal studies of social mobility: Blau and Duncan's \citeyearpar{blau1967american} model of status attainment and Zhou's \citeyearpar{zhou2019selection} model of conditional versus controlled mobility. We conclude with a discussion of current limitations and directions for future development.

To facilitate adoption, we provide open-source software for implementing cGNFs in Python and R, tailored for common applications in the social sciences. We also provide a series of online tutorials to help acquaint researchers with the software, its code, and the associated workflow. All these resources are available at \url{https://github.com/cGNF-Dev}.

\section{Directed Acyclic Graphs as Non-parametric SEMs} \label{sec:DAGs_NPSEMs}

Directed acyclic graphs (DAGs) represent causal relationships among a set of variables \citep{elwert2013graphical, pearl2009causality, pearl2010foundations}. They consist of nodes and directed edges. The nodes symbolize variables, while the directed edges between nodes represent causal effects of arbitrary form. The orientation of the edges signals the direction of influence from one variable to another, and the term ``acyclic" specifies that the graph must not include cycles. In other words, traversing the directed edges from any starting node should never loop back to the point of origin.

\begin{figure}[!ht]
\begin{centering}
\begin{tikzpicture}[yscale = 3, xscale = 3]

	\node[text centered] at (-1,0.5) (Uv1) {$\epsilon_{V_1}$};
	\node[text centered] at (-1,0) (v1) {$V_1$};
	\node[text centered] at (0,-1.15) (Uv2) {$\epsilon_{V_2}$};
	\node[text centered] at (0,-0.65) (v2) {$V_2$};
 	\node[text centered] at (0,0.5) (Uv3) {$\epsilon_{V_3}$};
	\node[text centered] at (0,0) (v3) {$V_3$};
 	\node[text centered] at (1,0.5) (Uv4) {$\epsilon_{V_4}$};
	\node[text centered] at (1,0) (v4) {$V_4$};
	
	\draw [->, line width= 1.25] (Uv1) -- (v1);
	\draw [->, line width= 1.25] (Uv2) -- (v2);
	\draw [->, line width= 1.25] (Uv3) -- (v3);
	\draw [->, line width= 1.25] (Uv4) -- (v4);
	\draw [->, line width= 1.25] (v1) -- (v2);
	\draw [->, line width= 1.25] (v1) -- (v3);
	\draw [->, line width= 1.25] (v2) -- (v3);
	\draw [->, line width= 1.25] (v2) -- (v4);
	\draw [->, line width= 1.25] (v3) -- (v4);
 
\end{tikzpicture}
\caption{A Simple Example of a Directed Acyclic Graph (DAG).\label{fig:A-Simple-DAG}}
\medskip{}
\par\end{centering}
Note: In this DAG, $V_1$ causes $V_2$ and $V_3$, $V_2$ causes $V_3$ and $V_4$, and $V_3$ causes $V_4$. The $\left\{\epsilon_{V_1},\epsilon_{V_2},\epsilon_{V_3},\epsilon_{V_4}\right\}$ terms are random disturbances.
\end{figure}
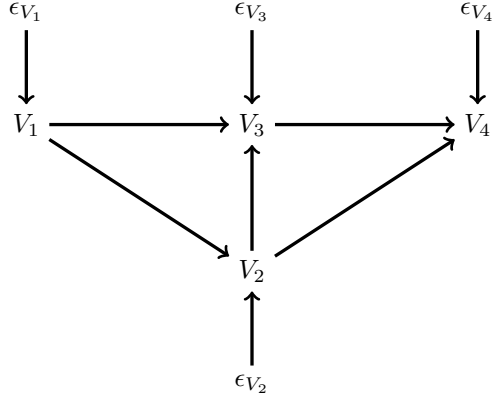

Figure \ref{fig:A-Simple-DAG} displays a simple DAG with four observed variables: $V_1$, $V_2$, $V_3$, and $V_4$, collectively denoted by $\mathbf{V}$. The directed edges in the graph establish the causal connections between these variables. Specifically, they indicate that $V_1$ causes $V_2$ and $V_3$, which in turn cause $V_4$. They also show that $V_3$ is caused by $V_2$. The epsilon terms, denoted as $\left\{\epsilon_{V_1},\epsilon_{V_2},\epsilon_{V_3},\epsilon_{V_4}\right\}$, represent random disturbances that account for unobserved factors influencing each observed variable. 

In a DAG, variables that are directly caused by a preceding variable are called its \textit{children}. Conversely, variables that directly cause a subsequent variable are identified as its \textit{parents}. To illustrate, in Figure \ref{fig:A-Simple-DAG}, $V_3$ has two parents, $V_1$ and $V_2$, but only a single child, $V_4$. 

A DAG can be interpreted as a non-parametric structural equation model (SEM), as it represents a set of causal relationships between variables without prescribing their functional form. Thus, any DAG can be translated into a corresponding system of assignment equations, where each child variable is determined by an unspecified function of its parents.

Consider, for instance, the DAG in Figure \ref{fig:A-Simple-DAG}. It can be represented using the following set of structural equations:
\begin{IEEEeqnarray}{rll}
V_1 &\enspace := \enspace & g_{V_1}(\epsilon_{V_1}) \hfill \nonumber \\
V_2 &\enspace := \enspace & g_{V_2}(V_1, \epsilon_{V_2}) \hfill \nonumber \\
V_3 &\enspace := \enspace & g_{V_3}(V_1, V_2, \epsilon_{V_3}) \hfill \nonumber \\
V_4 &\enspace := \enspace & g_{V_4}(V_2, V_3, \epsilon_{V_4}), \hfill \IEEEyesnumber \label{seq:NPSEM}
\end{IEEEeqnarray}
where the $:=$ symbol is an assignment operator used to indicate the direction of causal influence. In these equations, the set of functions $\left\{ g_{V_1},g_{V_2},g_{V_3},g_{V_4} \right\}$ do not impose any restrictions on the form of the causal relationships among variables. For example, the equation $V_3 := g_{V_3}\left(V_1, V_2, \epsilon_{V_3} \right)$ only signifies that $V_3$ is determined by an unrestricted function $g_{V_3}$ of its parent variables $V_1$ and $V_2$, along with a random disturbance $\epsilon_{V_3}$ that may follow any distribution.

The non-parametric SEM in Equation \eqref{seq:NPSEM} can also be summarized more succinctly as follows:
\begin{IEEEeqnarray}{rll}
V_i := \enspace g_{V_i}(\mathbf{V}_{i}^p, \epsilon_{V_i}), \enspace V_i \in \{V_1,V_2,V_3,V_4\} \IEEEyesnumber \label{seq:NPSEM_V}
\end{IEEEeqnarray}
where $\mathbf{V}_{i}^p$ represents the observed parents of any given variable $V_i$ in the set $\mathbf{V}=\{V_1,V_2,V_3,V_4\}$ and $\epsilon_{V_i}$ denotes the unobserved causes affecting this variable. As before, $g_{V_i}$ is an unrestricted function, requiring no commitment to any particular form for the relationship between $V_i$ and its parents.

Under the non-parametric SEM in Equation \eqref{seq:NPSEM}, a larger and more complex probability distribution for the observed data can be decomposed into several smaller, simpler distributions. These smaller distributions each involve only a subset of the observed variables, and they can be pieced back together to reconstruct the full distribution of all the variables taken together. 

In general, the joint probability distribution of any $k$ variables $\mathbf{X}=\{X_1,X_2,...,X_k\}$ can be decomposed into a product of conditional distributions.\footnote{We denote an arbitrary set of $k$ variables as $\mathbf{X}=\{X_1,X_2,...,X_k\}$, while $\mathbf{V}=\{V_1,V_2,...,V_k\}$ represents a set of variables with a defined causal structure.} For example, let $f_{\mathbf{X}}\left(x_{1},...,x_{k}\right)$ denote the joint probability that $X_{1}=x_{1}$, $X_{2}=x_{2}$, ..., and $X_{k}=x_{k}$. The \textit{product rule of joint probability} allows us to order these variables arbitrarily and then decompose their joint distribution as follows: 
\begin{IEEEeqnarray}{rll}
f_{\mathbf{X}}\left(x_{1},...,x_{k}\right) &\enspace = &\enspace f_{X_{1}}\left(x_{1}\right) f_{X_{2}|x_{1}}\left(x_{2}\right) \times ... \times f_{X_{k}|x_{1}...x_{k{-}1}}\left(x_{k}\right) \hfill \nonumber \\
&\enspace = &\enspace  f_{X_{1}}\left(x_{1}\right) \prod_{i=2}^{k}f_{X_{i}|x_{1}...x_{i{-}1}}(x_i) \hfill
\IEEEyesnumber
\label{eq:PDF_decomp}
\end{IEEEeqnarray}
where $f_{X_{i}|x_{1}...x_{i{-}1}}\left(x_{i}\right)$ denotes the conditional probability that $X_{i}=x_{i}$, given the values of $X_{1},...,X_{i{-}1}$. We refer to Equation \eqref{eq:PDF_decomp} as the \textit{autoregressive factorization} of the joint distribution, as it involves a decomposition where each variable is conditioned upon all of its predecessors.

Each variable, however, may not be sensitive to all of its predecessors. When the observed data are generated from a model resembling a DAG, variables are directly influenced only by their parents, and this enables a more economical decomposition of their joint distribution. Specifically, the joint distribution can be decomposed into a product of conditional probabilities, with each depending only on the parents of the variable in question. For example, the non-parametric SEM in Equation \eqref{seq:NPSEM} allows the following decomposition of the joint probability distribution for $\mathbf{V}$: 
\begin{IEEEeqnarray}{rll}
f_{\mathbf{V}}\left(v_1,v_2,v_3,v_4\right) &\enspace = \enspace & f_{V_1}\left(v_1\right) f_{V_2|v_1}\left(v_2\right) f_{V_3|v_{1}v_{2}}\left(v_3\right) f_{V_4|v_{2}v_{3}}\left(v_4\right) \hfill \nonumber \\
&\enspace = \enspace & \prod_{i=1}^{4}f_{V_{i}|v_{i}^{p}}\left(v_{i}\right), \enspace V_i \in \{V_1,V_2,V_3,V_4\}, \hfill \IEEEyesnumber \label{seq:PDF_NPSEM_decomp}
\end{IEEEeqnarray}
where $f_{V_i|v_{i}^p}\left(v_i\right)$ denotes the conditional probability that a variable $V_i$ takes the value $v_i$, given the values of its parents $v_{i}^p$. This decomposition is referred to as the \textit{Markov factorization} of the joint probability distribution for the observed data \citep{pearl2009causality}. 

While an observational distribution, such as Equation \eqref{seq:PDF_NPSEM_decomp}, describes the likelihood of seeing different values in the data under existing conditions, causal inference involves interventional distributions, which capture how these probabilities would change if certain variables were externally manipulated. 

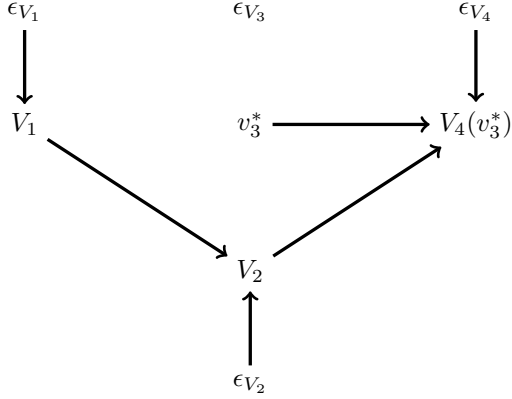
\begin{figure}[!ht]
\begin{centering}
\begin{tikzpicture}[yscale = 3, xscale = 3]
\begin{scope}

	\node[text centered] at (-1,0.5) (Uv1) {$\epsilon_{V_1}$};
	\node[text centered] at (-1,0) (v1) {$V_1$};
	\node[text centered] at (0,-1.15) (Uv2) {$\epsilon_{V_2}$};
	\node[text centered] at (0,-0.65) (v2) {$V_2$};
 	\node[text centered] at (0,0.5) (Uv3) {$\epsilon_{V_3}$};
	\node[text centered] at (0,0) (v3) {$v_{3}^*$};
 	\node[text centered] at (1,0.5) (Uv4) {$\epsilon_{V_4}$};
	\node[text centered] at (1,0) (v4) {$V_4(v_{3}^*)$};
	
	\draw [->, line width= 1.25] (Uv1) -- (v1);
	\draw [->, line width= 1.25] (Uv2) -- (v2);
	\draw [->, line width= 1.25] (Uv4) -- (v4);
	\draw [->, line width= 1.25] (v1) -- (v2);
	\draw [->, line width= 1.25] (v2) -- (v4);
	\draw [->, line width= 1.25] (v3) -- (v4);
 
\end{scope}
\end{tikzpicture}
\caption{A Mutilated Directed Acyclic Graph (DAG).\label{fig:A-Mutiliated-DAG}}
\medskip{}
\par\end{centering}
Note: In this mutilated DAG, $V_4(v_{3}^*)$ represents the potential outcome of $V_4$ under an intervention that sets $V_3$ equal to $V_{3}^{*}$.
\end{figure}

Interventions are represented in DAGs by ``mutilating" them. This involves removing incoming edges to the variable or variables being manipulated, and then setting these nodes to fixed values for deterministic interventions, or assigning them values drawn from a prescribed distribution in the case of stochastic interventions. To illustrate, consider a deterministic intervention where the variable $V_3$ is set to the value $v_{3}^{*}$ for everyone. Figure \ref{fig:A-Mutiliated-DAG} displays the mutilated version of our original DAG corresponding to this intervention. Here, the edges from $V_1$, $V_2$, and $\epsilon_{V_3}$ into $V_3$ have been deleted, and $V_3$ has been assigned the value $v_{3}^{*}$.

With a mutilated DAG, the corresponding set of non-parametric structural equations also takes a modified form. Specifically, these equations can now be expressed as follows:
\begin{IEEEeqnarray}{rll}
V_1 &\enspace := \enspace & g_{V_1}(\epsilon_{V_1}) \hfill \nonumber \\
V_2 &\enspace := \enspace & g_{V_2}(V_1, \epsilon_{V_2}) \hfill \nonumber \\
V_3 &\enspace := \enspace & v_{3}^{*} \hfill \nonumber \\
V_4(v_{3}^*) &\enspace := \enspace & g_{V_4}(V_2, v_{3}^{*}, \epsilon_{V_4}), \hfill \IEEEyesnumber \label{seq:NPSEM_mutilated}
\end{IEEEeqnarray}
where $V_4(v_{3}^*)$ denotes the potential outcome of $V_4$ when $V_3$ is set to $v_{3}^{*}$. 

By extension, the interventional joint distribution resulting from this manipulation is given as follows:
\begin{IEEEeqnarray}{rll}
f_{\mathbf{V}(v_{3}^*)}\left(v_1,v_2,v_{3}^*,v_4\right) = f_{V_1}\left(v_1\right) f_{V_{2}|v_1}\left(v_2\right) f_{V_{4}|v_{2}v_{3}^*}\left(v_4\right). \hfill \IEEEyesnumber \label{seq:PDF_NPSEM_mutilated}
\end{IEEEeqnarray}
This expression is obtained by removing the conditional probability $f_{V_{3}|v_{1}v_2}\left(v_3\right)$ for the manipulated variable $V_3$ from the Markov factorization of the joint distribution. The remaining probabilities are then conditioned on $V_3=v_{3}^{*}$ wherever this variable appears as a parent. Known as a \textit{truncated Markov factorization} \citep{pearl2009causality}, the resulting distribution, denoted by $f_{\mathbf{V}(v_{3}^*)}$, describes the joint probability of different outcomes for each variable under an intervention that exposes everyone to $v_{3}^{*}$.

With interventional distributions corresponding to different manipulations on the variable $V_3$, we can recover a variety of causal estimands, provided that they are identified from the observed data. For example, under the DAG outlined previously, the average total effect of $V_3$ on $V_4$ is given by the following expression:
\begin{IEEEeqnarray}{rll}
\mathbf{ATE}_{V_3{\rightarrow}V_4} &\enspace = \enspace & \mathbf{E}\left[V_{4}(v_{3}^{*})-V_{4}(v_{3})\right] \hfill \nonumber \\
&\enspace = \enspace & \sum_{v_{1}} \sum_{v_{2}} \sum_{v_{4}} v_4 \left(f_{\mathbf{V}(v_{3}^*)}\left(v_1,v_2,v_{3}^*,v_4\right) - f_{\mathbf{V}(v_{3})}\left(v_1,v_2,v_{3},v_4\right)\right). \hfill \IEEEyesnumber \label{seq:NPSEM_ATE_V3_to_V4}
\end{IEEEeqnarray}
In this equation, $f_{\mathbf{V}(v_{3}^*)}$ is the interventional distribution after setting $V_3$ to $v_{3}^*$, while $f_{\mathbf{V}(v_{3})}$ is the interventional distribution when $V_3$ is set to a different value $v_3$.\footnote{If these variables were strictly continuous, the probability-weighted sum in Equation \eqref{seq:NPSEM_ATE_V3_to_V4} would just be replaced with a density-weighted integral.}

A similar procedure can be used to recover other estimands identified from the DAG, such as the $\mathbf{ATE}_{V_2{\rightarrow}V_4}$ or the $\mathbf{ATE}_{V_2{\rightarrow}V_3}$, among a variety of other possibilities. In each case, the appropriate outcome variable is averaged over the relevant interventional distributions, and then the resulting averages are compared. 

Another approach to obtaining these estimands involves Monte Carlo sampling from the relevant interventional distributions, and then averaging these samples together. For example, the average total effect of $V_3$ on $V_4$ can also be formulated as follows:
\begin{IEEEeqnarray}{rll}
\mathbf{ATE}_{V_3{\rightarrow}V_4} &\enspace = \enspace & \mathbf{E}\left[V_{4}(v_{3}^{*})-V_{4}(v_{3})\right] \hfill \nonumber \\
&\enspace = &\enspace \lim_{J\rightarrow\infty}\frac{1}{J} \sum_{j=1}^J \tilde{V_{4}}^{j}(v_{3}^{*}) - \tilde{V_{4}}^{j}(v_{3}), \hfill \IEEEyesnumber \label{seq:MCE_ATE_V3_to_V4}
\end{IEEEeqnarray}
where $\tilde{V_{4}}^{j}(v_{3}^{*})$ and $\tilde{V_{4}}^{j}(v_{3})$ denote Monte Carlo samples drawn from interventional distributions with $V_3$ set to $v_{3}^*$ and $v_3$, respectively. As the total number of samples $J$ approaches infinity, the target estimand is recovered exactly.

In sum, the Markov factorization of the joint distribution, together with the interventional distributions obtained by truncating it in different ways, provide access to all the causal estimands that can be non-parametrically identified from a given DAG. If we had a distribution model based on the Markov factorization, we could then construct interventional distributions and draw Monte Carlo samples from them in order to quantify all these effects. In the next section, we introduce a class of distribution models known as normalizing flows, which are well-suited to this end. 

\section{An Introduction to Normalizing Flows} \label{sec:nf_intro}

A normalizing flow is an invertible transformation designed to map one variable--or a set of variables--onto another, which follows a standard normal distribution \citep{kobyzev2020NF, papamakarios2021NF_pmi, rezende2015variationalNF, tabak2010density, tabak2013family}. This mapping can then be used to simulate new data from either an observational or interventional distribution. This is achieved by drawing Monte Carlo samples from the standard normal distribution and then transforming these samples via the inverse of the flow.

\subsection{Univariate Normalizing Flows} \label{subsec:nf_uni}

For a single variable $X_{1}$ with a probability distribution $f_{X_{1}}$, a normalizing flow can be formally defined as follows:
\begin{IEEEeqnarray}{rll}
Z_{1} = \boldsymbol{h}(X_{1}) \sim \mathcal{N}(0,1), \hfill \IEEEyesnumber
\label{eq:uni_norm_flow}
\end{IEEEeqnarray}
where $\boldsymbol{h}$ denotes a function or composition of multiple functions that map $X_{1}$ onto another variable $Z_{1}$ equipped with the standard normal distribution. Essentially, normalizing flows are just a transformation of one variable with an arbitrary distribution into another that is normally distributed with zero mean and unit variance.

Normalizing flows accommodate discrete variables by first dequantizing them \citep{balgi2022cgnf, Uria2013RNADETR_uniformdeq, ZieglerR2019VAE_dequant}. Dequantization turns integers, like zeros and ones, into continuous values by adding a small amount of random noise, usually drawn from a uniform or normal distribution with minuscule variance. This recasts discrete variables as continuous, making them suitable for subsequent normalizing transformations. Conversely, discrete variables can be easily restored by rounding their dequantized values to the nearest integer. Thus, dequantization enables normalizing flows to map any type of variable--binary, ordinal, polytomous, and so on--to the standard normal distribution. In Part \ref{subsec:appendix_dequant} of the Appendix, we provide additional details on the use of dequantization with normalizing flows for discrete data.

In general, the functions that compose a normalizing flow $\boldsymbol{h}$ may assume any form, provided they are \textit{bijective}--that is, as long as they map each input to a single unique output. This constraint ensures that the standard normal variable $Z_{1}$ can be mapped back to the original variable $X_{1}$ by applying the inverse of the flow:
\begin{IEEEeqnarray}{rll}
X_{1} = \boldsymbol{h}^{-1}(Z_{1}) \sim f_{X_{1}}. \hfill \IEEEyesnumber
\label{eq:uni_norm_flow_inv}
\end{IEEEeqnarray}
It also allows for convenient Monte Carlo sampling from the arbitrary distribution $f_{X_{1}}$. This is accomplished by initially drawing Monte Carlo samples from the standard normal distribution and then transforming these samples using the inverse of the flow.

In addition, the transformation $\boldsymbol{h}$ encodes a particular form for the distribution of $X_{1}$. This form is given by the change of variables formula and can be expressed as follows:
\begin{IEEEeqnarray}{rll}
f_{X_{1}}(x_1) = f_{Z_{1}}(\boldsymbol{h}(x_1)) \left|\frac{\partial \boldsymbol{h}}{\partial x_{1}}\right|.  \hfill \IEEEyesnumber
\label{eq:uni_norm_flow_density}
\end{IEEEeqnarray}
In this expression, $f_{Z_{1}}$ represents the standard normal distribution, while $\big|\frac{\partial \boldsymbol{h}}{\partial x_{1}}\big|$ is the absolute value of the derivative of the normalizing flow with respect to $x_{1}$. This latter term captures the extent to which the transformation $\boldsymbol{h}$ modifies $X_{1}$ so that it conforms to a standard normal distribution.

To illustrate, consider a simple and familiar example. Suppose that $X_{1}$ were normally distributed with mean $\mu$ and variance $\sigma^2$. In this case, a normalizing flow for $X_{1}$ could be constructed using a single linear transformation:
\begin{IEEEeqnarray}{rll}
Z_1 = \boldsymbol{h}(X_{1}) = \frac{X_1-\mu}{\sigma} \sim \mathcal{N}(0,1). \hfill \IEEEyesnumber \label{eq:uni_norm_flow_stdnorm}
\end{IEEEeqnarray}
Conversely, inverting the flow maps the standard normal variable $Z_{1}$ back to $X_{1}$ as follows:
\begin{IEEEeqnarray}{rll}
X_1 = \boldsymbol{h}^{-1}(Z_{1}) = \sigma Z_1 + \mu \sim \mathcal{N}(\mu,\sigma^2). \hfill \IEEEyesnumber \label{eq:uni_stdnorm_invflow_norm}
\end{IEEEeqnarray}
This flow just mirrors the common practice of standardizing a normally distributed variable to compute z-scores. Here, a simple linear transformation converts a variable with an arbitrary normal distribution into another variable that adheres to the standard normal distribution.

Although this example serves as a useful illustration, it is contrived because the distribution of $X_1$ has a known parametric form, $f_{X_{1}}=\mathcal{N}(\mu,\sigma^2)$. Whenever this distribution is known, a normalizing flow can be easily derived, where $X_1$ and $Z_1$ can be mapped back and forth from one to another using relatively simple analytic expressions. In practice, however, the distribution $f_{X_{1}}$ and the normalizing flow $\boldsymbol{h}$ are typically unknown.\footnote{In general, the function $\boldsymbol{h}$ can be conceptualized as the composition of the inverse of the standard normal cumulative distribution function (CDF) and an arbitrary CDF for the variable of interest, $X_{1}$, provided that it is smooth with a finite first derivative. The composition of the inverse normal CDF with the CDF for $X_{1}$ yields a normalizing transformation because $F_{X_{1}}(X_{1}) = U_{1} \sim \text{Uniform}(0,1)$ and $F_{Z}^{-1}(U_1) = Z_1 \sim \mathcal{N}(0,1)$, where $F_{X_{1}}$ denotes the CDF of $X_{1}$ and $F_{Z}$ denotes the standard normal CDF. If $X_{1}$ is discrete and thus $F_{X_{1}}$ is not smooth, the variable can be dequantized using a smooth distribution for the added random noise, as detailed in Part \ref{subsec:appendix_dequant} of the Appendix.} In this situation, these functions must be inferred by fitting a highly expressive model, such as a deep neural network, to the available data. We address this challenge in the subsequent section, after extending normalizing flows to multivariate distributions.

\subsection{Multivariate Normalizing Flows} \label{subsec:nf_multi}

For a set of random variables $\mathbf{X}=\left\{ X_{1},...,X_{k}\right\}$ with a joint probability distribution $f_{\mathbf{X}}$, a multivariate normalizing flow can be formally defined as follows:
\begin{IEEEeqnarray}{rll}
\mathbf{Z} = \boldsymbol{h}(\mathbf{X}) \sim \mathcal{N}(\mathbf{0},\mathbf{I}). \hfill \IEEEyesnumber
\label{eq:multi_norm_flow}
\end{IEEEeqnarray}
Similar to the univariate case, $\boldsymbol{h}$ represents a composition of bijective functions that transform the set of variables $\mathbf{X}$ into another set $\mathbf{Z}=\left\{Z_{1},...,Z_{k}\right\}$, which follows a multivariate standard normal distribution. This distribution is denoted as $\mathcal{N}(\mathbf{0},\mathbf{I})$, where $\mathbf{I}$ is the identity matrix, indicating that all elements of $\mathbf{Z}$ are uncorrelated by construction.

The set of variables $\mathbf{Z}$ can be mapped back to $\mathbf{X}$ using the inverse of the flow as follows:
\begin{IEEEeqnarray}{rll}
\mathbf{X} = \boldsymbol{h}^{-1}(\mathbf{Z}) \sim f_{\mathbf{X}}. \hfill \IEEEyesnumber
\label{eq:multi_norm_flow_inv}
\end{IEEEeqnarray}
This allows for convenient Monte Carlo sampling--now from the joint distribution $f_{\mathbf{X}}$--where samples are initially drawn from $\mathcal{N}(\mathbf{0},\mathbf{I})$ and then transformed with $\boldsymbol{h}^{-1}$.

The form of the joint distribution for $\mathbf{X}$ can also be expressed using the change of variables formula. Specifically, in the multivariate setting, this distribution can be expressed as follows: 
\begin{IEEEeqnarray}{rll}
f_\mathbf{X}(\mathbf{x}) = f_\mathbf{Z}(\boldsymbol{h}(\mathbf{x})) \left|\mathrm{det}\mathbb{J}_{\boldsymbol{h}(\mathbf{x})}\right|, \hfill \IEEEyesnumber
\label{eq:multi_norm_flow_density}
\end{IEEEeqnarray}
where $f_\mathbf{Z}$ is the multivariate standard normal distribution and $\mathbb{J}_{\boldsymbol{h}(\mathbf{x})}$ is the Jacobian matrix associated with $\boldsymbol{h}(\mathbf{x})$. This matrix contains all the partial first derivatives of the transformation $\boldsymbol{h}(\mathbf{x})$ with respect to each component of $\mathbf{x}$. The term $\big|\mathrm{det}\mathbb{J}_{\boldsymbol{h}(\mathbf{x})}\big|$ represents the absolute value of its determinant. Conceptually, this determinant captures how the transformation $\boldsymbol{h}$ modifies each element of $\mathbf{x}$ to map these variables onto another set that follows a multivariate standard normal distribution.

Although the transformations that compose $\boldsymbol{h}$ can take any form as long as they are bijective, we focus on a subclass of multivariate flows with an autoregressive structure \citep{bengio1999modeling, frey1998graphical, kingma2016IAF, kobyzev2020NF}. These flows can be formally represented as follows:
\begin{IEEEeqnarray}{rll}
\mathbf{Z} &\enspace = &\enspace \boldsymbol{h}(\mathbf{X}) \hfill \nonumber \\
&\enspace = &\enspace \left\{h_{1}\left(X_{1};c_1\right),...,h_{i}\left(X_{i};c_{i}\left(X_1,...,X_{i{-}1}\right)\right),...,h_{k}\left(X_k;c_k\left(X_1,...,X_{k-1}\right)\right)\right\}. \hfill \IEEEyesnumber \label{eq:autoreg_flow}
\end{IEEEeqnarray}
In this expression, $c_i$ is a function of the first $i{-}1$ variables in $\mathbf{X}$, known as a \textit{conditioner}, while each transformation $h_{i}$, here and henceforth referred to as a \textit{normalizer}, is a function of its conditioner and the variable $X_i$. In substantive terms, an autoregressive flow like Equation \eqref{eq:autoreg_flow} orders the elements of $\mathbf{X}$ arbitrarily and then transforms each variable $X_i$ into a new variable $Z_i$ that follows a standard normal distribution, conditional on its predecessors $X_1,...,X_{i{-}1}$.\footnote{With autoregressive flows, each function $h_{i}$ can be conceptualized as the composition of the inverse of the standard normal cumulative distribution function (CDF) with an arbitrary \textit{conditional} CDF for $X_{i}$, given its predecessors.} Each conditioner $c_i$ determines the location of the distribution for $X_i$ as a function of $X_1,..., X_{i-1}$, and each normalizer $h_i$ adjusts the shape of this distribution, based on its location given by the conditioner, to follow a standard normal curve.\footnote{The conditioner for $X_1$, denoted by $c_1$, degenerates into a constant because it does not depend on any preceding variables.}

The joint distribution for $\mathbf{X}$ that follows from an autoregressive flow and the change of variables formula is given by:
\begin{IEEEeqnarray}{rll}
f_\mathbf{X}(\mathbf{x}) &\enspace = &\enspace f_\mathbf{Z}(\boldsymbol{h}(\mathbf{x})) \left|\mathrm{det}\mathbb{J}_{\boldsymbol{h}(\mathbf{x})}\right| \hfill \nonumber \\
&\enspace = &\enspace f_{Z}\left(h_1\left(x_1;c_1\right)\right) \prod_{i=2}^{k}f_{Z}\left(h_{i}\left(x_{i};c_{i}\left(x_{1},...,x_{i{-}1}\right)\right)\right) \left|\mathrm{det}\mathbb{J}_{\boldsymbol{h}(\mathbf{x})}\right| \hfill \nonumber \\
&\enspace = &\enspace f_{Z}\left(h_1\left(x_1;c_1\right)\right) \left|\frac{\partial h_{1}}{\partial x_{1}}\right| \prod_{i=2}^{k}f_{Z}\left(h_{i}\left(x_{i};c_{i}\left(x_{1},...,x_{i{-}1}\right)\right)\right) \left|\frac{\partial h_{i}}{\partial x_{i}}\right|, \IEEEyesnumber 
\label{eq:autoreg_flow_density}
\end{IEEEeqnarray}
where $f_{Z}$ represents the univariate standard normal distribution. In this expression, the second equality comes from factorizing the joint distribution of $\mathbf{Z}$ as $f_\mathbf{Z}(\mathbf{z})=\prod_{i=1}^{k}f_{Z}(z_i)$ using the product rule for independent and identically distributed variables. The final equality arises from the triangular structure of the Jacobian matrix associated with an autoregressive flow. With a triangular Jacobian matrix, its determinant is the product of its diagonal elements--that is, $\mathrm{det}\mathbb{J}_{\boldsymbol{h}(\mathbf{x})} = \prod_{i=1}^{k} \frac{\partial h_{i}}{\partial x_{i}}$.

The expression given by Equation \eqref{eq:autoreg_flow_density} closely resembles the joint distribution of $\mathbf{X}$ as factorized in Equation \eqref{eq:PDF_decomp} from the previous section. This is because the term $f_{Z}\left(h_1\left(x_1,c_1\right)\right) \big|\frac{\partial h_{1}}{\partial x_{1}}\big|$ encodes the marginal probability $f_{X_{1}}\left(x_{1}\right)$, while each component of the product, denoted by $f_{Z}\left(h_{i}\left(x_{i};c_{i}\left(x_{1},...,x_{i{-}1}\right)\right)\right)\big|\frac{\partial h_{i}}{\partial x_{i}}\big|$ for $i=2,...,k$, encodes the conditional probability $f_{X_{i}|x_{1}...x_{i{-}1}}\left(x_{i}\right)$. Thus, autoregressive flows are built upon the autoregressive factorization of the joint distribution $f_\mathbf{X}(\mathbf{x})$, which does not rely on any independence restrictions among the elements of $\mathbf{X}$ nor any predefined ordering of these variables.

What if we constructed a normalizing flow similar to Equation \eqref{eq:autoreg_flow}, but built upon the Markov factorization of the joint distribution, as given by an assumed DAG, instead of the autoregressive factorization? This approach would enable convenient Monte Carlo sampling not only from the observational joint distribution but also from a broad array of interventional distributions obtained by appropriately truncating the flow. This is the conceptual foundation underlying causal-graphical normalizing flows, which we introduce in the next section.

\section{causal-Graphical Normalizing Flows} \label{sec:cGNF}

A causal-graphical normalizing flow (cGNF) resembles an autoregressive flow, but with a key distinction: the conditioner for each variable is a function of its parents, as indicated by a directed acyclic graph (DAG). Specifically, for a set of $k$ causally ordered variables $\mathbf{V}=\{V_1,...,V_k\}$, a cGNF can be formulated as follows:
\begin{IEEEeqnarray}{rll}
\mathbf{Z} &\enspace = &\enspace \boldsymbol{h}(\mathbf{V}) \hfill \nonumber \\
&\enspace = &\enspace \left\{h_{1}\left(V_{1};c_1\right),...,h_{i}\left(V_{i};c_{i}\left(\mathbf{V}_{i}^p\right)\right),...,h_{k}\left(X_k;c_k\left(\mathbf{V}_{k}^p\right)\right)\right\}. \hfill \IEEEyesnumber \label{eq:cGNF}
\end{IEEEeqnarray}
In this expression, the conditioner $c_i$ depends on $\mathbf{V}_{i}^p$, which represents the parents of the variable $V_i$. The normalizer, denoted by $h_i$, is a function of both the conditioner and $V_i$ in turn. 

Essentially, a cGNF arranges the elements of $\mathbf{V}$ in causal order. It then maps each variable $V_i$ to a new variable $Z_i$, which follows a standard normal distribution conditional on the parents of $V_i$. To this end, each conditioner $c_i$ shifts the location of the distribution for $V_i$ as a function of its parents $\mathbf{V}_{i}^p$. The normalizer $h_i$ then transforms the shape of this distribution, given its location from the conditioner, to resemble the standard normal curve. The causal order and parent-child relationships among the variables of interest all come from a DAG supplied by the analyst.

The conditioners and normalizers within a cGNF are unknown functions that may be quite complex. To model these functions, we use a special class of artificial neural networks. Not only do these networks satisfy all the conditions that define a bijective map, they are also capable of approximating \textit{any} monotonic transformation of one variable or set of variables into another \citep{Huang2018NAF, wehenkel2019UMNN, wehenkel2020GNF}.

Artificial neural networks draw their inspiration from the structure and function of human brains \citep{chollet2021deep, goodfellow2016deep, lecun2015deep}. They consist of multiple interconnected nodes, also referred to as ``neurons" or ``units," organized into multiple layers. A typical architecture for a deep neural network includes an input layer, several intermediary or ``hidden" layers, and an output layer.

Data flows through a deep neural network in a hierarchical manner. Nodes within each layer receive inputs from preceding layers and then generate an output. These outputs, in turn, serve as the inputs for nodes in subsequent layers. The number of layers and the quantity of nodes within each layer shape the network's expressiveness--that is, its ability to approximate a wide range of complex functions. 

The connections between nodes in adjacent layers are governed by a set of weights and activation functions. Specifically, each node produces an output by applying an activation function to the weighted sum of its inputs. 

The weights that link nodes across layers are parameters that must be learned, or estimated, from data. The process of estimating these weights is known as \textit{training}. During training, the weights are initialized at random values and then adjusted using an algorithm designed to make the network's output as accurate as possible.

\begin{figure}[!hb]
\begin{centering}
\begin{tikzpicture}[shorten >=1pt,->,draw=gray!50, node distance=\layersep]

    \tikzstyle{every pin edge}=[<-,shorten <=1pt]
    \tikzstyle{neuron}=[circle,fill=black!25,minimum size=17pt,inner sep=0pt]
    \tikzstyle{input neuron}=[neuron, fill=gray!50];
    \tikzstyle{output neuron}=[neuron, fill=gray!50];
    \tikzstyle{hidden neuron}=[neuron, fill=gray!50];
    \tikzstyle{annot} = [text width=4em, text centered]

    \def\layersep{3cm}
    \newcounter{weightcounter}

    \foreach \name / \y in {1/-0.8,2/0.8}
    \path[yshift=0cm]
    node[input neuron, pin=left:$\mathrm{Input}_\name$] (I-\name) at (0,-\y cm) {};

    \foreach \name / \y in {1/-1.5,2/0,3/1.5}
    \path[yshift=0cm, xshift=1.5cm]
    node[hidden neuron] (H-\name) at (\layersep,-\y cm) {};
    \node[annot, above of=H-2, node distance=2.35cm] (hl) {Hidden Layer};

    \node[output neuron, pin={[pin edge={->}]right:Output}, right of=H-2] (O) at (2*\layersep, 0cm) {};

    \foreach \dest in {1,2,3} {
        \stepcounter{weightcounter}
        \path (I-1) edge node[pos=0.3, above] {$w_{\the\value{weightcounter}}$} (H-\dest);
    }
    \foreach \dest in {1,2,3} {
        \stepcounter{weightcounter}
        \path (I-2) edge node[pos=0.3, below] {$w_{\the\value{weightcounter}}$} (H-\dest);
    }
    \foreach \source in {1,2,3} {
        \stepcounter{weightcounter}
        \path (H-\source) edge node[pos=0.4, above] {$w_{\the\value{weightcounter}}$} (O);
    }
    
\end{tikzpicture}
\caption{An Example of a Deep Neural Network Depicted as a Computational Graph. \label{fig:A-Deep-NeuralNet}}
\medskip{}
\par\end{centering}
Note: In this network, there is an input layer with two nodes, a hidden layer with three nodes, and an output layer with a single node. Each layer is fully connected, as indicated by the arrows between nodes, and each connection between nodes is controlled by a weight $w_i$. 
\end{figure}
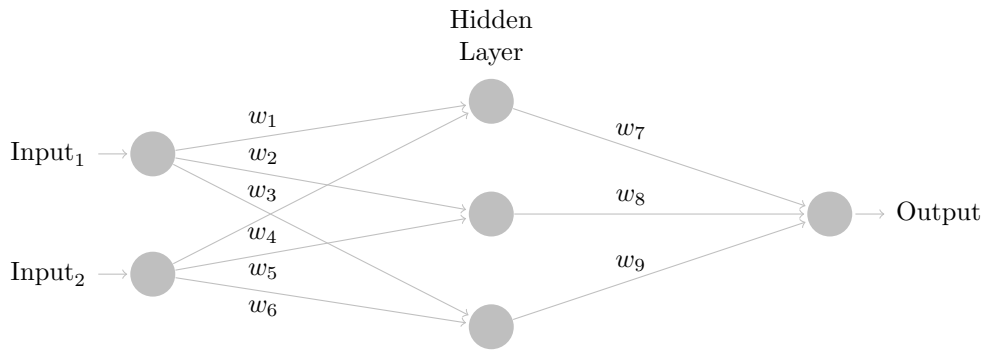

\clearpage

Figure \ref{fig:A-Deep-NeuralNet} illustrates a generic example of a deep neural network using a computational graph. By combining nodes, weights, and activation functions, layer upon layer, networks with this basic structure can approximate highly complex functions. Indeed, networks with sufficiently flexible architectures are \textit{universal function approximators}, capable of accurately modeling any continuous mapping from one finite-dimensional space to another \citep{hornik1989multilayer}.

In the following sections, we outline the specialized architecture of the neural networks used to model a cGNF, and discuss how these networks are trained. Next, we explain how a trained cGNF facilitates Monte Carlo sampling from the observational joint distribution of the data. Finally, we show how a trained cGNF also enables Monte Carlo sampling from interventional distributions, allowing for causal inference.

\subsection{Parameterizing cGNFs}

We use unconstrained monotonic neural networks (UMNNs) to model the transformations that compose a cGNF \citep{wehenkel2019UMNN, wehenkel2020GNF}. UMNNs are invertible and differentiable networks capable of approximating any monotonic transformation. Their architecture is based on the principle that a monotonic function must exhibit a strictly positive derivative, which can then be integrated to yield the desired transformation. Specifically, when used to model the transformations in a cGNF, these networks can be represented as follows:
\begin{IEEEeqnarray}{rll}
h_{i}\left(V_{i};c_{i}\left(\mathbf{V}_{i}^{p}\right);\boldsymbol{\theta}_i\right)=\int_{0}^{V_{i}}\beta_{i}\left(t;c_{i}\left(\mathbf{V}_{i}^{p};\psi_{i}\right);\phi_{i}\right)\mathrm{d}t+\alpha_{i}\left(c_{i}\left(\mathbf{V}_{i}^{p};\psi_{i}\right)\right), \hfill \IEEEyesnumber \label{seq:cGNF_UMNN}
\end{IEEEeqnarray}
where $\boldsymbol{\theta}_i$ denotes the union of the parameters $\phi_i$ and $\psi_i$.

In Equation \eqref{seq:cGNF_UMNN}, the conditioner is now modeled using a deep neural network, denoted by $c_{i}\left(\mathbf{V}_{i}^{p};\psi_{i}\right)$. To produce its output, this network takes the parents of $V_i$ as inputs and then transforms them using a set of weights, denoted by $\psi_{i}$, together with the rectified linear unit (ReLU) activation function.\footnote{The ReLU activation function returns the value zero for any negative input, but for any positive input, it returns the value of the input itself. Formally, this function can be expressed as $f(x)=\max(0,x)$.} Termed the \textit{embedding network}, $c_{i}\left(\mathbf{V}_{i}^{p};\psi_{i}\right)$ models how $V_i$ varies as a function of its parents $\mathbf{V}_{i}^{p}$ in the assumed DAG.

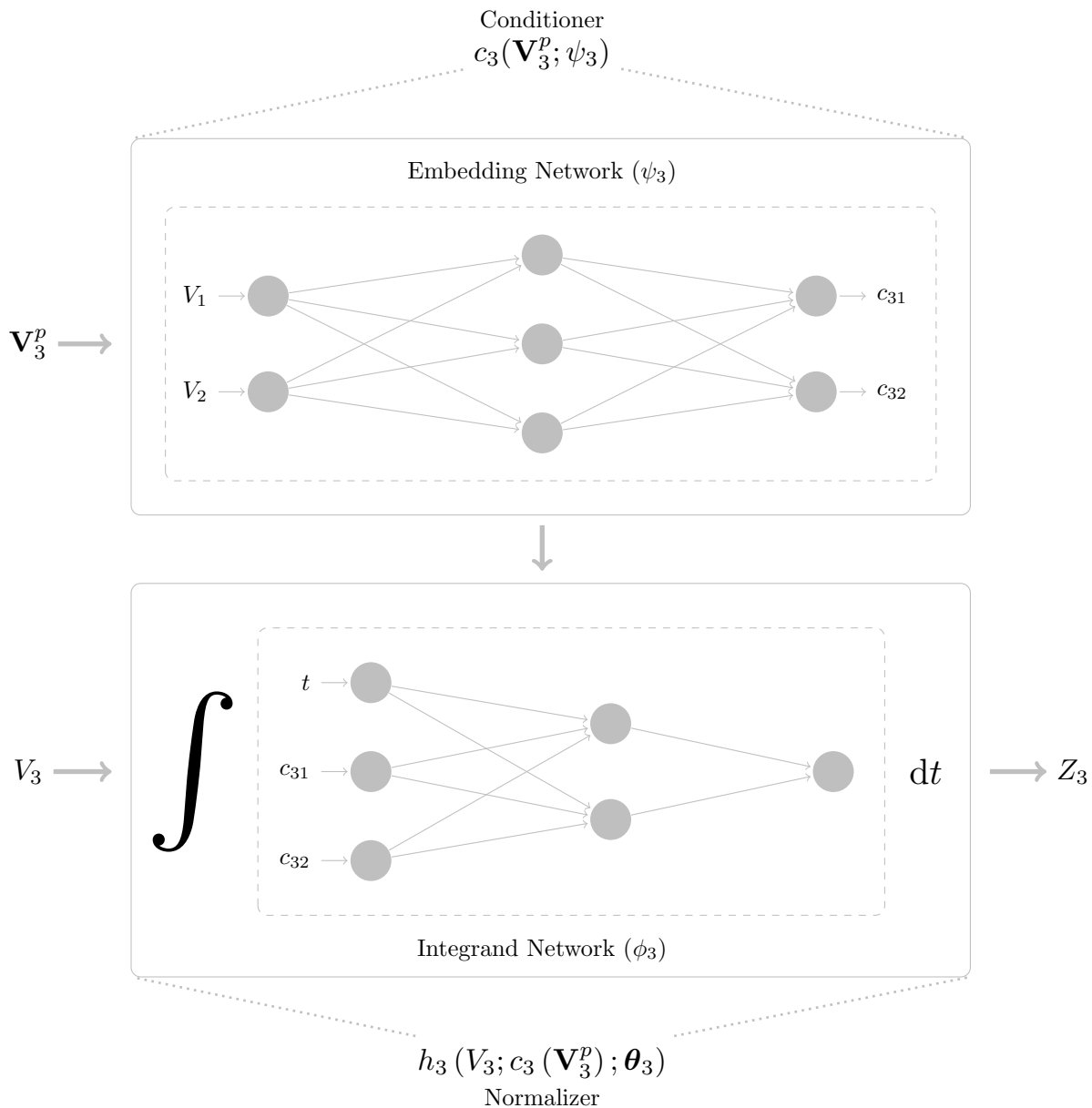
\begin{figure}
\begin{centering}
\begin{tikzpicture}[shorten >=1pt,->,draw=gray!50]

    \tikzstyle{every pin edge}=[<-,shorten <=1pt]
    \tikzstyle{neuron}=[circle,fill=black!25,minimum size=17pt,inner sep=0pt]
    \tikzstyle{input neuron}=[neuron, fill=gray!50];
    \tikzstyle{output neuron}=[neuron, fill=gray!50];
    \tikzstyle{hidden neuron}=[neuron, fill=gray!50];
    \tikzstyle{annot} = [text width=4em, text centered]

    \begin{scope}
        \node[scale=1.2] (V3p) at (-3.5cm, 0cm) {$\mathbf{V}_{3}^{p}$};
        \draw[->, line width=2pt] (V3p) -- (-2.25cm, 0cm);
    
        \foreach \name / \y in {1/-0.7,2/0.7}
        \path[yshift=0cm]
        node[input neuron, pin=left:$V_\name$] (I-\name) at (0,-\y cm) {};
    
        \foreach \name / \y in {1/-1.3,2/0,3/1.3}
        \path[yshift=0cm, xshift=1.5cm]
        node[hidden neuron] (H-\name) at (2.5cm,-\y cm) {};
    
        \node[output neuron, pin={[pin edge={->}]right:$c_{31}$}, right of=H-2] (O1) at (7cm, 0.7cm) {};
        \node[output neuron, pin={[pin edge={->}]right:$c_{32}$}, right of=H-2] (O2) at (7cm, -0.7cm) {};
    
        \foreach \source in {1,2}
            \foreach \dest in {1,2,3}
                \path (I-\source) edge (H-\dest);
    
        \foreach \source in {1,2,3}
            \path (H-\source) edge (O1);
        \foreach \source in {1,2,3}
            \path (H-\source) edge (O2);
    
        \draw[rounded corners, dashed] (-1.5cm, -2cm) rectangle (9.75cm, 2cm);
        \draw[rounded corners, solid] (-2cm, -2.5cm) rectangle (10.25cm, 3cm);
    
        \node at (4cm, 2.5cm) {Embedding Network ($\psi_{3}$)};
        \node at (4cm, 4.75cm) {Conditioner};
        \node[scale=1.25] (C3) at (4cm, 4.25cm) {$c_3(\mathbf{V}_{3}^{p};\psi_{3})$};
        \draw[dotted, -, line width=1pt] (C3) -- (-2cm, 3cm);
        \draw[dotted, -, line width=1pt] (C3) -- (10.25cm, 3cm);
    \end{scope}

    \begin{scope}[yshift=-6.25cm] 
        \node[input neuron, pin=left:$t$] (I-1) at (1.5, 1.3cm) {};
        \node[input neuron, pin=left:$c_{31}$] (I-2) at (1.5, 0cm) {};
        \node[input neuron, pin=left:$c_{32}$] (I-3) at (1.5, -1.3cm) {};
        
        \foreach \name / \y in {1/-0.7,2/0.7}
        \path[yshift=0cm, xshift=3.5cm]
        node[hidden neuron] (H-\name) at (1.5cm,-\y cm) {};
    
        \node[output neuron, right of=H-1] (O) at (7.25cm, 0cm) {};
    
        \foreach \source in {1,2,3}
            \foreach \dest in {1,2}
                \path (I-\source) edge (H-\dest);
    
        \foreach \source in {1,2}
            \path (H-\source) edge (O);
    
        \draw[rounded corners, dashed] (-0.15cm, -2.1cm) rectangle (9cm, 2.1cm);
        \draw[rounded corners, solid] (-2cm, -3cm) rectangle (10.25cm, 2.75cm);
        \node at (4cm, -2.6cm) {Integrand Network ($\phi_3$)};
        \node at (4cm, -4.75cm) {Normalizer};
        \node[scale=1.25] (h3) at (4cm, -4.25cm) {$h_3\left(V_{3};c_{3}\left(\mathbf{V}_{3}^{p}\right);\boldsymbol{\theta}_3\right)$};
        \draw[dotted, -, line width=1pt] (h3) -- (-2cm, -3cm);
        \draw[dotted, -, line width=1pt] (h3) -- (10.25cm, -3cm);
    
        \node[scale=6] at (-1.1cm, 0cm) {$\int$};
        \node[scale=1.5] at (9.6cm, 0cm) {d$t$};
        
        \node[scale=1.2] (V3) at (-3.5cm, 0cm) {$V_3$};
        \draw[->, line width=2pt] (V3) -- (-2.25cm, 0cm);
        \node[scale=1.2] (Z3) at (11.75cm, 0cm) {$Z_3$};
        \draw[<-, line width=2pt] (Z3) -- (10.5cm, 0cm);
    \end{scope}

    \draw[->, line width=2pt] (4cm, -2.65cm) -- (4cm, -3.35cm);

\end{tikzpicture}
\caption{An Unconstrained Monotonic Neural Network (UMNN) Depicted as a Computational Graph. \label{fig:A-UMNN-CompGraph}}
\bigskip{}
\par\end{centering}
Note: This figure illustrates a simple UMNN for variable $V_3$ in our DAG from Figure \ref{fig:A-Simple-DAG}. The conditioner, $c_3$, is modeled using an embedding network. This network has an input layer with two nodes, which correspond to the parents of $V_3$ (i.e., $V_1$ and $V_2$). The input layer is followed by a single hidden layer containing three nodes and an output layer with two nodes, labeled $c_{31}$ and $c_{32}$. The outputs of the embedding network serve as inputs for the integrand network. Specifically, the integrand network has an input layer with three nodes corresponding to $c_{31}$ and $c_{32}$ from the embedding network, as well as the integration points $t$. This input layer is followed by a single hidden layer with two nodes and an output layer with a single node. The normalizer, $h_3$, integrates the output of the integrand network, using $V_3$ as the upper limit of integration. This calculation yields $Z_3$, a transformed version of $V_3$ that conforms to the standard normal distribution.
\end{figure}

The output from the embedding network serves two purposes. First, it is used to generate a scalar offset term, denoted by $\alpha_{i}\left(c_{i}\left(\mathbf{V}_{i}^{p};\psi_{i}\right)\right)$. Second, it also serves as input to another neural network, denoted by $\beta_{i}\left(t;c_{i}\left(\mathbf{V}_{i}^{p};\psi_{i}\right);\phi_{i}\right)$. This other network uses a distinct set of weights, $\phi_{i}$, and a special activation function--specifically, the exponential linear unit incremented by one (ELUPlus)--to generate an output that is strictly positive.\footnote{The ELUPlus activation function is designed to produce an output greater than zero, regardless of its input. Formally, this function can be expressed as $f(x)=x+1$ if $x>0$, and $f(x)=\exp(x)$ if $x\leq0$.} Referred to as the \textit{integrand network}, it models the derivative, at a given point $t$, of a monotonic function intended to map the variable $V_i$ onto the standard normal distribution, using the output from the embedding network. 

The normalizer, then, is constructed with both the embedding and integrand networks in tandem. It comes from integrating the output of the integrand network, $\beta_{i}\left(t; c_{i}\left(\mathbf{V}_{i}^{p};\psi_{i}\right);\phi_{i}\right)$, from $t=0$ to the observed value of $V_i$, and then adding the offset term $\alpha_{i}\left(c_{i}\left(\mathbf{V}_{i}^{p};\psi_{i}\right)\right)$. The integration is performed using Clenshaw-Curtis quadrature, a numerical technique for approximating the area under a curve \citep{clenshaw1960method}. The resulting function, denoted by $h_{i}\left(V_{i};c_{i}\left(\mathbf{V}_{i}^{p}\right);\boldsymbol{\theta}_i\right)$, is an invertible, monotonic transformation of the variable $V_i$. Its parameters, $\boldsymbol{\theta}_i=\left\{\phi_{i},\psi_{i}\right\}$, are weights associated with the embedding and integrand networks. These weights can be trained to transform $V_i$ into a new variable $Z_i$, which follows the standard normal distribution conditional on its parents ${V}_{i}^{p}$. Figure \ref{fig:A-UMNN-CompGraph} illustrates a simple example of this model using a computational graph.\footnote{The UMNN described here can be conceptualized as a model for the following transformation: $z_i=F_{Z}^{-1}(F_{V_{i}|v_{i}^p}(v_{i}))$, where $F_{V_{i}|v_{i}^p}$ is the cumulative distribution function (CDF) for the variable $V_{i}$, conditional on its parents, and $F_{Z}^{-1}$ is the inverse of the standard normal CDF. The composition of the inverse normal CDF with any arbitrary CDF for a continuous variable yields a monotonic normalizing transformation.} 

Thus, for an entire set of $k$ causally ordered variables $\mathbf{V}=\{V_1,...,V_k\}$, a cGNF parameterized by $\boldsymbol{\theta}=\left\{\boldsymbol{\theta}_1,...,\boldsymbol{\theta}_k\right\}$ can be compactly expressed as follows:
\begin{IEEEeqnarray}{rll}
\mathbf{Z} = \boldsymbol{h}(\mathbf{V}; \boldsymbol{\theta}) \sim \mathcal{N}(\mathbf{0},\mathbf{I}). \hfill \IEEEyesnumber
\label{seq:multi_norm_flow_cgnf}
\end{IEEEeqnarray}
In this equation, $\mathcal{N}(\mathbf{0},\mathbf{I})$ is the multivariate standard normal distribution, $\boldsymbol{h}$ denotes the composition of normalizers for each variable in the data, and $\boldsymbol{\theta}$ represents the full collection of weights from the UMNNs used to model each normalizer.

The architecture for the embedding and integrand networks that compose a UMNN can be customized with a varying number of layers, nodes per layer, and inter-layer connections. In general, more elaborate architectures--with additional layers, nodes, and connections--are better equipped to approximate more complex transformations. When configured with a sufficiently flexible architecture, UMNNs function as \textit{universal density approximators}, capable of modeling any distribution irrespective of its complexity \citep{Huang2018NAF, wehenkel2019UMNN}. 

For empirical applications in the social sciences, we suggest architectures with at least four hidden layers, a minimum of 10 to 20 nodes per layer, and a fully connected configuration among them. For simplicity, we also recommend using identical architectures across all the normalizers, from $h_1$ to $h_k$, which does not appear to compromise their performance in practice \citep{wehenkel2019UMNN}.

Although increasing the complexity of UMNNs enables more accurate approximation, it also introduces the risk of over-fitting. Over-fitting occurs when the network begins to model random variation in the sample data used to train it, capturing noise in addition to the underlying signal. As a result, the trained network can yield estimates that are less reliable and suffer from greater uncertainty. Nevertheless, the problem of over-fitting can be mitigated through the use of regularization techniques during training, as we discuss below. Moreover, recent studies suggest that over-parameterizing neural networks, such that their number of weights exceeds the number of observations available for training, might actually enhance performance, regardless of whether regularization methods are employed \citep{allen2019learning, brutzkus2017sgd, yang2020rethinking, zhang2021understanding}.

\subsection{Training cGNFs}

To train a cGNF, we adjust the weights, denoted by $\boldsymbol{\theta}=\left\{\boldsymbol{\theta}_1,...,\boldsymbol{\theta}_k\right\}$, in the embedding and integrand networks that compose the UMNNs. When adjusting these weights, the objective is to minimize a loss function that quantifies the fit of the model to a set of observed sample data. Specifically, the loss function used to train a cGNF is the negative log-likelihood, derived from the joint distribution for $\mathbf{V}$ using the change of variables formula. This loss function can be formally expressed as follows:
\begin{IEEEeqnarray}{rll}
-\mathcal{LL}\left(\boldsymbol{\theta}\right) &\enspace = &\enspace -\ln\left(\prod_{l=1}^{n}f_{\mathbf{V}}\left(\mathbf{v}^l;\boldsymbol{\theta}\right)\right) \hfill \nonumber \\
&\enspace = &\enspace -\ln\left(\prod_{l=1}^{n}f_{\mathbf{Z}}\left(\boldsymbol{h}\left(\mathbf{v}^l;\boldsymbol{\theta}\right)\right)\left|\mathrm{det}\mathbb{J}_{\boldsymbol{h}(\mathbf{v}^l;\boldsymbol{\theta})}\right|\right) \hfill \nonumber \\
&\enspace = &\enspace -\sum_{l=1}^{n}\ln\left(f_{\mathbf{Z}}\left(\boldsymbol{h}\left(\mathbf{v}^l;\boldsymbol{\theta}\right)\right)\right)-\sum_{l=1}^{n}\ln\left(\left|\mathrm{det}\mathbb{J}_{\boldsymbol{h}(\mathbf{v}^l;\boldsymbol{\theta})}\right|\right) \hfill \nonumber \\
&\enspace = &\enspace -\sum_{l=1}^{n}\sum_{i=1}^{k}\ln\left(f_{Z}\left(h_{i}\left(v_{i}^l;c_{i}\left(\mathbf{v}_{i}^{p,l}\right);\boldsymbol{\theta}_{i}\right)\right)\right)-\sum_{l=1}^{n}\sum_{i=1}^{k}\ln\left(\left|\frac{\partial h_{i}}{\partial v_{i}^l}\right|\right), \hfill \IEEEyesnumber 
\label{seq:neg_LL_loss}
\end{IEEEeqnarray}
where $l=1,...,n$ indexes observations in the sample data. As before, $i=1,...,k$ indexes the variables under consideration, $f_\mathbf{Z}$ denotes the multivariate standard normal distribution, $f_Z$ denotes the univariate standard normal distribution, and $h_{i}\left(V_{i};c_{i}\left(\mathbf{V}_{i}^{p}\right);\boldsymbol{\theta}_i\right)$ denotes the UMNN intended to normalize each variable conditional on its parents. 

To find values for the network weights that minimize the negative log-likelihood, we use the method of \textit{stochastic gradient descent} (SGD), an algorithm that iteratively adjusts the weights based on the gradient of the loss function \citep{lecun2015deep, chollet2021deep, goodfellow2016deep}. The gradient refers to the set of partial derivatives of the loss function with respect to each weight in the network. Essentially, the gradient is like an arrow pointing toward the steepest increase in the loss function due to a small change in the weights. SGD adjusts the weights by moving them in the opposite direction of the gradient, thereby reducing the loss and improving the model's fit to the sample data. 

At each iteration of the algorithm, the weights are adjusted using the gradient computed on a random subset of the data, known as a \textit{mini-batch}, while the size of the adjustments is governed by a hyper-parameter called the \textit{learning rate}. When the learning rate is set at a small value, the weights are only adjusted a little bit after each mini-batch is processed, which prevents the algorithm from overshooting their optimal values. 

Completing one pass through all mini-batches in the sample data marks the end of one training \textit{epoch}. The algorithm adjusts the weights repeatedly, cycling through mini-batches and epochs over and over, until it reaches a stopping criterion. In our case, the algorithm terminates when further adjustments to the weights no longer reduce the loss, as measured on a separate validation sample held out from the data used for training. This stopping criterion functions as a form of implicit regularization, helping to prevent over-fitting. This is achieved by halting the training process before further adjustments to the network weights serve mainly to fit random noise in the training data. 

To summarize, the process of training a cGNF using SGD involves the following steps: 

\begin{enumerate}
\item Partition the sample data randomly into training and validation subsets.
\item Initialize the network weights with random values.
\item Randomly divide the training data into mini-batches.
\item For each mini-batch:
    \begin{enumerate}
    \item Compute the gradient of the loss function.
    \item Adjust the weights incrementally in the opposite direction of the gradient.
    \end{enumerate}
\item Evaluate the loss function using the validation data.
\item Repeat steps 3 to 5 until the validation loss ceases to improve over a set number of epochs.
\end{enumerate}

In general, we advise reserving 20 percent of the sample data for validation while training the cGNF on the remaining 80 percent. Mini-batch sizes should fall between 64 and 512 observations for best performance. We also suggest a learning rate less than or equal to 0.001 and terminating the training algorithm after the validation loss stagnates for 30 to 50 epochs.

\subsection{Sampling with cGNFs}

After the cGNF is trained, it can be used for Monte Carlo sampling from the observational joint distribution of the data. This is achieved by generating samples from the standard normal distribution and then transforming them using the inverse of the trained cGNF. The inverse of the cGNF is found using a bisection algorithm, which identifies the inverse by iteratively narrowing the range of values within which it must lie.\footnote{The algorithm starts with an interval defined by the minimum and maximum values of the function, which necessarily bound the desired output. The algorithm then finds the midpoint of these values and uses it to replace the endpoint of the current range that does not contain the inverse, halving the interval's size. This process is iterated until the interval is extremely small, and the midpoint within this range is taken as the value of the inverse. Specifically, for a value $z$ sampled from the standard normal distribution, the algorithm starts with an interval $\left[x_a,x_b\right]$ that must contain the inverse. The algorithm replaces this interval with $\left[x_a,x_{(b-a)/2}\right]$ if $h(x_{(b-a)/2})>z$, and with $\left[x_{(b-a)/2},x_b\right]$ otherwise. This process of halving the interval is repeated until its range is minuscule and, then, the midpoint of the final interval is returned as the inverse.}

To illustrate the sampling process, consider a cGNF trained with the DAG in Figure \ref{fig:A-Simple-DAG}. In this case, the sampling process begins with the first variable in causal order, denoted by $V_1$. To simulate its values, we initially create $J$ Monte Carlo samples from the standard normal distribution. These samples, each denoted by $\tilde{Z}_{1}^j$ for $j=1,...,J$, are then transformed as follows: $\tilde{V}_{1}^j = h_{1}^{-1} \left( \tilde{Z}_{1}^j; c_1; \hat{\boldsymbol{\theta}}_1 \right)$, where the ``hat" indicates that the network weights have been estimated by SGD. The result of this transformation, $\tilde{V}_{1}^j$, represents a Monte Carlo sample from the marginal distribution of $V_1$. It is obtained by applying the inverse of the normalizer for $V_1$ to a sample from the standard normal distribution.

For the next variable, $V_2$, we generate another set of Monte Carlo samples from the standard normal distribution, each denoted by ${Z}_{2}^j$ for $j=1,...,J$. We then transform these samples as follows: $\tilde{V}_{2}^j = h_{2}^{-1} \left( \tilde{Z}_{2}^j; c_2\left( \tilde{V}_{1}^j\right); \hat{\boldsymbol{\theta}}_2 \right)$, where $\tilde{V}_{1}^j$ is carried over from the previous step. The result of this transformation, $\tilde{V}_{2}^j$, represents a Monte Carlo sample from the conditional distribution of $V_2$, given its only parent $V_1$. It comes from transforming a standard normal sample using the inverse of the normalizer for $V_2$.

Continuing this process for $V_3$, the next variable in causal order, we generate a third set of Monte Carlo samples from the standard normal distribution. These samples, denoted by ${Z}_{3}^j$ for $j=1,...,J$, are then transformed as follows: $\tilde{V}_{3}^j = h_{3}^{-1} \left( \tilde{Z}_{3}^j; c_3\left( \tilde{V}_{1}^j, \tilde{V}_{2}^j \right); \hat{\boldsymbol{\theta}}_3 \right)$, where $\tilde{V}_{1}^j$ and $\tilde{V}_{2}^j$ are both carried over from the previous steps. The result of this transformation, $\tilde{V}_{3}^j$, represents a Monte Carlo sample from the conditional distribution of $V_3$, given its parents $V_1$ and $V_2$. It is obtained by passing a standard normal sample through the inverse of the normalizer for $V_3$.

For the final variable $V_4$, we generate another set of standard normal samples, each denoted by ${Z}_{4}^j$ for $j=1,...,J$. They are then subjected to the following transformation: $\tilde{V}_{4}^j = h_{4}^{-1} \left( \tilde{Z}_{4}^j; c_4\left( \tilde{V}_{2}^j, \tilde{V}_{3}^j \right); \hat{\boldsymbol{\theta}}_4 \right)$, where $\tilde{V}_{2}^j$ and $\tilde{V}_{3}^j$ are samples obtained from the previous steps. This transformation yields $\tilde{V}_{4}^j$, which represents a Monte Carlo sample from the conditional distribution of $V_4$, given its parents $V_2$ and $V_3$. As before, it comes from applying the inverse of the normalizer, now for $V_4$, to a Monte Carlo sample from the standard normal distribution.

Combining the samples for each variable together, we obtain a random vector $\tilde{\mathbf{V}}^j=\left\{ \tilde{V}_{1}^j,\tilde{V}_{2}^j,\tilde{V}_{3}^j,\tilde{V}_{4}^j \right\}$. This vector represents a Monte Carlo sample from the observational joint distribution, as approximated by the trained cGNF.

In general, for a set of $k$ variables arranged in causal order, Monte Carlo sampling from their observational joint distribution is accomplished by first generating standard normal samples, indexed by $j=1,...,J$, and then transforming them recursively as follows:
\begin{IEEEeqnarray}{rll}
\tilde{V}_{i}^j = h_{i}^{-1} \left( \tilde{Z}_{i}^j; c_i\left( \tilde{\mathbf{V}}_{i}^{p,j} \right); \hat{\boldsymbol{\theta}}_i \right)  \enspace \mathrm{for} \enspace i=1,...,k. \hfill \IEEEyesnumber 
\label{eq:cGNF_obv_inv}
\end{IEEEeqnarray}
In this expression, $\tilde{Z}_{i}^j$ denotes a standard normal sample, $h_{i}^{-1}$ represents the inverse of the normalizer for variable $V_i$, and $\tilde{\mathbf{V}}_{i}^{p,j}$ denotes the simulated values for the parents of $V_i$, which are obtained from previous steps in the sampling algorithm. Transforming $\tilde{Z}_{i}^j$ with $h_{i}^{-1}$ yields $\tilde{V}_{i}^j$, a Monte Carlo sample from the conditional distribution of $V_i$, given its parents. Cycling through these transformations for each variable in causal order generates samples from the full joint distribution of the data, as modeled by the cGNF.

\subsection{Estimating Causal Effects with cGNFs}\label{subsec:estimation}

A cGNF enables Monte Carlo sampling not only from the observational joint distribution but also from various interventional distributions. To simulate values from an interventional distribution, Monte Carlo samples are first selected from the standard normal distribution, as before. Next, these samples are transformed using the inverse of the trained cGNF, after truncating it in accordance with the desired intervention. Causal effects are then estimated by averaging and comparing the samples drawn from different interventional distributions. 

This approach to estimating causal effects is an implementation of the g-computation algorithm, first proposed by \citet{robins1986new} and later extended by others \citep{daniel2011gformula, imai2010general, wang2015g}. When implemented in conjunction with a cGNF, the algorithm is versatile enough to estimate any causal effect that is non-parametrically identified from the observed data, without imposing any functional form restrictions on their distribution. 

\subsubsection{Total Effects}\label{subsubsec:ATE_est}

To illustrate, suppose we trained a cGNF based on the DAG in Figure \ref{fig:A-Simple-DAG}, and we were interested in estimating the average total effect of $V_3$ on $V_4$, defined formally as $\mathbf{ATE}_{V_3{\rightarrow}V_4}=\mathbf{E}\left[V_{4}(v_{3}^{*})-V_{4}(v_{3})\right]$. For this estimand, we construct an estimate by Monte Carlo sampling from the interventional distributions that arise after setting $V_3$ at two different values, $v_{3}^*$ and $v_3$, respectively. To this end, we modify the sampling algorithm outlined in the previous section as follows: (i) we skip the step where samples of $V_3$ are generated, and (ii) we draw Monte Carlo samples for all the other variables after setting $V_3$ to $v_{3}^*$ and $v_3$, in turn, wherever this variable appears in the conditioners of the cGNF.

Specifically, to simulate samples from the interventional distribution when $V_3$ is set at $v_{3}^{*}$, we first generate $J$ Monte Carlo samples from the standard normal distribution, each denoted by $\tilde{Z}_{1}^j$ for $j=1,...,J$. These samples are then transformed by computing $\tilde{V}_{1}^j = h_{1}^{-1} \left( \tilde{Z}_{1}^j; c_1; \hat{\boldsymbol{\theta}}_1 \right)$. Next, we create another $J$ Monte Carlo samples from the standard normal distribution, each denoted by $\tilde{Z}_{2}^j$ for $j=1,...,J$, and then transform them by computing $\tilde{V}_{2}^j = h_{2}^{-1} \left( \tilde{Z}_{2}^j; c_2\left( \tilde{V}_{1}^j \right); \hat{\boldsymbol{\theta}}_2 \right)$. These steps mirror the sampling algorithm for the observational joint distribution, as outlined in the previous section, because $V_1$ and $V_2$ causally precede $V_3$, the variable subject to intervention.

At this juncture, however, the procedure diverges from the sampling algorithm outlined previously. After generating Monte Carlo samples for $V_1$ and $V_2$, we now skip drawing samples for $V_3$ and proceed directly to sampling for $V_4$. Here, we generate a final set of $J$ standard normal samples, each denoted by ${Z}_{4}^j$ for $j=1,...,J$. These samples are then transformed by computing $\tilde{V}_{4}^j\left(v_{3}^*\right) = h_{4}^{-1} \left( \tilde{Z}_{4}^j; c_4\left( \tilde{V}_{2}^j, v_{3}^* \right); \hat{\boldsymbol{\theta}}_4 \right)$, where, in the conditioner, $\tilde{V}_{2}^j$ is carried over from the previous step and $V_3$ is set to $v_{3}^*$. The result of this transformation, denoted by $\tilde{V}_{4}^j\left(v_{3}^*\right)$, represents a Monte Carlo sample of $V_4$ from the interventional distribution with $V_3$ set at $v_{3}^*$. 

To simulate samples from the interventional distribution with $V_3$ set at $v_{3}$, rather $v_{3}^*$, we repeat the previous step, only now using this other value for the variable subject to intervention. In particular, we again transform the final set of standard normal samples, denoted as ${Z}_{4}^j$ for $j=1,...,J$, by computing $\tilde{V}_{4}^j\left(v_{3}\right) = h_{4}^{-1} \left( \tilde{Z}_{4}^j; c_4\left( \tilde{V}_{2}^j, v_{3} \right); \hat{\boldsymbol{\theta}}_4 \right)$. In the conditioner of this transformation, $\tilde{V}_{2}^j$ is carried over, as before, while $V_3$ is now set to its alternative value $v_{3}$. As a result, we obtain $\tilde{V}_{4}^j\left(v_{3}\right)$, which represents a Monte Carlo sample of $V_4$ from the interventional distribution with $V_3$ set to $v_3$. 

An estimate for $\mathbf{ATE}_{V_3{\rightarrow}V_4}$, the average total effect of $V_3$ on $V_4$, can then be constructed as follows:
\begin{IEEEeqnarray}{rll}
\widehat{\mathbf{ATE}}_{V_3{\rightarrow}V_4} &\enspace = &\enspace \frac{1}{J} \sum_{j=1}^J \tilde{V_{4}}^{j}(v_{3}^{*}) - \tilde{V_{4}}^{j}(v_{3}), \hfill \IEEEyesnumber \label{seq:MCE_ATE_V3_to_V4_est}
\end{IEEEeqnarray}
where $\tilde{V_{4}}^{j}(v_{3}^{*})$ and $\tilde{V_{4}}^{j}(v_{3})$ denote Monte Carlo samples drawn from interventional distributions with $V_3$ set to $v_{3}^*$ and $v_3$, respectively. A similar procedure can be used to estimate any other total effect, such as $\mathbf{ATE}_{V_2{\rightarrow}V_3}=\mathbf{E}\left[V_{3}(v_{2}^{*})-V_{3}(v_{2})\right]$ or $\mathbf{ATE}_{V_2{\rightarrow}V_4}=\mathbf{E}\left[V_{4}(v_{2}^{*})-V_{4}(v_{2})\right]$, as long as they are non-parametrically identified. For these other estimands, the sampling algorithm would just be modified to reflect the different interventions of interest.

\subsubsection{Conditional Effects}

The sampling algorithm we described for estimating average total effects can also be used to estimate conditional average effects, which reflect the impact of interventions within particular subgroups. For example, suppose we were interested in estimating the conditional average effect of $V_3$ on $V_4$, given $V_2=v_2$, with a cGNF based on the DAG in Figure \ref{fig:A-Simple-DAG}. This estimand can be formally defined as follows:
\begin{IEEEeqnarray}{rll}
\mathbf{CATE}_{V_3{\rightarrow}V_4|v_2} = \mathbf{E} \left[V_{4}(v_{3}^{*}) - V_{4}(v_{3})|V_2=v_2\right]. 
\hfill \IEEEyesnumber \label{seq:CATE_V3_to_V4}
\end{IEEEeqnarray}
It captures the effect of $V_3$ on $V_4$ within the subpopulation for which $V_2=v_2$.

To estimate this effect, we execute the sampling algorithm exactly as outlined previously for the average total effect of $V_3$ on $V_4$. The only difference lies in the final step, where we restrict our comparison of the resulting Monte Carlo samples to those that fall within a particular subgroup. Specifically, we compute the following quantity:
\begin{IEEEeqnarray}{rll}
\widehat{\mathbf{CATE}}_{V_3{\rightarrow}V_4|v_2} &\enspace = &\enspace \frac{1}{J_{v_2}} \sum_{j:\tilde{V}_{2}^j=v_2} \tilde{V_{4}}^{j}(v_{3}^{*}) - \tilde{V_{4}}^{j}(v_{3}).
\hfill \IEEEyesnumber \label{seq:MCE_CATE_V3_to_V4_est}
\end{IEEEeqnarray}
In this expression, the sum is taken over Monte Carlo samples for which $\tilde{V}_{2}^j=v_2$, and $J_{v_2}$ denotes the total number of such samples. The simulated variables $\tilde{V}_{2}^j$, $\tilde{V_{4}}^{j}(v_{3}^{*})$, and $\tilde{V_{4}}^{j}(v_{3})$ are all defined and generated as in Section \ref{subsubsec:ATE_est}. Other conditional effects can be estimated using a similar procedure.\footnote{For example, if $V_2$ were continuous, we could define the conditional effect over an interval of values and then average the Monte Carlo samples of $\tilde{V_{4}}^{j}(v_{3}^{*})$ and $\tilde{V_{4}}^{j}(v_{3})$ among cases with values of $\tilde{V}_{2}^j$ that belong to this interval.}

\subsubsection{Joint Effects}

Joint effects refer to causal estimands that involve interventions on multiple variables simultaneously. For example, the average joint effect of $V_2$ and $V_3$ on $V_4$ can be expressed as follows:
\begin{IEEEeqnarray}{rll}
\mathbf{AJE}_{V_{2},V_{3}{\rightarrow}V_4} = \mathbf{E} \left[V_{4}(v_{2}^{*},v_{3}^{*}) - V_{4}(v_{2},v_{3}) \right].
\hfill \IEEEyesnumber \label{seq:AJE_V2andV3_to_V4}
\end{IEEEeqnarray}
In this equation, $V_4(v_{2}^{*},v_{3}^{*})$ represents the potential outcome of $V_4$ when $V_2$ and $V_3$ are set to $v_{2}^{*}$ and $v_{3}^{*}$, respectively. This potential outcome is contrasted with another, denoted by $V_4(v_{2},v_{3})$, which is defined analogously. The resulting effect captures the combined impact of $V_2$ and $V_3$ on $V_4$.

This effect can also be estimated by Monte Carlo sampling from different interventional distributions, using an inverted and appropriately truncated cGNF. Here, the sampling algorithm is modified to reflect an intervention on multiple variables at once: that is, we skip drawing samples for both $V_2$ and $V_3$, and then we generate samples for the remaining variables after setting $V_2$ and $V_3$ at specific values, wherever they appear in the conditioners of the flow.

Specifically, to simulate samples from an interventional distribution when $V_2$ is set at $v_{2}^{*}$ and $V_3$ is set at $v_{3}^{*}$, we begin by generating Monte Carlo samples for $V_1$, as outlined previously. We then skip drawing samples for both $V_2$ and $V_3$ and proceed directly to sampling for $V_4$. To this end, we generate $J$ Monte Carlo samples from the standard normal distribution, denoted by ${Z}_{4}^j$ for $j=1,...,J$, and we then transform them by computing $\tilde{V}_{4}^j\left(v_{2}^{*}, v_{3}^*\right) = h_{4}^{-1} \left( \tilde{Z}_{4}^j; c_4\left( v_{2}^{*},v_{3}^{*} \right); \hat{\boldsymbol{\theta}}_4 \right)$. The result of this transformation, denoted by $\tilde{V}_{4}^j\left(v_{2}^{*}, v_{3}^{*}\right)$, represents a Monte Carlo sample of $V_4$ from the interventional distribution with $V_2$ and $V_3$ set at $v_{2}^*$ and $v_{3}^*$, respectively. 

To simulate samples from the interventional distribution with $V_2$ and $V_3$ set at $v_{2}$ and $v_{3}$, we repeat the previous calculations, now using these other values for the variables subject to intervention. Thus, we again transform the set of standard normal samples, ${Z}_{4}^j$ for $j=1,...,J$, this time by computing $\tilde{V}_{4}^j\left(v_{2}, v_{3}\right) = h_{4}^{-1} \left( \tilde{Z}_{4}^j; c_4\left( v_{2}, v_{3} \right); \hat{\boldsymbol{\theta}}_4 \right)$, where $V_2$ and $V_3$ are set to their alternative values in the conditioner of the flow. 

An estimate for $\mathbf{AJE}_{V_{2},V_{3}{\rightarrow}V_4}$, the average joint effect of $V_2$ and $V_3$ on $V_4$, can then be constructed as follows:
\begin{IEEEeqnarray}{rll}
\widehat{\mathbf{AJE}}_{V_{2},V_{3}{\rightarrow}V_4} &\enspace = &\enspace \frac{1}{J} \sum_{j=1}^J \tilde{V_{4}}^{j}(v_{2}^{*},v_{3}^{*}) - \tilde{V_{4}}^{j}(v_{2},v_{3}). \hfill \IEEEyesnumber \label{seq:MCE_AJE_V2andV3_to_V4_est}
\end{IEEEeqnarray}
A similar procedure can be used to estimate any other joint effect, including controlled direct effects and interaction effects \citep{vanderweele2009distinctionheterogeneity, vanderweele2015explanation}.

\subsubsection{Mediation Effects}

Monte Carlo sampling, in conjunction with a trained cGNF, can also be used to analyze different types of causal mediation, including direct, indirect, and path-specific effects \citep{pearl2022direct, vanderweele2015explanation, zhou2023tracing}. These effects are all formulated in terms of cross-world potential outcomes, where an exposure variable is set at one value but the mediating variable or variables are set at their values under a different exposure condition. 

For example, consider the natural indirect effect of $V_2$ on $V_4$ operating via $V_3$, given the DAG from Figure \ref{fig:A-Simple-DAG}. This effect can be formally defined as follows:
\begin{IEEEeqnarray}{rll}
\mathbf{NIE}_{V_2{\rightarrow}V_3{\rightarrow}V_4} = \mathbf{E} \left[V_{4}\left(v_{2}^{*}\right)-V_{4}\left(v_{2}^{*},V_{3}\left(v_{2}\right)\right)\right].
\hfill \IEEEyesnumber \label{seq:NDE_V2_to_V4_via_V3}
\end{IEEEeqnarray}
In this expression, $V_{4}\left(v_{2}^*\right)=V_{4}\left(v_{2}^*,V_{3}\left(v_{2}^*\right)\right)$ represents the conventional potential outcome of $V_4$ when $V_2$ is fixed at $v_{2}^*$ and, by extension, $V_3$ takes its natural value under the same exposure setting. Conversely, $V_{4}\left(v_{2}^{*},V_{3}\left(v_{2}\right)\right)$ represents the potential outcome of $V_4$ when $V_2$ is set at $v_{2}^*$ but $V_3$ assumes the value it would naturally take if $V_2$ had been set at $v_2$ instead. This is known as a \textit{cross-world potential outcome}, since it fixes $V_2$ at one value, $v_{2}^*$, while setting $V_3$ at its value from an alternative counterfactual scenario where $V_2$ is instead set at $v_2$. The natural indirect effect captures the influence of $V_2$ on $V_4$ that is transmitted through $V_3$ by comparing the conventional and cross-world potential outcomes. 

To estimate the natural indirect effect, we begin by generating $J$ Monte Carlo samples, each drawn from the standard normal distribution and denoted by $\tilde{Z}_{1}^j$ for $j=1,...,J$. We then transform them to yield samples for $V_{1}$, computing $\tilde{V}_{1}^j = h_{1}^{-1} \left( \tilde{Z}_{1}^j; c_1; \hat{\boldsymbol{\theta}}_1 \right)$ as outlined previously. 

Next, we skip drawing samples for $V_2$ and proceed directly to sampling for $V_3$. Specifically, we generate $J$ Monte Carlo samples from the standard normal distribution, denoted by $\tilde{Z}_{3}^j$ for $j=1,...,J$, and we then transform them by computing $\tilde{V}_{3}^j\left(v_{2}^*\right) = h_{3}^{-1} \left( \tilde{Z}_{3}^j; c_3\left( \tilde{V}_{1}^j, v_{2}^* \right); \hat{\boldsymbol{\theta}}_3 \right)$. In addition, we also transform these samples by computing $\tilde{V}_{3}^j\left(v_{2}\right) = h_{3}^{-1} \left( \tilde{Z}_{3}^j; c_3\left( \tilde{V}_{1}^j, v_{2} \right); \hat{\boldsymbol{\theta}}_3 \right)$. The first of these transformations yields $\tilde{V}_{3}^j\left(v_{2}^{*}\right)$, a Monte Carlo sample of $V_3$ from the interventional distribution with $V_2$ set at $v_{2}^*$, while the second produces $\tilde{V}_{3}^j\left(v_{2}\right)$, a sample from the interventional distribution with $V_2$ now set at $v_2$. 

In the final step, we create Monte Carlo samples for $V_4$. Here, we again initiate the sampling process by generating $J$ Monte Carlo samples from the standard normal distribution, denoted by ${Z}_{4}^j$ for $j=1,...,J$. We then apply two different transformations to these samples. The first transformation computes $\tilde{V}_{4}^j\left(v_{2}^*\right) = h_{4}^{-1} \left( \tilde{Z}_{4}^j; c_3\left( v_{2}^*, \tilde{V}_{3}^j\left(v_{2}^*\right) \right); \hat{\boldsymbol{\theta}}_4 \right)$, where, in the conditioner, $\tilde{V}_{3}^j\left(v_{2}^*\right)$ is carried over from the preceding step. The second transformation computes $\tilde{V}_{4}^j\left(v_{2}^*,V_{3}\left(v_{2}\right)\right) = h_{4}^{-1} \left( \tilde{Z}_{4}^j; c_3\left( v_{2}^*, \tilde{V}_{3}^j\left(v_{2}\right) \right); \hat{\boldsymbol{\theta}}_4 \right)$, in which $\tilde{V}_{3}^j\left(v_{2}\right)$ is carried over from the previous step as well.

Upon completing the sampling algorithm, we obtain obtain Monte Carlo samples of the conventional and cross-world potential outcomes, denoted by $\tilde{V}_{4}^j\left(v_{2}^*\right)$ and $\tilde{V}_{4}^j\left(v_{2}^*,V_{3}\left(v_{2}\right)\right)$, respectively. With these samples, an estimate for the natural indirect effect can then be constructed as follows:
\begin{IEEEeqnarray}{rll}
\widehat{\mathbf{NIE}}_{V_2{\rightarrow}V_3{\rightarrow}V_4} = \frac{1}{J} \sum_{j=1}^J \tilde{V}_{4}^j\left(v_{2}^*\right) - \tilde{V}_{4}^j\left(v_{2}^*,V_{3}\left(v_{2}\right)\right).
\hfill \IEEEyesnumber \label{seq:MCE_NDE_V2_to_V4_via_V3_est}
\end{IEEEeqnarray}
Similar procedures can be used to estimate natural direct effects, pure indirect effects, and path-specific effects. In each case, the sampling algorithm is adapted to simulate the particular conventional and cross-world potential outcomes that compose these other estimands.

\subsection{Sensitivity Analysis}

Identifying and consistently estimating any causal effect hinges on assumptions about the absence of unobserved confounding. Unobserved confounding arises when two variables are influenced by a common cause, termed a confounder. In such cases, an association between these two variables may not reflect a causal effect of one on the other. Instead, it could be a result of spurious co-variation due to their shared connection with an unobserved factor. 

\begin{figure}[!ht]
\begin{centering}
\begin{tikzpicture}[yscale = 3, xscale = 3]

\begin{scope}

	\node[text centered] at (-1,0.5) (Uv1) {$\epsilon_{V_1}$};
	\node[text centered] at (-1,0) (v1) {$V_1$};
	\node[text centered] at (0,-1.15) (Uv2) {$\epsilon_{V_2}$};
	\node[text centered] at (0,-0.65) (v2) {$V_2$};
 	\node[text centered] at (0,0.5) (Uv3) {$\epsilon_{V_3}$};
	\node[text centered] at (0,0) (v3) {$V_3$};
 	\node[text centered] at (1,0.5) (Uv4) {$\epsilon_{V_4}$};
	\node[text centered] at (1,0) (v4) {$V_4$};
	\node[below=1.2cm, scale=0.95, anchor=west] at (-1.085,-1.25) 
		{A. Disturbance Terms are Not Independent};
  
	\draw [->, line width= 1.25] (Uv1) -- (v1);
	\draw [->, line width= 1.25] (Uv2) -- (v2);
	\draw [->, line width= 1.25] (Uv3) -- (v3);
	\draw [->, line width= 1.25] (Uv4) -- (v4);
	\draw [->, line width= 1.25] (v1) -- (v2);
	\draw [->, line width= 1.25] (v1) -- (v3);
	\draw [->, line width= 1.25] (v2) -- (v3);
	\draw [->, line width= 1.25] (v2) -- (v4);
	\draw [->, line width= 1.25] (v3) -- (v4);
    \draw [<->, dashed, line width=1.25] (Uv3) to[out=40, in=140] (Uv4);
    
\end{scope}

\begin{scope}[yshift=0cm, xshift=3cm]

	\node[text centered] at (-1,0.5) (Uv1) {$\epsilon_{V_1}$};
	\node[text centered] at (-1,0) (v1) {$V_1$};
    \node[text=gray, centered] at (-1,1.5) (z1) {$Z_1$};
	\node[text centered] at (0,-1.15) (Uv2) {$\epsilon_{V_2}$};
	\node[text centered] at (0,-0.65) (v2) {$V_2$};
    \node[text=gray, centered] at (-1,-1.15) (z2) {$Z_2$};
 	\node[text centered] at (0,0.5) (Uv3) {$\epsilon_{V_3}$};
	\node[text centered] at (0,0) (v3) {$V_3$};
    \node[text=gray, centered] at (0,1.5) (z3) {$Z_3$};
 	\node[text centered] at (1,0.5) (Uv4) {$\epsilon_{V_4}$};
	\node[text centered] at (1,0) (v4) {$V_4$};
 	\node[text=gray, centered] at (1,1.5) (z4) {$Z_4$};
    \node[text=gray, centered] at (0.5,1.6) (rho) {$\rho_{Z_{3},Z_{4}}$};
    \node[text=gray, rotate=90, centered] at (-0.85,1) (h2) {\scalebox{0.95}{$h_{1}\left(V_{1};c_{1}\right)$}};
    \node[text=gray, centered] at (-0.5,-1.3) (h2) {\scalebox{0.95}{$h_{2}\left(V_{2};c_{2}\left(\mathbf{V}_{2}^p\right)\right)$}};
    \node[text=gray, rotate=90, centered] at (-0.15,1) (h3) {\scalebox{0.95}{$h_{3}\left(V_{3};c_{3}\left(\mathbf{V}_{3}^p\right)\right)$}};
    \node[text=gray, rotate=90, centered] at (1.15,1) (h4) {\scalebox{0.95}{$h_{4}\left(V_{4};c_{4}\left(\mathbf{V}_{4}^p\right)\right)$}};
	\node[below=1.2cm, scale=0.95, anchor=west] at (-1.085,-1.25) 
		{B. Normalized Disturbances are Correlated};
  
	\draw [->, line width= 1.25] (Uv1) -- (v1);
	\draw [->, line width= 1.25] (Uv2) -- (v2);
	\draw [->, line width= 1.25] (Uv3) -- (v3);
	\draw [->, line width= 1.25] (Uv4) -- (v4);
    \draw [gray, ->, line width= 1.25] (Uv1) -- (z1);
	\draw [gray, ->, line width= 1.25] (Uv2) -- (z2);
	\draw [gray, ->, line width= 1.25] (Uv3) -- (z3);
	\draw [gray, ->, line width= 1.25] (Uv4) -- (z4);
	\draw [->, line width= 1.25] (v1) -- (v2);
	\draw [->, line width= 1.25] (v1) -- (v3);
	\draw [->, line width= 1.25] (v2) -- (v3);
	\draw [->, line width= 1.25] (v2) -- (v4);
	\draw [->, line width= 1.25] (v3) -- (v4);
    \draw [<->, dashed, line width=1.25] (Uv3) to[out=40, in=140] (Uv4);
    \draw [gray, <->, dashed, line width=1.25] (z3) to[out=40, in=140] (z4);
    
\end{scope}

\end{tikzpicture}
\caption{Directed Acyclic Graphs (DAGs) Depicting Unobserved Confounding.\label{fig:DAGs-with-Confounding}}
\medskip{}
\par\end{centering}
Note: In these DAGs, the dashed bidirectional arrow connecting $\epsilon_{V_3}$ and $\epsilon_{V_4}$ indicates that these disturbance terms are not independent (i.e., $V_3$ and $V_4$ share an unobserved common cause). The dashed bidirectional arrow connecting $Z_3$ and $Z_4$ denotes that these variables are correlated, where $\rho_{Z_{3},Z_{4}}$ captures the direction and strength of the relationship. The normalizing transformations, denoted by $h_{i}\left(V_{i};c_{i}\left(\mathbf{V}_{i}^p\right)\right)$ for $i=1,...,4$, map each variable $V_i$ to a standard normal variable $Z_i$, conditional on its parents $\mathbf{V}_{i}^p$. Because each variable $V_i$, given its parents $\mathbf{V}_{i}^p$, varies only as a function of its disturbance $\epsilon_{V_i}$, the transformation $h_{i}$ can be also conceptualized as mapping this disturbance term to a standard normal variable $Z_i$. Thus, $Z_3$ and $Z_4$ represent normalized transformations of $\epsilon_{V_3}$ and $\epsilon_{V_4}$, respectively, and $\rho_{Z_{3},Z_{4}}$ captures their correlation. 
\end{figure}
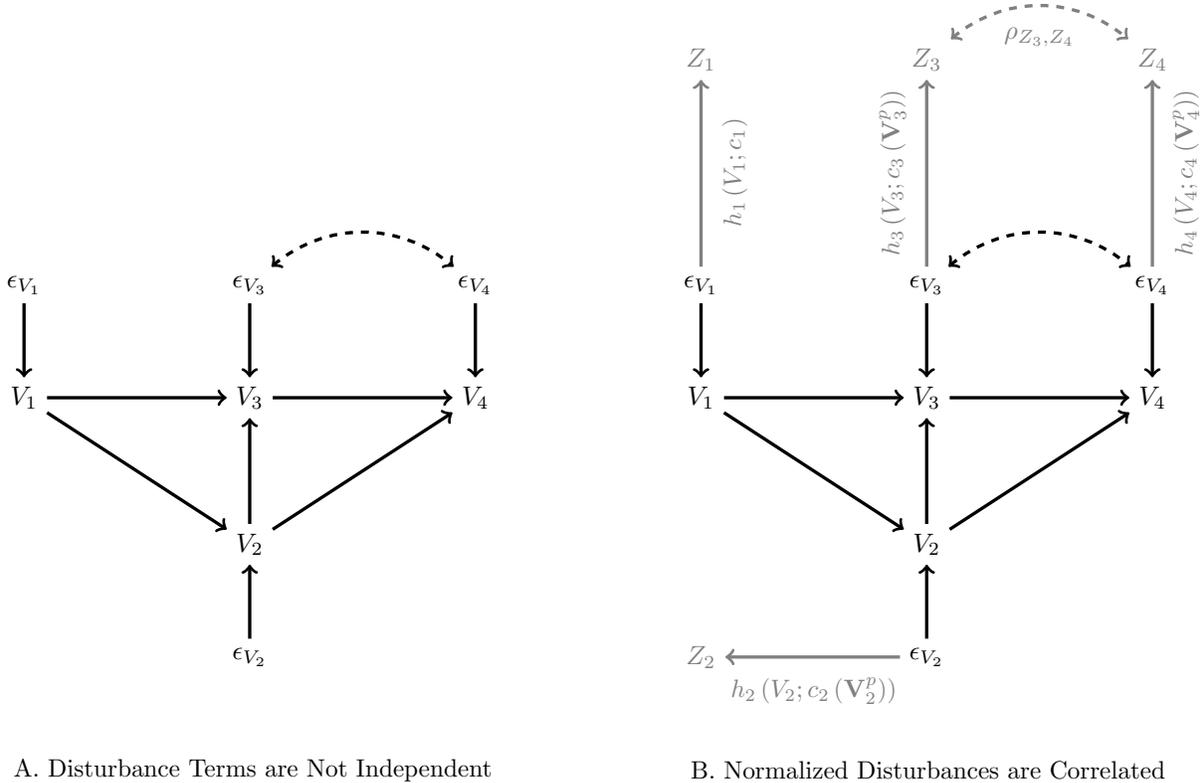

In a DAG, unobserved confounding is often represented by dashed or bidirectional arrows \citep{elwert2013graphical, morgan2015counterfactuals}. For example, Panel A of Figure \ref{fig:DAGs-with-Confounding} uses a dashed, bidirectional arrow to denote that the disturbance terms, $\epsilon_{V_3}$ and $\epsilon_{V_4}$, are not independent because $V_3$ and $V_4$ share an unobserved common cause. In this scenario, certain estimands cannot be identified--specifically, those whose identification relies on the absence of unobserved confounding between $V_3$ and $V_4$, including the $\mathbf{ATE}_{V_3{\rightarrow}V_4}$, $\mathbf{CATE}_{V_3{\rightarrow}V_4|v_2}$, $\mathbf{AJE}_{V_{2},V_{3}{\rightarrow}V_4}$, and $\mathbf{NIE}_{V_2{\rightarrow}V_3{\rightarrow}V_4}$, among others. As a result, cGNFs may fail to provide accurate estimates for these effects. 

Because the absence of unobserved confounding is neither empirically verifiable nor automatically ensured by common research designs in the social sciences, it is important to assess the sensitivity of effect estimates to potential bias. cGNFs offer a straightforward and intuitive approach for conducting such analyses \citep{balgi2022rho}. It involves specifying a set of sensitivity parameters, which represent correlations between normalized disturbance terms in the assumed DAG. These correlations are then used to modify the processes through which a cGNF is trained and utilized to generate Monte Carlo samples. By modifying the training and sampling procedures in this way, the cGNF is recalibrated to adjust for possible biases due to unobserved confounding.

To better appreciate this approach, note that mapping any variable $V_i$, conditional on its parents $\mathbf{V}_{i}^p$, to a standard normal variable $Z_i$ is functionally equivalent to mapping its disturbance term $\epsilon_{V_i}$ to the same variable $Z_i$. This is because the only source of variation in $V_i$, holding its parents constant, comes from its disturbance term. The equivalence of these mappings is represented graphically in Panel B of Figure \ref{fig:DAGs-with-Confounding}, where each variable $Z_i$ is depicted as a transformation of a corresponding disturbance term $\epsilon_{V_i}$.

In analyses devoid of unobserved confounding, where all the disturbance terms are pairwise independent, a cGNF should transform them into a set of mutually independent, standard normal variables. In other words, when unobserved confounding is assumed away entirely, a cGNFs is designed to map the disturbance terms to a new set of variables distributed as $\mathcal{N}(\mathbf{0},\mathbf{I})$. However, if the disturbance terms are not independent due to unobserved confounding, a cGNF can be modified to map them instead to a multivariate normal distribution that preserves their dependence structure.

Specifically, to accommodate dependent disturbances, a cGNF can be reformulated as follows: 
\begin{IEEEeqnarray}{rll}
\mathbf{Z} = \boldsymbol{h}(\mathbf{V};\boldsymbol{\theta}) \sim \mathcal{N}(\mathbf{0},\Sigma_\mathbf{Z}). \hfill \IEEEyesnumber
\label{seq:multi_norm_flow_rhognf}
\end{IEEEeqnarray}
In this expression, the normalizers now map the set of variables $\mathbf{V}=\{V_1,...,V_k\}$ into a new set $\mathbf{Z}=\{Z_1,...,Z_k\}$, which are distributed according to a multivariate normal distribution with a mean vector of zeros and a covariance matrix given by $\Sigma_\mathbf{Z}$ rather than the identify matrix $\mathbf{I}$. 

This distribution can be represented in greater detail as follows: 
\begin{IEEEeqnarray}{C}
\mathcal{N}(\mathbf{0},\Sigma_\mathbf{Z}) = \mathcal{N}\left(\begin{bmatrix}
0 \\
\vdots \\
0
\end{bmatrix},\begin{bmatrix}
1 & \rho_{Z_1,Z_2} & \cdots & \rho_{Z_1,Z_k} \\
\rho_{Z_1,Z_2} & \ddots & \ddots & \vdots \\
\vdots & \ddots & \ddots & \vdots \\
\rho_{Z_1,Z_k} & \cdots & \cdots & 1
\end{bmatrix}\right), \enspace\IEEEyesnumber \label{seq:normal_z_sigma}
\end{IEEEeqnarray}
where the off-diagonal elements of $\Sigma_\mathbf{Z}$ are a set of Pearson correlations. These correlations reflect the direction and magnitude of the relationships between different disturbance terms, after each has been transformed such that its marginal distribution is univariate standard normal. Stronger correlations between these normalized disturbances correspond to a greater degree of unobserved confounding.

To adapt a cGNF for mapping the set of variables $\mathbf{V}$ into a new set $\mathbf{Z}$ that is distributed as $\mathcal{N}(\mathbf{0},\mathbf{\Sigma_Z})$ rather than $\mathcal{N}(\mathbf{0},\mathbf{I})$, we need only modify the loss function used to optimize the weights of the UMNNs during training. This modified loss function can be expressed as follows:
\begin{IEEEeqnarray}{rll}
-\mathcal{LL}\left(\boldsymbol{\theta;\Sigma_\mathbf{Z}}\right) &\enspace = &\enspace -\ln\left(\prod_{l=1}^{n}f_{\mathbf{V}}\left(\mathbf{v}^l;\boldsymbol{\theta}\right)\right) \hfill \nonumber \\
&\enspace = &\enspace -\ln\left(\prod_{l=1}^{n}f_{\mathbf{Z}}\left(\boldsymbol{h}\left(\mathbf{v}^l;\boldsymbol{\theta}\right);\Sigma_\mathbf{Z}\right)\left|\mathrm{det}\mathbb{J}_{\boldsymbol{h}(\mathbf{v}^l;\boldsymbol{\theta})}\right|\right) \hfill \nonumber \\
&\enspace = &\enspace -\sum_{l=1}^{n}\ln\left(f_{\mathbf{Z}}\left(\boldsymbol{h}\left(\mathbf{v}^l;\boldsymbol{\theta}\right);\Sigma_\mathbf{Z}\right)\right) - \sum_{l=1}^{n}\sum_{i=1}^{k}\ln\left(\left|\frac{\partial h_{i}}{\partial v_{i}^l}\right|\right), \hfill \IEEEyesnumber 
\label{seq:neg_LL_loss_rhognf}
\end{IEEEeqnarray}
where $l=1,...,n$ indexes observations in the sample data, $i=1,...,k$ indexes the variables, and $f_{\mathbf{Z}}(\mathbf{z}^l;\Sigma_\mathbf{Z})$ represents the multivariate normal distribution with a mean vector of zeros and a covariance matrix $\Sigma_\mathbf{Z}$, evaluated at $\boldsymbol{h}\left(\mathbf{v}^l;\boldsymbol{\theta}\right)$. The only difference between this loss function and the one used to train a standard cGNF lies in the substitution of the covariance matrix $\Sigma_\mathbf{Z}$ for the identity matrix $\mathbf{I}$ in the multivariate normal distribution $f_{\mathbf{Z}}$.

After training a cGNF with this modified loss function, the Monte Carlo sampling algorithm must also be adapted to properly generate effect estimates adjusted for confounding bias. To modify the sampling algorithm, we simply draw our initial $J$ Monte Carlo samples of $\tilde{Z}_{i}^j$, for $j=1,...,J$ and $i=1,...,k$, from the multivariate normal distribution given by $\mathcal{N}(\mathbf{0},\mathbf{\Sigma_Z})$. These samples are then transformed using the inverse of a cGNF that has been optimized with the loss function from Equation \eqref{seq:neg_LL_loss_rhognf}. Otherwise, the sampling algorithm proceeds exactly as outlined in the previous section.

Effect estimates generated by this modified training and sampling procedure are adjusted for confounding bias, as represented through the correlations between normalized disturbance terms in $\mathbf{\Sigma_Z}$. Their sensitivity to different forms of unobserved confounding can then be assessed by training and computing estimates from multiple cGNFs, using a range of values for these correlations.

To illustrate, consider the goal of estimating the average total effect of $V_3$ on $V_4$ in the presence of unobserved confounding, as depicted in Figure \ref{fig:DAGs-with-Confounding}. To construct bias-adjusted estimates in this scenario, we first train a cGNF with the following form: 
\begin{IEEEeqnarray}{C}
\mathbf{Z} = \boldsymbol{h}(\mathbf{V};\boldsymbol{\theta}) = \begin{bmatrix}
Z_1 \\
Z_2 \\
Z_3 \\
Z_4
\end{bmatrix}\sim \mathcal{N}\left(\begin{bmatrix}
0 \\
0 \\
0 \\
0
\end{bmatrix},\begin{bmatrix}
1 & 0 & 0 & 0 \\
0 & 1 & 0 & 0 \\
0 & 0 & 1 & \rho_{Z_3,Z_4} \\
0 & 0& \rho_{Z_3,Z_4} & 1
\end{bmatrix}\right)=\mathcal{N}(\mathbf{0}, \Sigma_\mathbf{Z}). \enspace \IEEEyesnumber \label{seq:normal_z_sigma_V3_V4_corr}
\end{IEEEeqnarray}
In this model, $Z_3$ and $Z_4$ can be interpreted as normalized transformations of the disturbance terms $\epsilon_{V_3}$ and $\epsilon_{V_4}$, respectively, with $\rho_{Z_{3},Z_{4}}$ denoting their correlation on this transformed scale. By specifying a range of values for $\rho_{Z_{3},Z_{4}}$ and training different cGNFs using a modified loss function, we could then obtain a corresponding range of effect estimates adjusted for different types of unobserved confounding. To generate these estimates, we implement the sampling algorithm for the $\mathbf{ATE}_{V_3{\rightarrow}V_4}$ exactly as described in Section \ref{subsubsec:ATE_est}, but with two modifications: our initial samples of $\tilde{Z}_{i}^j$ are drawn from $\mathcal{N}(\mathbf{0}, \Sigma_\mathbf{Z})$, and then they are transformed using the inverse of a cGNF optimized with the loss function from Equation \eqref{seq:neg_LL_loss_rhognf}.

In sum, cGNFs seamlessly integrate with methods of sensitivity analysis for unobserved confounding. This is accomplished by modifying the training and sampling process to reflect a prescribed set of correlations among normalized versions of the disturbance terms. This approach to sensitivity analysis is extremely versatile. It enables construction of bias-adjusted estimates for a wide range of effects across many different types of unobserved confounding, simply by varying the correlation structure among the transformed disturbances. 

\subsection{Summary}

A cGNF arranges a set of variables in causal order and then maps each one to the standard normal distribution, conditional on its parents, as defined by an assumed DAG. This mapping is accomplished with UMNNs--a special class of artificial neural networks that can approximate any monotonic transformation--trained on sample data by the method of SGD. Because UMNNs are invertible, a trained cGNF enables Monte Carlo sampling from the observational joint distribution of the data by first drawing samples from the standard normal distribution and then recursively applying the inverse of the flow. Moreover, this sampling procedure can be selectively truncated to facilitate Monte Carlo sampling from many different interventional distributions, with the resulting samples used to construct estimates for a wide range of causal effects. All these procedures can be  modified to assess robustness to different types of unobserved confounding.

The workflow for implementing an analysis based on cGNFs can thus be summarized as follows:

\begin{enumerate}
\item Draw a DAG, using theory and prior knowledge.
\item Determine which estimands can be identified from the observed data, given the DAG.
\item Train a cGNF on the observed data.
    \begin{enumerate}
    \item Specify the UMNN architecture.
    \item Specify the SGD hyperparameters.
    \item Execute the training algorithm.
    \end{enumerate}
\item Use the cGNF for Monte Carlo sampling from the relevant interventional distributions.
\item Construct estimates for the target estimands using the Monte Carlo samples.
\item Assess the sensitivity of these estimates to unobserved confounding.
\end{enumerate}

The accuracy of the effect estimates produced by this procedure hinges crucially on whether the assumed DAG is correct. If the assumed DAG is incorrect, estimates from the cGNF will be biased. The accuracy of these estimates also depends on the expressiveness of the neural networks used to model the cGNF. If the UMNNs are not sufficiently expressive, they may not provide an accurate approximation for the true but unknown joint distribution. In this situation, cGNFs may also produce biased estimates, even when the assumed DAG is accurate. 

Beyond the accuracy of the DAG and the expressiveness of the networks, the performance of cGNFs also depends on the amount of sample data available to train them, whether the training algorithm converges to an optimal solution from its random initialization, and the number of Monte Carlo samples generated from the trained model for estimation. To quantify the uncertainty in estimates due to sampling error, training error, and simulation error taken together, we recommend constructing confidence intervals using the percentiles of a synthetic sampling distribution generated via the non-parametric bootstrap \citep{tibshirani1993introduction}. Although computationally demanding to implement, the bootstrap can quantify uncertainty in a wide variety of predictions from artificial neural networks with a high degree of accuracy \citep{franke2000bootstrapping, heskes1996practical}. 

\section{Empirical Illustrations}

We illustrate the utility of cGNFs by revisiting two seminal studies of social mobility. The first is Blau and Duncan's (\citeyear{blau1967american}) classic analysis of status attainment, which pioneered the study of mobility processes using parametric structural equation models (SEMs). The second, Zhou's (\citeyear{zhou2019selection}) recent analysis of conditional versus controlled mobility, leveraged a directed acyclic graph (DAG) to guide semi-parametric estimation of several specific estimands in a broader causal system.

\subsection{Reanalysis of Blau and Duncan (1967)}

In \textit{The American Occupational Structure}, Blau and Duncan \citeyearpar{blau1967american} introduced a groundbreaking approach to studying inter-generational social mobility in the United States. Specifically, they employed linear path analysis--a form of parametric structural equation modeling--to assess the influence of family background, education, and early career achievements on later occupational attainment. Drawing from the 1962 ``Occupational Changes in a Generation'' (OCG) survey, the study analyzed data from about 20,000 American men between the ages of 20 and 64, offering a comprehensive view of the adult male workforce at the time.

The study centered its analysis on two main variables: educational attainment and occupational status, relating these measures across generations (i.e., between fathers and sons). In the OCG survey, educational attainment was originally measured in years of formal schooling, which was then grouped into nine distinct categories, ranging from ``no formal education'' to ``postgraduate studies.'' Occupational status was measured using the Duncan Socioeconomic Index (SEI), a scale that assigned an estimated prestige score to each occupational category in the OCG survey, ranging from 0 to 96. This score was calculated by weighting and combining the average income and educational levels of different occupations, with weights based on the relation of these characteristics to a separate set of occupational prestige ratings.

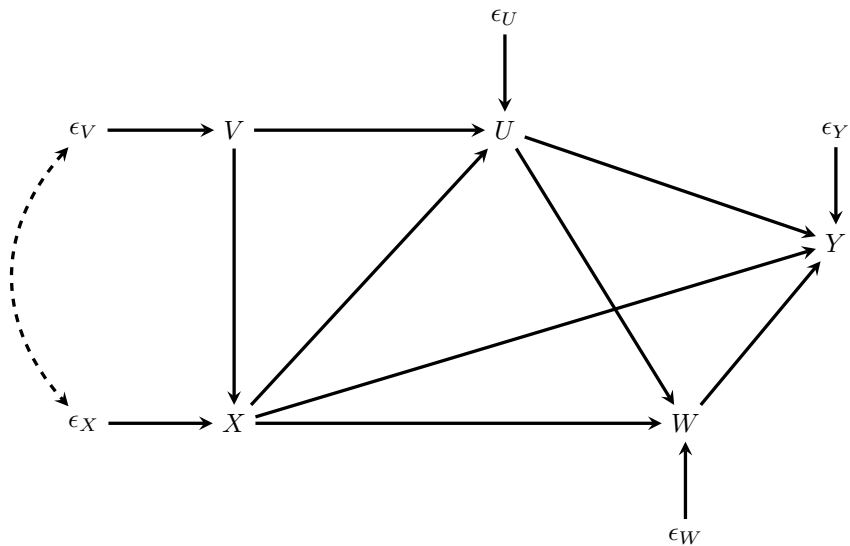
\begin{figure}[!ht]
\begin{centering}
\begin{tikzpicture}[yscale=3, xscale=4, >=stealth]

    \node[text centered] at (-1.5,0.65) (e_v) {$\epsilon_V$};
	\node[text centered] at (-1.0,0.65) (v) {$V$};
     \node[text centered] at (-1.5,-0.65) (e_x) {$\epsilon_X$};
	\node[text centered] at (-1.0,-0.65) (x) {$X$};
    \node[text centered] at (-0.1,1.15) (e_u) {$\epsilon_U$};
 	\node[text centered] at (-0.1,0.65) (u) {$U$};
 	\node[text centered] at (0.5,-1.15) (e_w) {$\epsilon_W$};
	\node[text centered] at (0.5,-0.65) (w) {$W$};
    \node[text centered] at (1,0.65) (e_y) {$\epsilon_Y$};
	\node[text centered] at (1,0.15) (y) {$Y$};
	
  	\draw [->, line width=1.25] (e_v) -- (v);
 	\draw [->, line width=1.25] (e_x) -- (x);
	\draw [->, line width=1.25] (e_u) -- (u);
 	\draw [->, line width=1.25] (e_w) -- (w);
  	\draw [->, line width=1.25] (e_y) -- (y);
	\draw [->, line width=1.25] (v) -- (u);
    \draw [->, line width=1.25] (v) -- (x);
    \draw [->, line width=1.25] (x) -- (u);
	\draw [->, line width=1.25] (x) -- (w);
    \draw [->, line width=1.25] (x) -- (y);
	\draw [->, line width=1.25] (u) -- (w);
	\draw [->, line width=1.25] (u) -- (y);
	\draw [->, line width=1.25] (w) -- (y);
    \draw [<->, dashed, line width=1.25] (e_v) to[out=-125, in=125] (e_x);
 
\end{tikzpicture}
\caption{A DAG corresponding to Blau and Duncan's \citeyearpar{blau1967american} Linear Path Model.}
\label{fig: Blau & Duncan DAG}
\medskip{}
\par\end{centering}
Note: $V$ denotes father's educational attainment, $X$ denotes father's occupational status, $U$ denotes son's educational attainment, $W$ denotes the occupational status of the son's first job, and $Y$ denotes the occupational status of the son's job in 1962. The dashed, bidirectional arrow connecting $\epsilon_V$ and $\epsilon_X$ indicates that these disturbances are correlated (i.e., $V$ and $X$ share unobserved common causes).
\end{figure}

Using these data, \citet{blau1967american} fit a linear path model that included five variables: father's educational attainment ($V$), father's occupational status ($X$), son's educational attainment ($U$), the occupational status of the son's first job ($W$), and the occupational status of the son's job in 1962 ($Y$), when the OCG survey was fielded.\footnote{We adopt the same notation for these variables as in \citet{blau1967american}.} The hypothesized relationships among them are depicted in Figure \ref{fig: Blau & Duncan DAG}, which contains a DAG corresponding to Blau and Duncan's \citeyearpar{blau1967american} original model. In this graph, $V$ affects $X$ and $U$; $X$ affects $U$, $W$, and $Y$; $U$ affects $W$ and $Y$; and $W$ affects $Y$. The dashed, bidirectional arrow connecting the disturbance terms for $V$ and $X$ indicates that these variables share unobserved common causes (e.g., grandfather's occupational status). The DAG differs from Blau and Duncan's \citeyearpar{blau1967american} linear path model in that it does not impose any restrictions on the functional form of these relationships.

In our reanalysis, we employ a cGNF based on the DAG in Figure \ref{fig: Blau & Duncan DAG}, training it on the OCG data. Our model architecture includes an embedding network with five hidden layers, each composed of 100, 90, 80, 70, and 60 nodes, in succession. Similarly, the integrand network also features five hidden layers with 60, 50, 40, 30, and 20 nodes, respectively. We train the cGNF using stochastic gradient descent (SGD) to minimize the negative log-likelihood, adopting a batch size of 128 and a learning rate of 0.0001. The training process is halted when there has been no reduction in the negative log-likelihood for 50 consecutive epochs, as computed on a one-fifth validation sample held out from the OCG data. 

After training the cGNF, we use it to estimate several different effects of interest, including the average total effect of father's occupational status ($X$) on son's occupational status ($Y$), the average total effect of son's education ($U$) on son's occupational status ($Y$), and the natural direct and indirect effects of father's occupational status ($X$) on son's occupational status ($Y$), as mediated by son's education ($U$). To quantify the uncertainty in these estimates, we construct 90 percent confidence intervals using the non-parametric bootstrap with 600 replications. This involves repeatedly retraining the cGNF and recalculating effect estimates on random samples from the OCG data, drawn with replacement. The confidence intervals are then given by the 95th and 5th percentiles of the resulting bootstrap distribution. Replication files for this analysis are available at \url{https://github.com/gtwodtke/deep_learning_with_DAGs/tree/main/blau_duncan_1967}.

\begin{figure}[ht!]
\centering
\includegraphics[width=\linewidth,height=\linewidth,keepaspectratio]{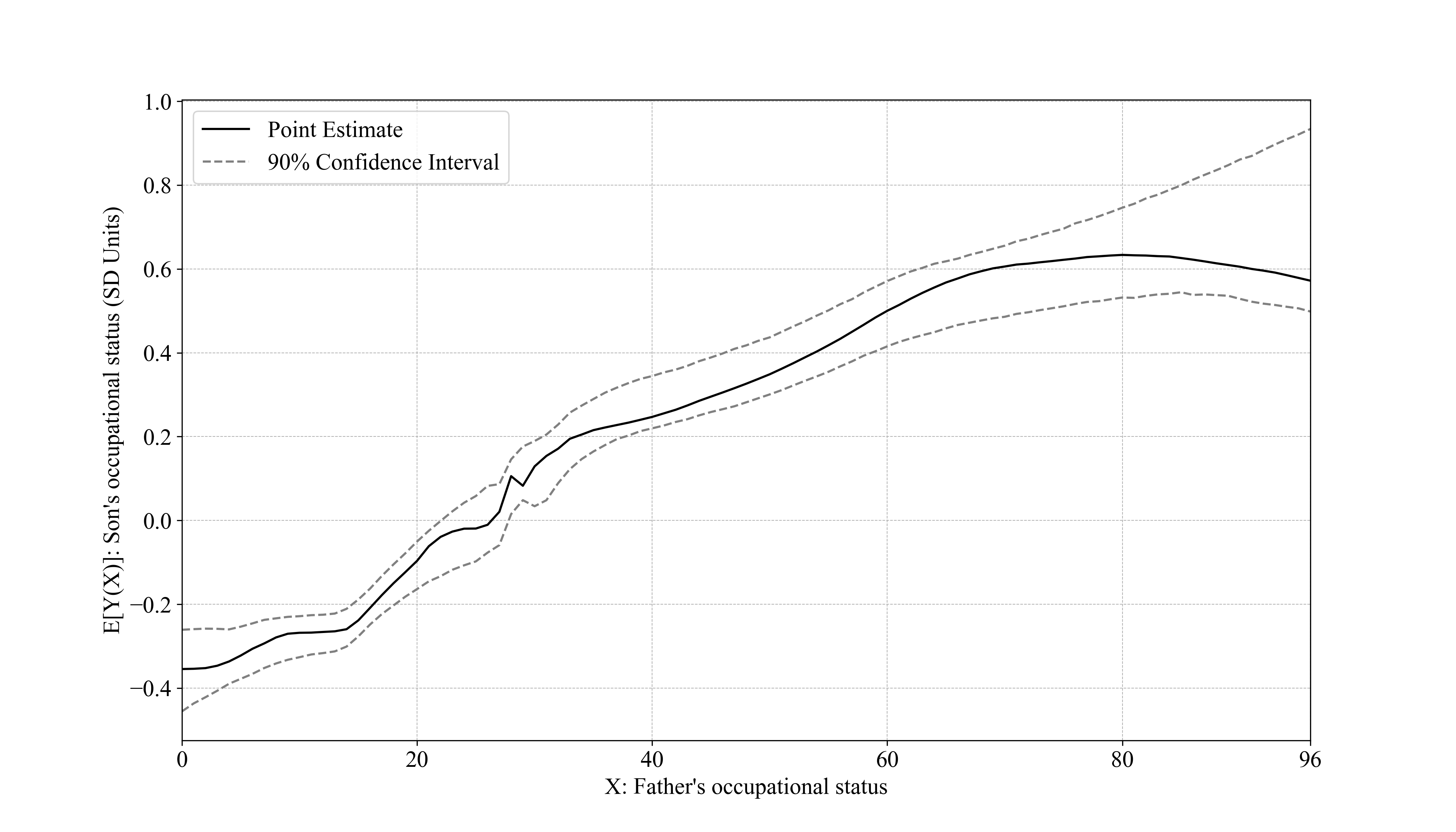}
\caption{Effect of $X$ (Father's Occupational Status) on $Y$ (Son's Occupational Status).}
\label{fig:est_ATE_X_Y}
\end{figure}

Figure \ref{fig:est_ATE_X_Y} presents an estimated response function describing the average total effect of father's occupational status on son's occupational status. Overall, these results are consistent with Blau and Duncan's \citeyearpar{blau1967american} linear path model, which showed a strong positive relationship between between occupational attainment across generations. However, estimates from the cGNF also reveal some non-linearity in this relationship, particularly at the upper and lower ends of the status spectrum among fathers.

\begin{figure}[ht!]
\centering
\includegraphics[width=\linewidth,height=\linewidth,keepaspectratio]{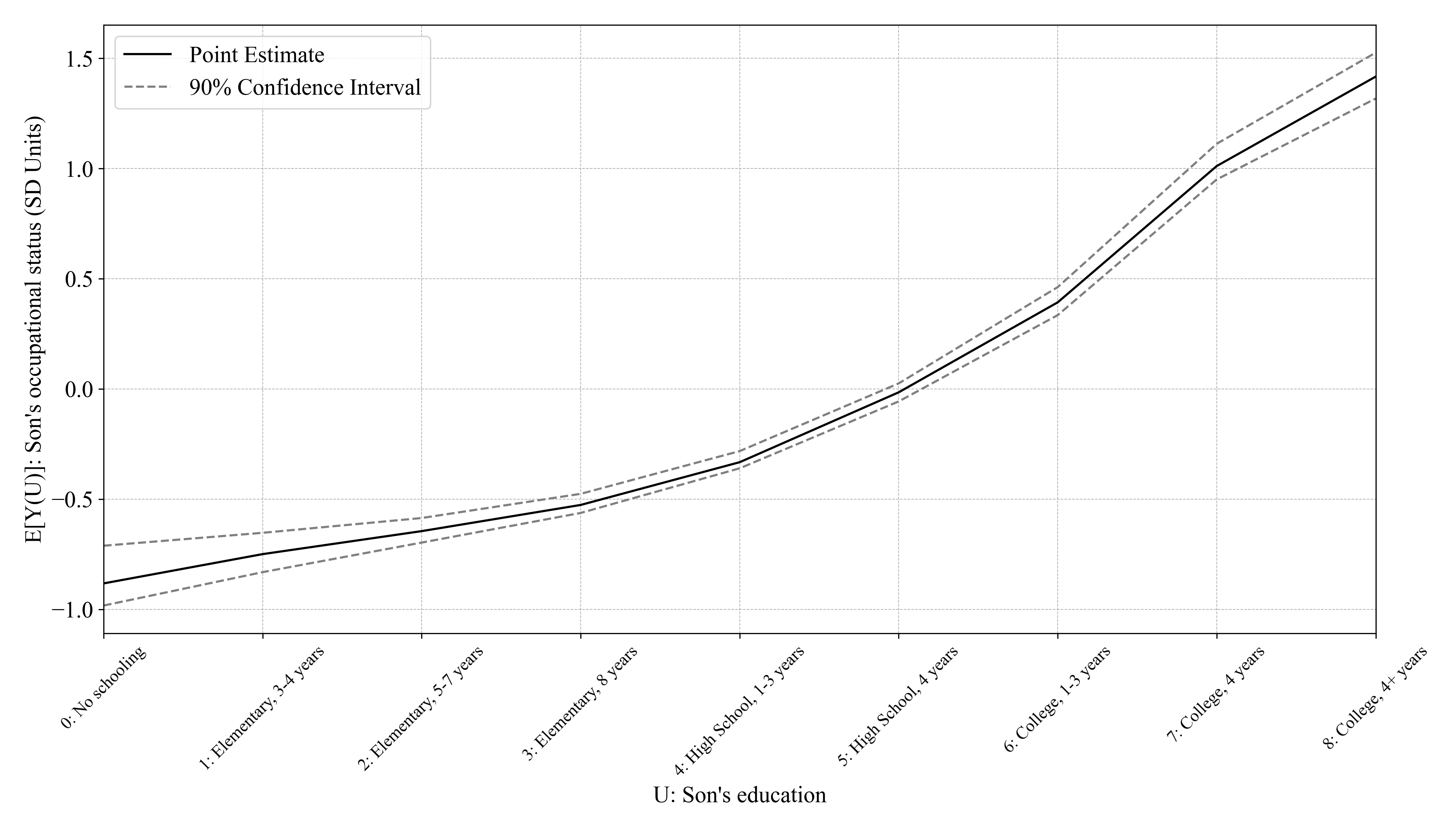}
\caption{Effect of $U$ (Son's Education) on $Y$ (Son's Occupational Status).}
\label{fig:est_ATE_U_Y}
\end{figure}

In Figure \ref{fig:est_ATE_U_Y}, we present a response function that summarizes the average total effect of a son's education on his subsequent occupational status. These results show a strong positive relationship between education and occupational attainment, aligning with Blau and Duncan \citeyearpar{blau1967american}. As before, estimates from the cGNF also uncover considerable non-linearity, where differences in educational attainment up to high school exert a comparatively modest impact compared to those at the secondary level and beyond.

\begin{figure}[ht!]
\begin{centering}
\includegraphics[width=\linewidth,height=\linewidth,keepaspectratio]{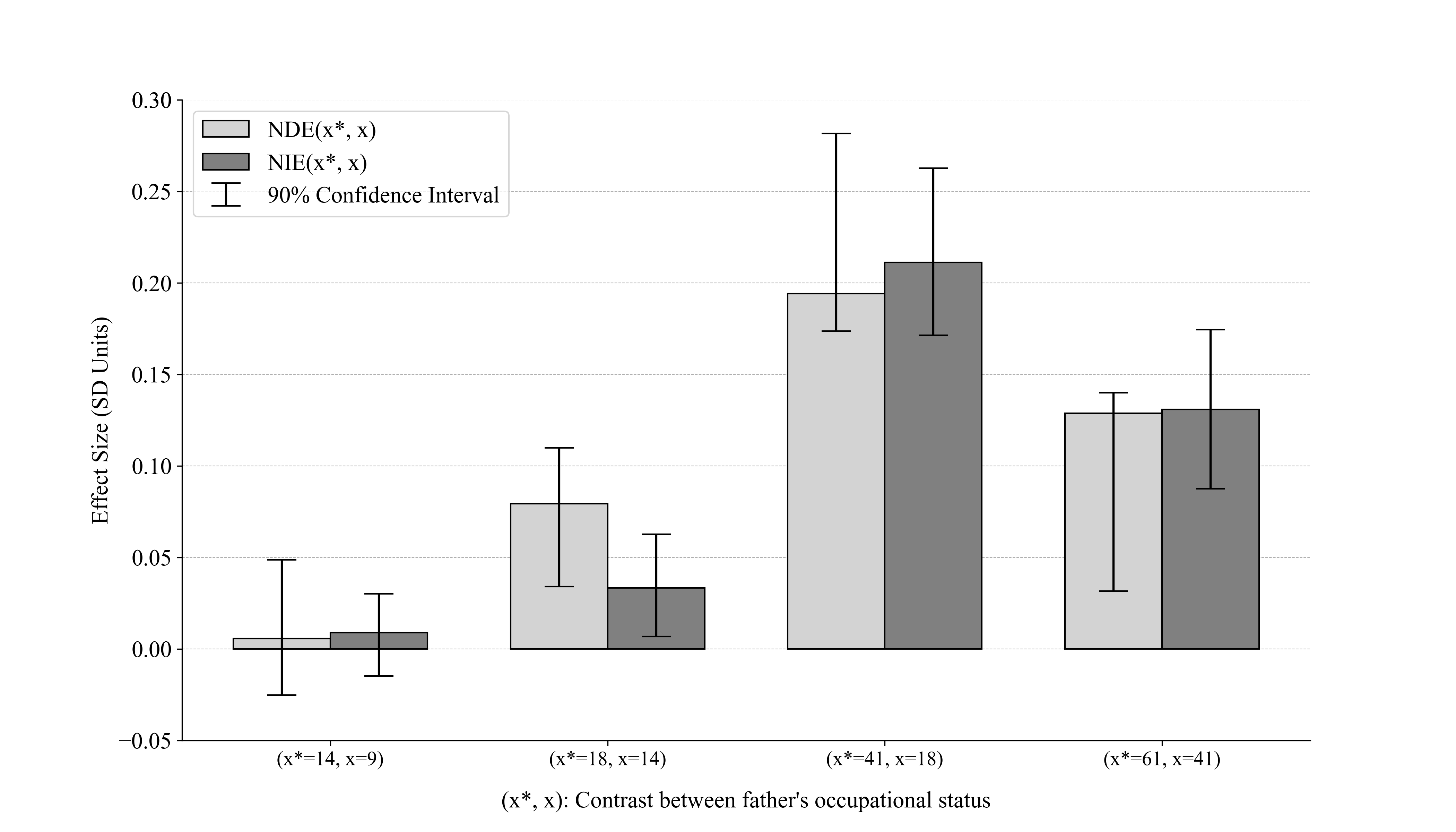}
\caption{Natural direct and indirect effects of $X$ (Father's Occupational Status) on $Y$ (Son's Occupational Status), as mediated via $U$ (Son's Education)}
\label{fig:est_NDE_NIE_X_Y_via_U}
\medskip{}
\par\end{centering}
Note: The natural direct and indirect effects are here based on contrasts between adjacent quintiles of the sample distribution for father's occupational status. The first through fifth quintiles of this variable are 9, 14, 18, 41, and 61, respectively.
\end{figure}

Figure \ref{fig:est_NDE_NIE_X_Y_via_U} presents the natural direct and indirect effects of father’s occupational status on son’s occupational status, as mediated by son’s education. These effects are based on contrasts between adjacent quintiles of the status distribution among fathers, and they are consistent with the linear path model of Blau and Duncan \citeyearpar{blau1967american}, which demonstrated a sizeable mediating role for educational attainment. Nevertheless, estimates from the cGNF suggest that education does not seem to play a very important mediating role at the lower end of the status distribution among fathers. In contrast, at the higher end, education is a powerful mediator, as indicated by the large indirect effects.

All these estimates are predicated on the critical assumption that there is no unobserved confounding of the focal relationships. If, for example, factors like motivation or prior achievement in high school, which are not observed in the OCG data, confound the relationship between a son's education and occupational status, then our estimates involving the effects of $U$ on $Y$ would be biased.

\begin{figure}[ht!]
\centering
\includegraphics[width=\linewidth,height=\linewidth,keepaspectratio]{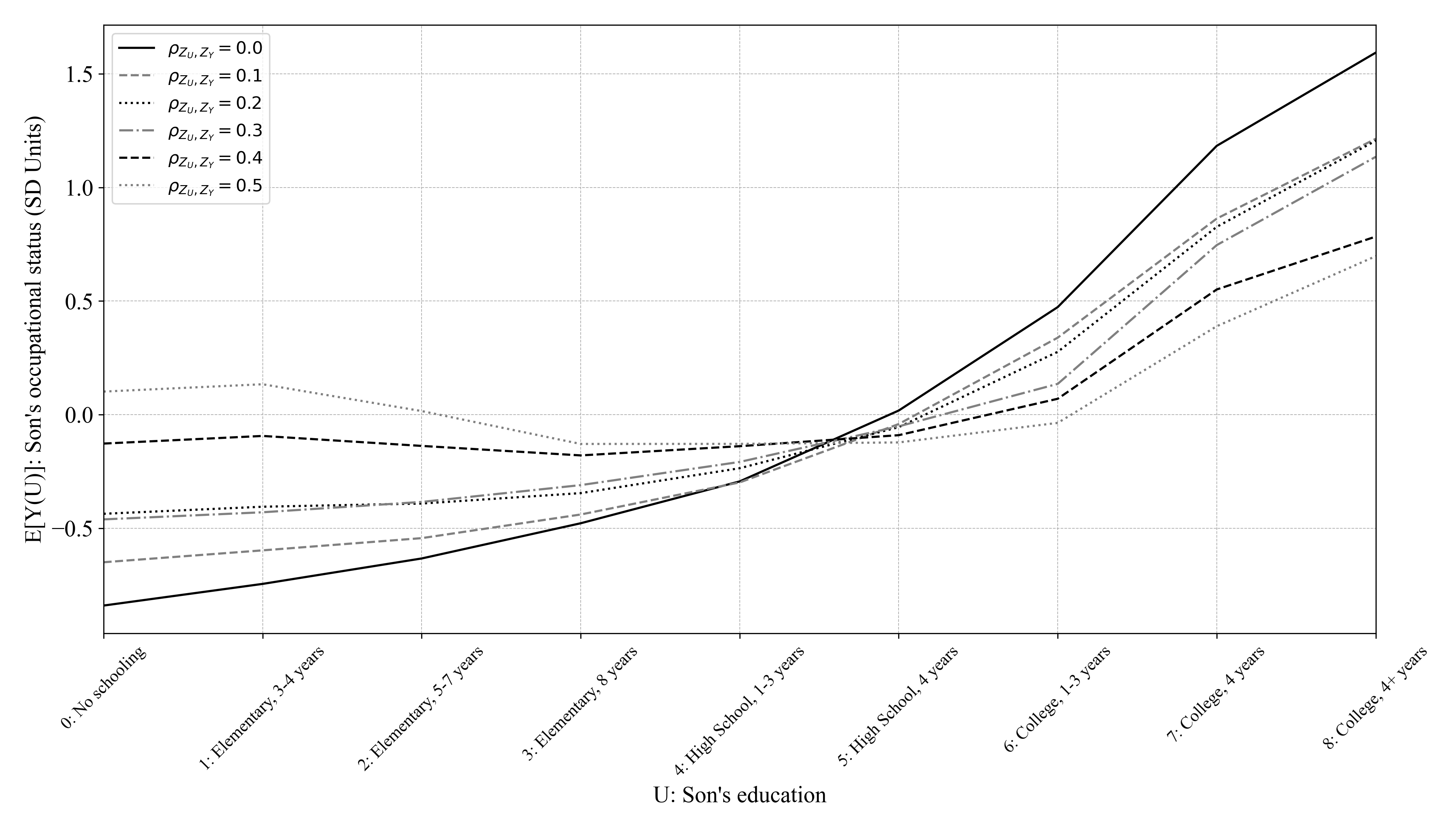}
\caption{Sensitivity of the Effect of $U$ (Son's Education) on $Y$ (Son's Occupational Status) to Unobserved $U$-$Y$ Confounding.}
\label{fig:est_ATE_U_Y_sens}
\end{figure}

Figure \ref{fig:est_ATE_U_Y_sens} presents results from a sensitivity analysis assessing the robustness of our findings to hypothetical patterns of unobserved confounding. It displays bias-adjusted estimates for the effect of son's education on their occupational status, plotted across values of the sensitivity parameter $\rho_{Z_{U},Z_{Y}}$. Positive values for this parameter imply that individuals select into higher levels of education on the basis of unobserved factors that also lead them to attain higher-status occupations, with larger values signaling stronger forms of selection. 

The results show that, across a range of plausible values for $\rho_{Z_{U},Z_{Y}}$, the estimated response function remains relatively stable. The response function begins to flatten out only at large values for $\rho_{Z_{U},Z_{Y}}$, but even under this rather extreme level of unobserved confounding, post-secondary education continues to exert a strong effect on occupational attainment. This stability suggests that our inferences about the causal relationship between $U$ and $Y$ are likely robust to unobserved confounding. 

In sum, our reanalysis indicates that the status attainment process involves nontrivial departures from linearity, a pattern that traditional path models fail to capture. By utilizing a cGNF, we circumvent the restrictive assumptions of standard SEMs, gain the ability to uncover and describe more complex relationships, and enable a seamless assessment of how these relationships vary under different forms of unobserved confounding.

\subsection{Reanalysis of Zhou (2019)}

\citet{zhou2019selection} examined the influence of higher education on inter-generational income mobility in the United States. The study showed that the relationship between the incomes of parents and their children is weaker for college graduates than for those with less education. However, this observed pattern could stem from non-random selection into post-secondary education. The process through which children select into college is complex, as it involves factors like academic performance in high school, which not only confound the effects of college among children but are also influenced by parental income. To navigate this complicated selection process, \citet{zhou2019selection} used a DAG to guide the design of a semi-parametric estimator capable of accurately evaluating whether a college degree actually promotes greater mobility.

The study analyzed data from 4,673 respondents in the 1979 National Longitudinal Survey of Youth (NLSY), all under 19 years old at baseline. The analysis centered on the following variables: the income rank of parents when the respondent was in high school ($X$), the income rank of the respondent as an adult ($Y$), and whether the respondent graduated from college ($C$). A set of baseline controls ($B$) was also incorporated into the analysis. These include measures of gender, race, urban residence, family structure, and parental education. Furthermore, the analysis included two additional variables: the respondent's educational expectations and scores on the Armed Forces Qualification Test (AFQT), measured during their high school years. These variables, denoted by $L$ and $A$, respectively, may be influenced by parental income and may also confound the relationship of education to income among respondents.

\begin{figure}[!ht]
\vspace{-20mm}
\begin{centering}
\begin{tikzpicture}[yscale=1.5, xscale=2, >=stealth]

    \node[text centered] at (0,0) (b) {$B$};
    \node[text centered] at (2,1) (l) {$L$};
    \node[text centered] at (1.5,-1) (x) {$X$};
    \node[text centered] at (3,0) (a) {$A$};
    \node[text centered] at (3.5,-1) (c) {$C$};
    \node[text centered] at (4.5,0) (y) {$Y$};
    
    \draw [->, line width=1.25] (b) to (l);
    \draw [->, line width=1.25] (b) to (x);
    \draw [->, line width=1.25] (b) to (a);
    \draw [->, line width=1.25] (b) .. controls +(1.5, -2) and +(+60: -1) .. (c);
    \draw [->, line width=1.25] (b) .. controls +(4, 3.5) and +(-100: -0.5) .. (y);
    \draw [->, line width=1.25] (x) to [bend left=30] (l);
    \draw [->, line width=1.25] (x) to (a);
    \draw [->, line width=1.25] (x) to (c);
    \draw [->, line width=1.25] (x) .. controls +(3.5, -1.2) and +(+100: -0.5) .. (y);
    \draw [->, line width=1.25] (l) to (a);
    \draw [->, line width=1.25] (l) to [bend left=45] (c);
    \draw [->, line width=1.25] (l) to [bend left=30] (y);
    \draw [->, line width=1.25] (a) to (c);
    \draw [->, line width=1.25] (a) to (y);
    \draw [->, line width=1.25] (c) to [bend right=20] (y);
    
\end{tikzpicture}
\caption{DAG based on \citet{zhou2019selection}.}
\label{fig:Zhou_DAG}
\medskip{}
\par\end{centering}
Note: $B$ denotes a set of baseline confounders; $X$ denotes parental income rank; $L$ and $A$ denote educational expectations and AFQT scores when the respondent was in high school, respectively; $C$ denotes whether the respondent graduated from college; and $Y$ denotes the respondent's income rank as an adult. For simplicity, the random disturbances are suppressed from this graph.
\end{figure}

We conceptually replicate Zhou's \citeyearpar{zhou2019selection} analysis using a cGNF based on the DAG in Figure \ref{fig:Zhou_DAG}. In this graph, the baseline confounders ($B$) affect all downstream variables. Parental income ($X$) influences a respondent's income ($Y$) and their likelihood of college graduation ($C$). It also influences a respondent's educational expectations ($L$) and AFQT scores ($A$) in high school. Additionally, attaining a college degree affects respondent income, and this effect is confounded by their expectations and test scores.

We train our cGNF with the NLSY data, adopting the same architecture, hyper-parameters, and stopping criterion as outlined in the previous section. After training, we then use our model to estimate two effects that reflect the influence of education on income mobility. 

First, we estimate the conditional effect of parental income on respondent income, given a respondent's educational attainment. This effect can be formally expressed as $\mathbf{CATE}_{X{\rightarrow}Y|C} = \mathbf{E} \left[Y(x^{*}) - Y(x)|C=c\right]$. It captures the influence of parental income on respondent income among sub-populations of respondents categorized by their observed level of educational attainment.

Second, we also estimate a joint effect of parental income and respondent education on respondent income. This estimand can be formally expressed as $\mathbf{AJE}_{X,C{\rightarrow}Y} = \mathbf{E} \left[Y(x^{*},c) - Y(x,c)\right]$, which is a type of controlled direct effect \citep{vanderweele2015explanation}. It captures the influence of parental income on respondent income under a hypothetical intervention where all respondents attain the level of education given by $c$.

In addition, we also estimate the path-specific effects of parental income on respondent income that operate through educational expectations, AFQT scores, and college graduation. As special types of indirect effects, the estimands we consider here capture the mediating role of a given variable in the causal chain connecting parental with respondent income, net of other potential mediators that precede it \citep{zhou2023tracing}.

Specifically, the first of these effects can be defined as follows:
\begin{IEEEeqnarray}{rCl}
\mathbf{PSE}_{X{\rightarrow}C{\rightarrow}Y} &=& \mathbf{E} \big[ Y(x^{*},L(x),A(x,L(x))) - Y(x^{*},L(x),A(x,L(x)),C(x,L(x),A(x,L(x))) \big]. \hfill \IEEEyesnumber 
\label{eq:PSE_XtoCtoY}
\end{IEEEeqnarray}
This expression captures an effect of parental income on respondent income mediated through a respondent's educational attainment, but not their prior expectations or test scores. It is represented by the $X{\rightarrow}C{\rightarrow}Y$ path in isolation. 

Similarly, another path-specific effect captures the influence of parental income on respondent income mediated through AFQT scores, but not prior expectations. This effect can be expressed as follows:
\begin{IEEEeqnarray}{rCl}
\mathbf{PSE}_{X{\rightarrow}A{\rightsquigarrow}Y} &=& \mathbf{E} \big[ Y(x^{*},L(x)) - Y(x^{*},L(x),A(x,L(x))) \big]. \hfill \IEEEyesnumber 
\label{eq:PSE_XtoAtoY}
\end{IEEEeqnarray}
It reflects the influence of parental income transmitted along the $X{\rightarrow}A{\rightarrow}Y$ and $X{\rightarrow}A{\rightarrow}C{\rightarrow}Y$ paths combined. 

The last effect of interest can be defined as follows:
\begin{IEEEeqnarray}{rCl}
\mathbf{PSE}_{X{\rightarrow}L{\rightsquigarrow}Y} &=& \mathbf{E} \big[ Y(x^{*}) - Y(x^{*},L(x)) \big]. \hfill \IEEEyesnumber 
\label{eq:PSE_XtoLtoY}
\end{IEEEeqnarray}
This expression captures the influence of parental income on respondent income mediated through educational expectations, as transmitted along the $X{\rightarrow}L{\rightarrow}Y$, $X{\rightarrow}L{\rightarrow}A{\rightarrow}Y$, and $X{\rightarrow}L{\rightarrow}A{\rightarrow}C{\rightarrow}Y$ paths together. It is equivalent to the natural indirect effect of $X$ on $Y$ via $L$.

These particular effects, while not examined by \citet{zhou2019selection}, are important nonetheless for evaluating the study's broader theoretical model of mobility and selection. They illuminate the potential for parental income to influence intermediate variables, like expectations and test scores in high school, which subsequently shape post-secondary attainment and income later in adulthood. In other words, they reflect the complex selection process hypothesized to contaminate the observed relationship of higher education with greater income mobility. 

\begin{table}[h!]
    \begin{centering}
    \caption{cGNF Estimates of the Conditional, Controlled, and Path-Specific Effects of Parental Income on Respondent Income}
    \label{tab:zhou_mobility_effects}
    \begin{tabular}{l@{\hspace{2.5cm}}c@{\hspace{0.75cm}}c} 
        \hline
        \hline
        \rule{0pt}{2.5ex}Estimand & Point Est. & 90\% Bootstrap CI \\
        \midrule
        \addlinespace[0.3cm] 
        \multicolumn{3}{l}{\textbf{Conditional Effects}} \\
        \(\quad \mathbf{CATE}_{X \rightarrow Y|C=0}\) & .102 & (.062, .135) \\
        \(\quad \mathbf{CATE}_{X \rightarrow Y|C=1}\) & .121 & (.051, .146) \\
        
        \addlinespace[0.3cm]
        \multicolumn{3}{l}{\textbf{Controlled Direct Effects}} \\
        \(\quad \mathbf{AJE}_{X,C=0 \rightarrow Y}\) & .097 & (.062, .131) \\
        \(\quad \mathbf{AJE}_{X,C=1 \rightarrow Y}\) & .077 & (-.026, .103) \\

        \addlinespace[0.3cm]
        \multicolumn{3}{l}{\textbf{Path-specific Effects}} \\
        \(\quad \mathbf{PSE}_{X \rightarrow C \rightarrow Y}\) & .003 & (-.002, .007) \\
        \(\quad \mathbf{PSE}_{X \rightarrow A \rightsquigarrow Y}\) & .023 & (.008, .025) \\
        \(\quad \mathbf{PSE}_{X \rightarrow L \rightsquigarrow Y}\) & .008 & (-.002, .019) \\
        \bottomrule
    \end{tabular}
\medskip{}
\small
\par\end{centering}
Note: All effects contrast the first with the third sample quartile of parental income ($X$). Confidence intervals are based on the 5th and 95th percentiles of a bootstrap distribution with 600 replicates.
\end{table}

Table \ref{tab:zhou_mobility_effects} presents the conditional, controlled, and path-specific effects of interest, as estimated by our cGNF. All of them contrast the first quartile with the third quartile of parental income. Overall, they are consistent with the results reported by \citet{zhou2019selection}. Estimates for the conditional effects indicate that the influence of parental income on respondent income--after adjusting for observed selection--is fairly similar among those with and without a college education ($\widehat{CATE}_{X \rightarrow Y|C=0}=.102$ versus $\widehat{CATE}_{X \rightarrow Y|C=1}=.121$). Furthermore, estimates for the controlled direct effects suggest that the influence of parental income on respondent income would also be fairly similar regardless of any intervention to expand or contract access to higher education ($\widehat{AJE}_{X,C=0 \rightarrow Y}=.097$ versus $\widehat{AJE}_{X,C=1 \rightarrow Y}=.077$). This suggests that increasing access to college is unlikely to boost income mobility very much, in line with Zhou's \citeyearpar{zhou2019selection} conclusions.

Our estimates for the path-specific effects point toward a potential explanation. They suggest that the influence of parental income operating exclusively through its impact on respondent education, but not through upstream factors like expectations and test scores in high school, is small ($\widehat{PSE}_{X \rightarrow C \rightarrow Y}=0.003$). Rather, the most important mechanism through which parental income affects respondent income appears to involve achievement test scores in high school ($\widehat{PSE}_{X \rightarrow A \rightsquigarrow Y}=0.023$). These findings suggest that higher parental incomes may lead to higher incomes for the next generation partly because they contribute to better college preparedness, thereby enhancing the likelihood of graduation and subsequent success in the labor market. Replication files for this analysis are available at \url{https://github.com/gtwodtke/deep_learning_with_DAGs/tree/main/zhou_2019}.

\section{Discussion}

In this study, we introduced causal-graphical normalizing flows (cGNFs), a novel approach to analyzing systems of causal relationships between variables. This approach integrates directed acyclic graphs (DAGs) with unconstrained monotonic neural networks (UMNNs) to model entire causal systems, without relying on restrictive parametric assumptions. The key advantage of cGNFs lies in their ability to flexibly approximate the full joint distribution of the data, using its Markov factorization. This facilitates estimation via Monte Carlo sampling for a wide range of causal estimands, including but not limited to total, conditional, direct, indirect, and path-specific effects. Extending recent advances in machine learning and causal inference \citep{athey2019grf, chernozhukov2018debiased, koch2021deep, vanderlaan2006targeted}, cGNFs transcend the prevailing focus on a narrow set of estimands and unlock the possibility of comprehensively evaluating more elaborate causal theories.

We illustrated the utility of cGNFs by reanalyzing two seminal studies of social mobility in the United States. The first reanalysis, drawing on Blau and Duncan's \citeyearpar{blau1967american} study of the status attainment process, highlighted the method's ability to uncover and describe nonlinear relationships. These relationships are often obscured by the functional form constraints traditionally imposed by parametric structural equation models (SEMs). Our second reanalysis focused on Zhou's \citeyearpar{zhou2019selection} study of conditional versus controlled mobility. In this illustration, we demonstrated the ability of cGNFs to detect complex forms of effect moderation and interaction, to adjust for dynamic selection processes, and to concurrently evaluate multiple aspects of a broader causal theory. Together, these empirical examples highlight the potential of cGNFs for modeling entire causal systems.

While cGNFs offer numerous advantages, they are certainly not without limitations, which in turn suggest important directions for future research. A central limitation is that cGNFs depend on the analyst to identify a correct DAG summarizing the causal relationships of interest. If the assumed DAG is incorrect, cGNFs may produce inaccurate estimates for certain effects in the causal system. This limitation is not unique to cGNFs. It afflicts almost any method of causal inference, including regression imputation, propensity score matching or weighting, and instrumental variables, where incorrect assumptions about the underlying structural model can lead to faulty estimates. However, the task of identifying a correct DAG is arguably more important for cGNFs, since their objective is to evaluate multiple features of a broader causal system. Notably, cGNFs can still provide accurate estimates for certain causal effects, even when parts of the DAG are incorrectly specified. For example, they can still accurately estimate the total effect of one variable on another, even if the DAG assumes an inaccurate set of relationships among the variables that confound this effect. Future research must systematically establish, in generalizable terms, which regions of the DAG must be accurately specified for cGNFs to reliably estimate particular effects within the broader causal system.

Similarly, cGNFs do not avoid any of the challenges associated with unobserved confounding, which are ubiquitous in the social sciences. Identifying and measuring the many different variables that may confound an effect of interest is a formidable task, regardless of the approach to modeling and estimation. In general, if a target estimand is not identified due to unobserved confounding, cGNFs will yield inaccurate estimates for this effect. In the present study, we demonstrated how cGNFs can be seamlessly integrated with methods for assessing the sensitivity of estimates to unobserved confounding. Beyond this approach, future research should also explore how cGNFs might be combined with other methods designed to identify causal effects when unobserved confounding is present. These include non-parametric instrumental variable models \citep{newey2003instrumental, newey2013nonparametric} and deep learning techniques for inferring and controlling latent confounders by proxy \citep{louizos2017causal}.

Even with a correct DAG and a cleanly identified estimand, cGNFs face another limitation: at present, there is no theoretical guarantee that they will provide consistent point estimates or that bootstrap intervals will have asymptotically valid coverage rates. As universal density approximators, UMNNs possess the capacity to model any distribution to an arbitrary degree of accuracy \citep{Huang2018NAF, wehenkel2019UMNN}. Thus, with a sufficiently expressive architecture and enough data for training, they can, in principle, recover any feature of the target distribution exactly. However, the precise conditions required for UMNNs to reach this level of accuracy are not yet fully understood. What are the architectures, hyper-parameter settings, and volume of data needed for cGNFs to estimate causal effects with no more than a trivial degree of error?

In Part \ref{subsec:appendix_bias_variance} of the Appendix, we present results from a series of Monte Carlo experiments indicating that cGNFs yield estimates with low bias and variance in sufficiently large samples, even with relatively simple architectures and standard hyper-parameter settings. In general, the bias and variance appear to decline monotonically as the sample size increases, and for samples of 16,000 cases or more, cGNF estimates for a wide range of effects are approximately unbiased and exhibit high stability under repeated sampling. We observe this pattern of results whether the data generating process is very simple (i.e., linear and additive with normal disturbances), incorporates discrete variables, or involves substantial non-linearity, effect heterogeneity, and non-normality.

Existing simulation studies additionally suggest that bootstrap methods generate confidence intervals with satisfactory coverage rates for neural network predictions \citep{franke2000bootstrapping, heskes1996practical}. In some cases, bootstrap intervals even appear to be conservative \citep{papadopoulos2001confidence}. In Part \ref{subsec:appendix_coverage} of the Appendix, we present results from a Monte Carlo experiment that align with these prior studies. Specifically, our experiment suggests that, with a sufficiently large sample ensuring minimal bias in cGNF estimates, bootstrap intervals have coverage rates that are slightly conservative. Nevertheless, until the asymptotic properties of cGNFs are conclusively established, any inferential statistics derived from them should be interpreted with caution.

To address uncertainties surrounding inference, a potential solution involves utilizing cGNFs to generate the components of multiply robust estimators with known asymptotic properties. These estimators include targeted maximum likelihood, augmented inverse probability weighting, and other related approaches based on the efficient influence curve \citep{glynn2010introduction, vanderlaan2006targeted, zhou2022semiparametric}. Because cGNFs provide access to the full joint distribution of the data, they can be used to construct all the terms that compose these estimators, including propensity scores, conditional means, and residuals. Using cGNFs in this way would produce effect estimates endowed with desirable properties, such as consistency, efficiency, and asymptotic normality. The advantage of cGNFs over other machine learning methods is their ability to generate the components for many different robust estimators simultaneously, each tailored to a particular effect in a broader causal system. Thus, future research should further explore the possibility of combining cGNFs with multiply robust approaches to estimation. 

Another challenge for cGNFs involves the limited capacity of neural networks to interpolate or extrapolate beyond the observed data. Although cGNFs can accurately model any joint distribution in theory, their estimates may be less reliable in regions of the data space with few observations. Consequently, the performance of cGNFs may vary inversely with the degree of sparsity, or in other words, their ability to model entire causal systems, sans functional form assumptions, comes with an increased hunger for data. Researchers must therefore exercise caution in assessing the suitability of any positivity conditions necessary for non-parametric identification of their target estimands.

The computational complexity of training cGNFs presents an additional challenge. For example, in our reanalysis of \citet{blau1967american}, training the model and generating point estimates required about 3 hours of wall time on an Nvidia A100 graphical processing unit (GPU). Computing the 600 bootstrap estimates in this analysis took another 44 hours using a high-performance computing cluster (HPC), with each bootstrap iteration distributed across a separate central processing unit (CPU).\footnote{Specifically, we parallelized the calculations for each bootstrap sample across 15 HPC nodes and 40 CPUs per node.} While executing this analysis on a standard personal computer, which is typically equipped with 4 to 16 CPUs and a consumer-grade GPU, remains possible, the total computation time would significantly increase. Future research should therefore prioritize increasing the speed and efficiency of the computations needed to train a cGNF and then generate effect estimates.

Beyond the challenges of sparsity and computational complexity, model selection is another area ripe for additional research \citep{koch2021deep}. In our context, model selection involves comparing and choosing among a set of candidate cGNFs after training, each distinguished by differences in its architecture or hyper-parameters. These models might also vary based on the values used to initialize their weights, as the loss functions associated with cGNFs are non-convex, and it is possible for the training process to terminate in a local rather than global minimum. The range of approaches for selecting architectures and hyper-parameters is broad, spanning from entirely random choices to exhaustive grid searches. Alternatively, model selection could also be directed by theoretical insights and prior knowledge, focusing on models presumed to be sufficiently expressive and to train smoothly. A common metric used for these comparisons is the best validation loss achieved before the training algorithm terminates, although other metrics may be worth exploring (e.g., Akaike or Bayesian information criteria). 

In Part \ref{subsec:appendix_hyper} of the Appendix, we provide evidence from another set of Monte Carlo experiments, indicating that the performance of cGNFs is fairly robust to modest variations in model architecture and hyper-parameter settings. The experiments demonstrate that, across a range of sensible architectures, batch sizes, and learning rates, cGNFs consistently produce estimates with low bias and variance in sufficiently large samples. Despite these encouraging results, the development of more sophisticated model selection techniques--specifically tailored for deep neural networks designed to estimate causal effects--remains a critical necessity. Future research focusing on model selection with cGNFs will be essential to further enhance the reliability of these methods for causal inference \citep{alaa2019validating, parikh2022validating}.

Deep neural networks, including cGNFs, are also not immune to the challenges posed by measurement error. Inaccuracies in the input data can distort the training process, resulting in erroneous output. While parametric structural equation models (SEMs) can presently incorporate an extensive set of techniques for addressing measurement error \citep{bollen1989structural}, analogous methods for use with deep learning models are still in their infancy \citep{hu2022measurement}. Because measurement error is pervasive in social science data, future research should explore the possibility of integrating measurement models into the cGNF architecture. This could potentially mitigate the impact of inaccurately measured inputs and improve the network's ability to distinguish signal from noise.

These limitations notwithstanding, cGNFs offer enormous potential for learning about causal systems in the social sciences, transcending parametric SEMs that are typically based on naive assumptions about functional form. We expect that they will find wide application, not only in research on social mobility but wherever interest lies in studying broader systems of causal relationships. While current limitations may forestall the prudent application of cGNFs in some cases, this article lays the foundation for future research aimed at addressing these challenges, including those related to sparsity, valid inference, computational complexity, model selection, and measurement error. By building on recent advances in machine learning and causal inference, cGNFs represent a significant step toward a more complete integration of the deep learning and causal revolutions \citep{pearl2018theoretical, pearl2018bookofwhy, sejnowski2018deep}.

\bibliographystyle{apalike}
\bibliography{SMRrefs.bib}

\newpage

\appendix

\section{Appendix}\label{sec:appendix}

\subsection{Dequantization}\label{subsec:appendix_dequant}

A normalizing flow may assume any form, provided that it is bijective and that the transformation and its inverse are both smooth with finite first derivatives. Transformations that satisfy these criteria are known as \textit{diffeomorphisms} \citep{milnor1997diffeomorphism}. For a continuous variable $X_1$, the normalizing flow $Z_1=\boldsymbol{h}\left(X_1\right)$ can be conceptualized as $Z_1=\boldsymbol{h}\left(X_1\right)=F_{Z}^{-1}(F_{X_1}(X_1))$, where $F_{X_1}$ is the cumulative distribution function (CDF) for $X_1$ and $F_{Z}^{-1}$ is the inverse of the standard normal CDF. For a continuous variable with a smooth, invertible, and differentiable CDF, the composition of this CDF with the inverse normal CDF is a diffeomorphism, and by extension, normalizing flows can readily model continuous distributions by mapping them to and from the standard normal distribution. 

However, if $X_1$ is discrete, its CDF $F_{X_1}$ is neither bijective nor smooth with a finite first derivative. Thus, to adapt normalizing flows for discrete data, research in this area has focused on integrating them with different forms of dequantization (e.g., \citealt{Nielsen2020DeqGap, Uria2013RNADETR_uniformdeq, ZieglerR2019VAE_dequant}). Dequantization converts a discrete variable into a continuous variable by adding a small amount of random noise to each of the discrete values. The amount of random noise is selected so that the original discrete values can be easily recovered by rounding off the dequantized variable to its nearest integer. 

In our application of normalizing flows, we dequantize discrete variables by adding normally distributed noise with zero mean and a variance of $1/36$, following \citet{balgi2022cgnf}. With a variance ${\leq}1/36$, nearly all the random noise added to each discrete value will lie in the range $[-0.5, +0.5]$, such that the discrete values can be covered by rounding to the nearest integer with virtually no loss of information. The addition of random noise drawn from $\mathcal{N}(0,1/36)$ converts a discrete variable into a continuous variable that follows a multimodal Guassian mixture distribution, with modes at each of the original values on the discrete variable. The CDF of this Guassian mixture is bijective and smooth with a finite first derivative, and thus its composition with the inverse normal CDF is a diffeomorphism, as above. 

A normalizing flow can then be used to map the dequantized continuous variable with a multimodal Gaussian mixture distribution to the standard normal distribution. In addition, its inverse can map from the standard normal distribution back to the Guassian mixture, and then the original discrete variable can be recovered by rounding the dequantized variable to its nearest integer. In this way, normalizing flows with Gaussian dequantization model discrete distributions by approximating them with a multimodal normal mixture, and then they recover the original mass points of the discrete distribution by rounding.

\subsection{Monte Carlo Experiments on Bias, Variance, and Asymptotic Behavior}\label{subsec:appendix_bias_variance}

To examine the performance of causal-Graphical Normalizing Flows (cGNFs), we conducted a series of Monte Carlo experiments. These experiments involved training cGNFs and using them to estimate causal effects with simulated data generated from three different structural equation models. The first data-generating model was linear and additive with normally distributed disturbances. It can be formally represented as follows:
\[
\begin{aligned}
C &\sim N(0, 1) \\
A &\sim N(0.1C, 1) \\
L &\sim N(0.2A + 0.2C, 1) \\
M &\sim N(0.1A + 0.2C + 0.25L, 1) \\
Y &\sim N(0.1A + 0.1C + 0.25M + 0.25L, 1) \\
\end{aligned}
\]
The second data-generating model was more complex, as it involves only discrete variables and also incorporates non-additive relationships among them. This model can be formally represented as follows:
\[
\begin{aligned}
C &\sim \text{Multinomial}(0.3, 0.5, 0.2) \\
A &\sim \text{Bernoulli}(0.3 + 0.1C) \\
L &\sim \begin{cases} 
\text{Multinomial}(0.5, 0.3, 0.2) & \text{ if } A = 1 \text{ and } C = 1 \\
\text{Multinomial}(0.3, 0.5, 0.2) & \text{ if } A = 1 \text{ and } C = 2 \\
\text{Multinomial}(0.2, 0.3, 0.5) & \text{ if } A = 1 \text{ and } C = 3 \\
\text{Multinomial}(0.6, 0.2, 0.2) & \text{ if } A = 0 
\end{cases} \\
M &\sim \text{Bernoulli}\left(\text{logit}^{-1}\left(-0.5 + 0.4A + 0.2C + 0.3L\right)\right) \\
Y &\sim \text{Bernoulli}\left(\text{logit}^{-1}\left(-0.5 + 0.3A + 0.1C + 0.3M + 0.3AM + 0.3L\right)\right)
\end{aligned}
\]
The last data-generating model is even more complex. It incorporates both discrete and continuous variables, nonlinear relationships, effect heterogeneity, heteroscedasticity, and highly non-normal disturbances. Specifically, this model can be formally represented as follows:
\[
\begin{aligned}
C &\sim \text{Laplace}(0, 1) \\
A &\sim \text{Bernoulli}\left(\text{logit}^{-1}\left(0.1C\right)\right) \\
L &\sim \text{Tukey-Lambda}(0.2A + 0.2C + 0.1AC, 1, 0.3, 0.7) \\
M &\sim\text{Student's t}(10) + 0.1A + 0.2C^2 + 0.25L + 0.15AL \\
Y &\sim \text{Normal}(0.1A + 0.1C^2 + 0.2M + 0.2AM + 0.25L^2, |C|)
\end{aligned}
\]

We simulated 400 to 600 datasets from each model, with sample sizes progressively doubling from 2,000 to 128,000 cases. For each dataset, we trained a cGNF using a batch size of 128, a learning rate of 0.0001, and a stopping criterion of no reduction in the loss for 50 consecutive epochs, computed on a one-fifth validation sample. The architecture of the cGNF included an embedding network with five hidden layers, containing 100, 90, 80, 70, and 60 nodes respectively, and an integrand network also composed of five hidden layers, but with 60, 50, 40, 30, and 20 nodes each.

After training, we used each cGNF to estimate several causal effects. First, we estimated the average total effects of $A$ on $M$ and $Y$, denoted as $\mathbf{ATE}_{A \rightarrow M} = \mathbf{E}[M(a^*) - M(a)]$ and $\mathbf{ATE}_{A \rightarrow Y} = \mathbf{E}[Y(a^*) - Y(a)]$ respectively. Second, we also estimated the natural direct and indirect effects $A$ on $M$, as mediated by $L$. These effects can be expressed as $\mathbf{NDE}_{A \rightarrow L \rightarrow M} = \mathbf{E}[M(a^*, L(a)) - M(a)]$ and $\mathbf{NIE}_{A \rightarrow L \rightarrow M} = \mathbf{E}[M(a^*) - M(a^*, L(a))]$. Lastly, we estimated the path-specific effects of $A$ on $Y$, operating both directly and through $L$ and $M$. These effects can be formally represented as $\mathbf{PSE}_{A \rightarrow Y} = \mathbf{E}[Y(a^*, L(a), M(a, L(a))) - Y(a)]$, $\mathbf{PSE}_{A \rightarrow M \rightarrow Y} = \mathbf{E}[Y(a^*, L(a)) - Y(a^*, L(a), M(a, L(a)))]$, and $\mathbf{PSE}_{A \rightarrow L \rightsquigarrow Y} = \mathbf{E}[Y(a^*) - Y(a^*, L(a))]$. In all cases, we contrasted $a^{*}=1$ with $a=0$. The true values of these estimands under each data-generating model are provided in Table \ref{tab:ground_truth}. Replication files are available at \url{https://github.com/gtwodtke/deep_learning_with_DAGs/tree/main/MCEs}.

\begin{table}[h!]
    \begin{centering}
    \caption{True Values of Target Estimands in each Monte Carlo Experiment}
    \label{tab:ground_truth}
    \begin{tabular}{lccc} 
        \hline
        \hline
        \rule{0pt}{2.5ex}Estimand & Linear DGM & Discrete DGM & Non-linear DGM \\
        \midrule
        \(\mathbf{ATE}_{A \rightarrow Y}\) & .180 & .143 & .325 \\
        \(\mathbf{PSE}_{A \rightarrow Y}\) & .100 & .109 & .189 \\
        \(\mathbf{PSE}_{A \rightarrow L \rightsquigarrow Y}\) & .060 & .022 & .085 \\
        \(\mathbf{PSE}_{A \rightarrow M \rightarrow Y}\) & .020 & .012 & .051 \\
        \(\mathbf{ATE}_{A \rightarrow M}\) & .150 & .113 & .207 \\
        \(\mathbf{NDE}_{A \rightarrow M}\) & .100 & .092 & .127 \\
        \(\mathbf{NIE}_{A \rightarrow L \rightarrow M}\) & .050 & .020 & .080 \\
        \bottomrule
    \end{tabular}
\medskip{}
\small
\par\end{centering}
Note: The true values for our target estimands in the first and second data-generating models (DGMs) were calculated analytically using their non-parametric identification formulas. For the third DGM, these values were computed numerically using 10 billion Monte Carlo samples drawn from of mutilated models. All estimands contrast $a^{*}=1$ with $a=0$.
\end{table}

\begin{figure}[ht!]
\begin{centering}
    \begin{subfigure}[b]{0.8\linewidth} 
        \includegraphics[width=\linewidth,height=0.5\textheight,keepaspectratio]{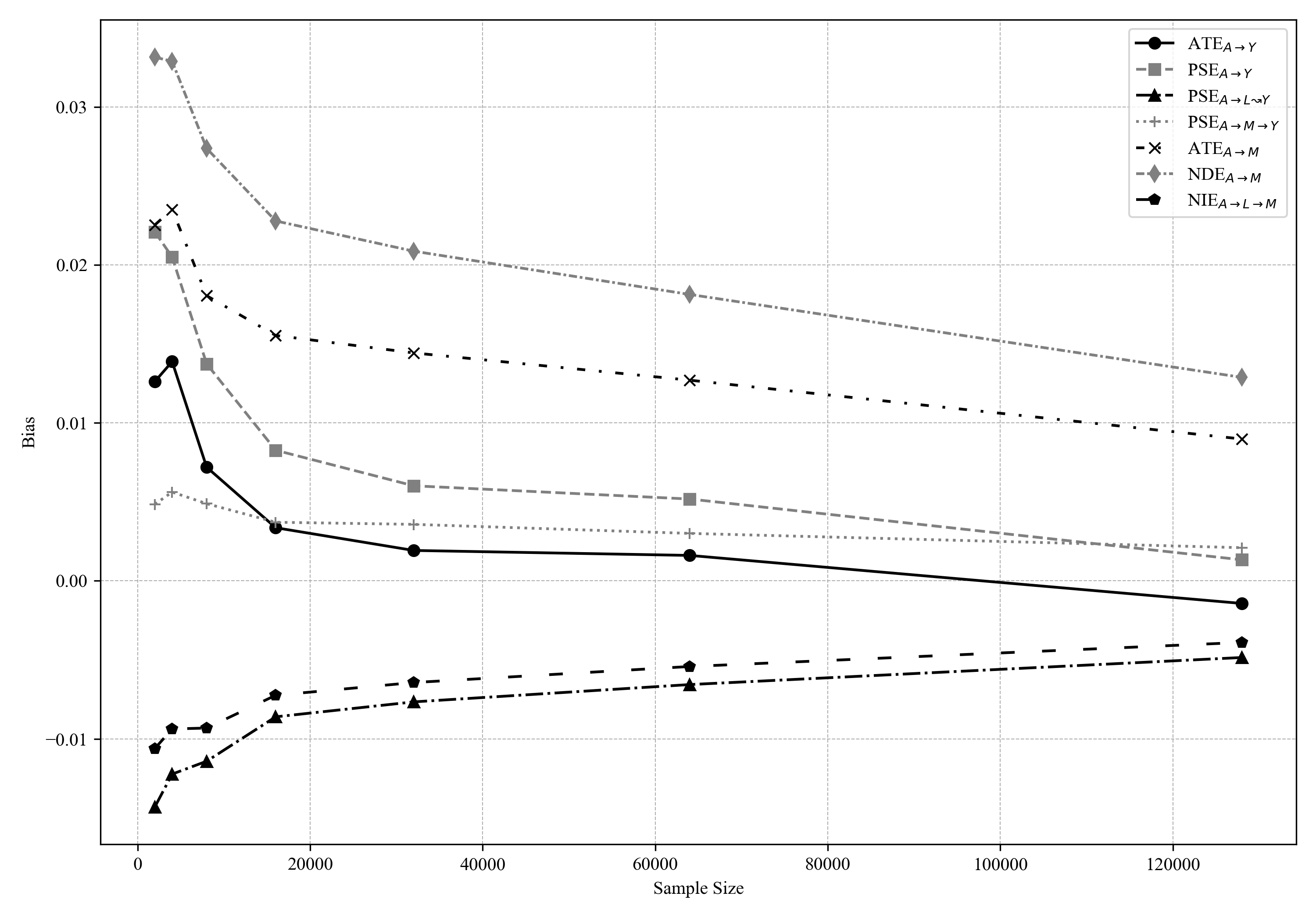}
        \caption{Bias of cGNF Effect Estimates}
        \label{fig:linear_bias}
    \end{subfigure}
    \\[1ex] 
    \begin{subfigure}[b]{0.8\linewidth} 
        \includegraphics[width=\linewidth,height=0.5\textheight,keepaspectratio]{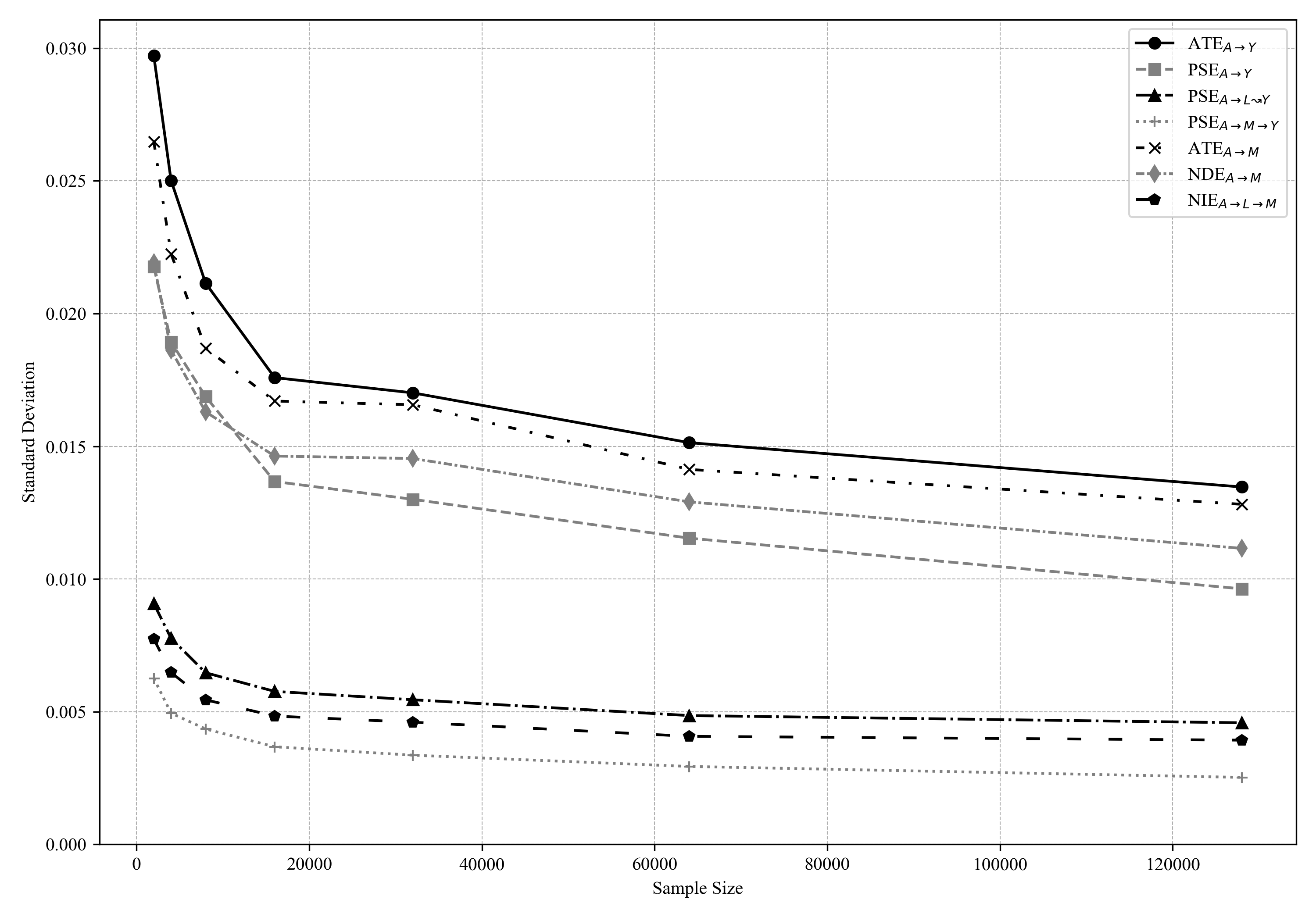}
        \caption{Standard deviation of cGNF Effect Estimates}
        \label{fig:linear_sd}
    \end{subfigure}
    \caption{Performance of cGNFs in Monte Carlo Experiments Based on the Normal, Linear, and Additive Data Generating Process.}
    \label{fig:combined_linear_MCEs}
\medskip{}
\par\end{centering}
Note: Each set of results is based on 600 simulated datasets. All cGNFs have an embedding network with five hidden layers, composed of 100, 90, 80, 70, and 60 nodes, and an integrand network also with five hidden layers, composed of 60, 50, 40, 30, and 20 nodes. The models are trained using a learning rate of 0.0001 and a batch size of 128. 
\end{figure}

Figure \ref{fig:combined_linear_MCEs} presents results from the experiments based on the normal, linear, and additive data-generating model. Panel (a) plots the bias of the cGNF estimates against the sample size, while Panel (b) displays their standard deviation. With data simulated from a relatively simple model, cGNF estimates exhibit low bias and variance, even in relatively small samples. As the sample size increases, the bias and variance converge toward zero, with both metrics stabilizing around 16,000 cases for all target estimands and then declining more slowly thereafter. When parallelized over 15 nodes of a high-performance computing cluster (HPC), each with 40 central processing units (CPUs), the wall time for these experiments was approximately 1 day and 8 hours. 

\begin{figure}[ht!]
\begin{centering}
    \begin{subfigure}[b]{0.8\linewidth} 
        \includegraphics[width=\linewidth,height=0.5\textheight,keepaspectratio]{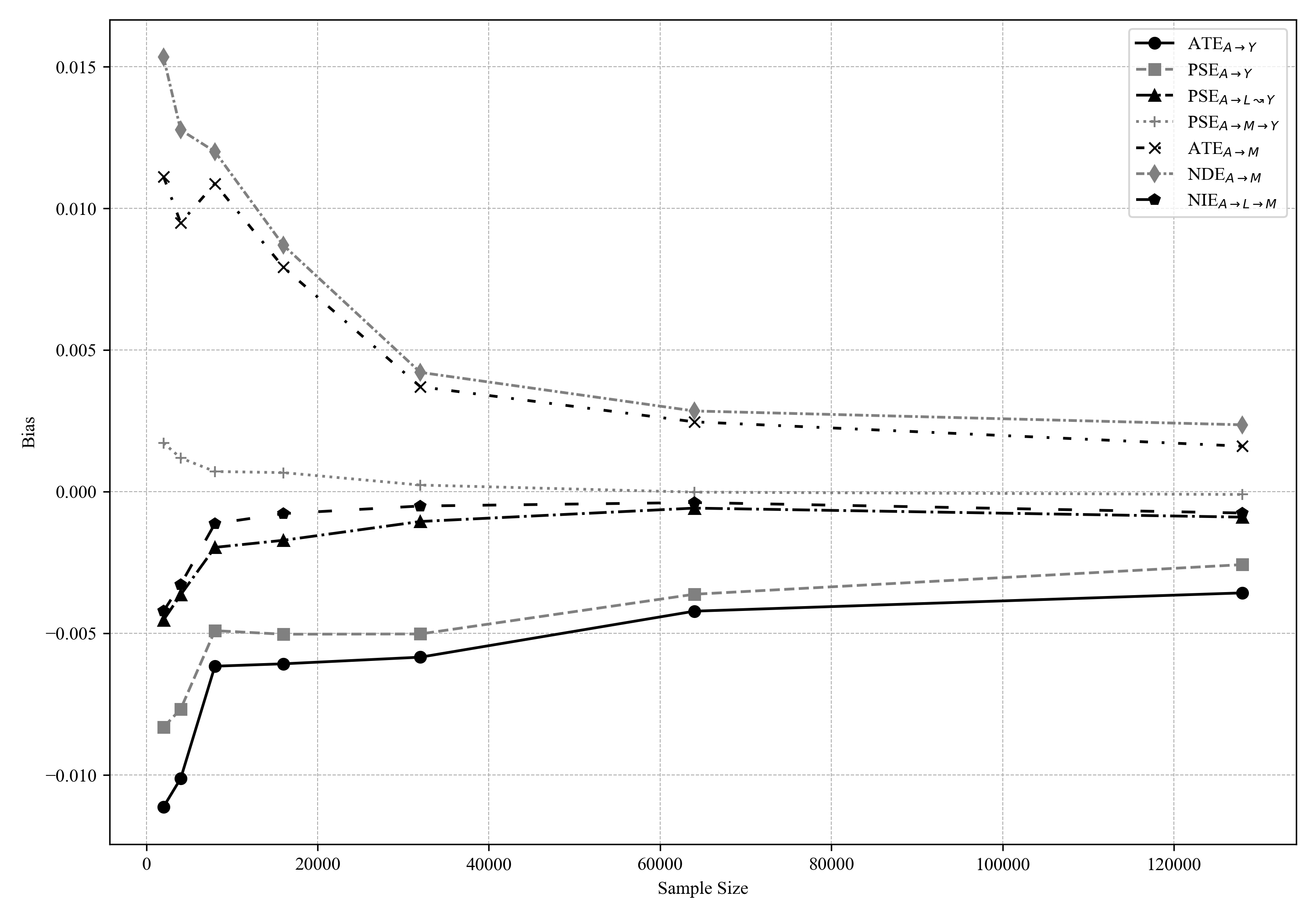}
        \caption{Bias of cGNF Effect Estimates}
        \label{fig:discrete_bias}
    \end{subfigure}
    \\[1ex] 
    \begin{subfigure}[b]{0.8\linewidth} 
        \includegraphics[width=\linewidth,height=0.5\textheight,keepaspectratio]{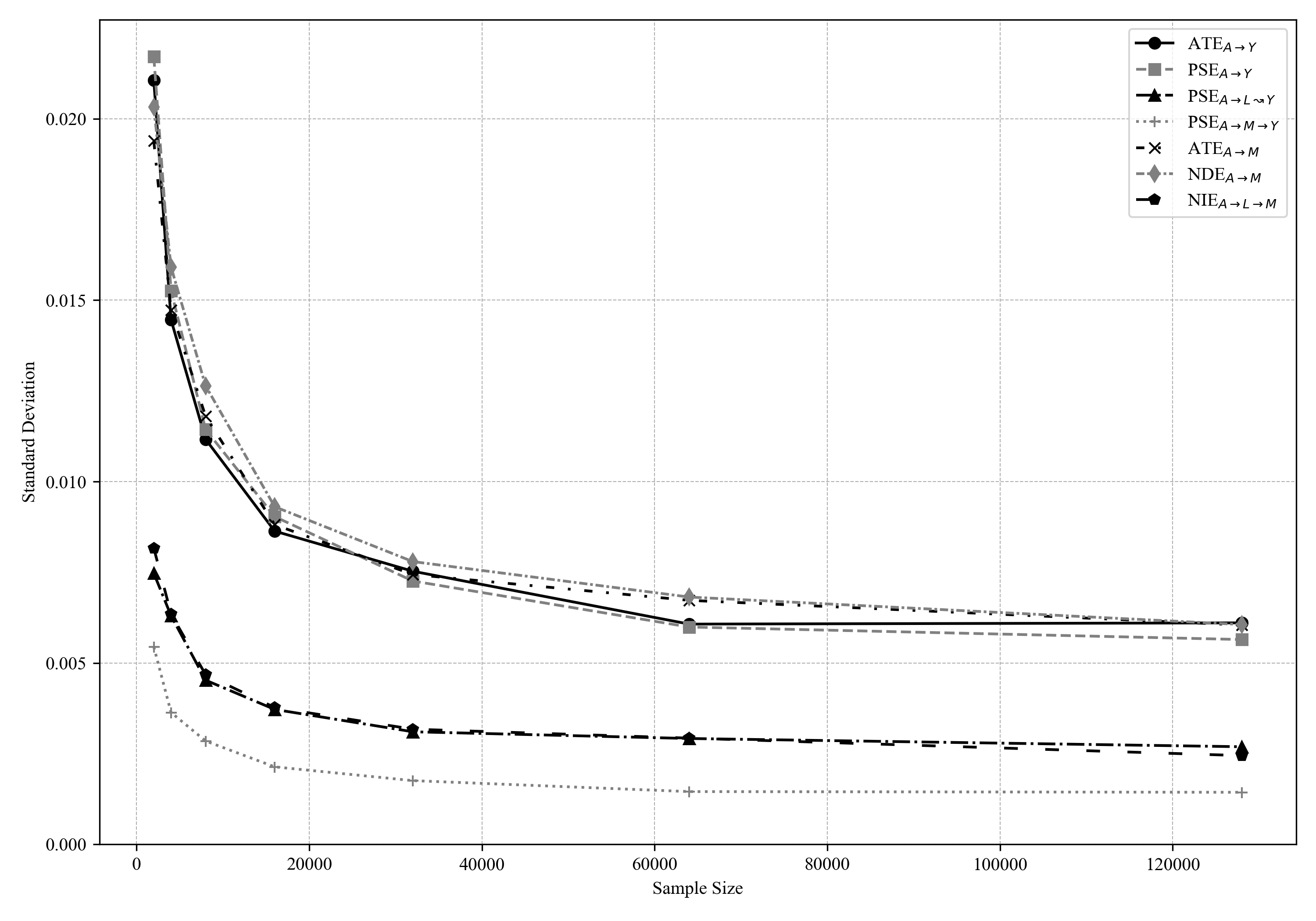}
        \caption{Standard deviation of cGNF Effect Estimates}
        \label{fig:discrete_sd}
    \end{subfigure}
    \caption{Performance of cGNFs in Monte Carlo Experiments Based on the Discrete Data Generating Process.}
    \label{fig:combined_discrete_MCEs}
\medskip{}
\par\end{centering}
Note: Each set of results is based on 400 simulated datasets. All cGNFs have an embedding network with five hidden layers, composed of 100, 90, 80, 70, and 60 nodes, and an integrand network also with five hidden layers, composed of 60, 50, 40, 30, and 20 nodes. The models are trained using a learning rate of 0.0001 and a batch size of 128. 
\end{figure}

Figure \ref{fig:combined_discrete_MCEs} presents results from the experiments based on the discrete data-generating model. With discrete data, cGNF estimates also generally exhibit low bias and variance, and as the sample size increases, the bias and variance again appear to converge toward zero. Both metrics stabilize at low levels between 16,000 and 32,000 cases, declining more slowly thereafter. The wall time for these experiments was approximately 1 day and 22 hours, when parallelized over 10 nodes of a HPC with 40 CPUs each. 

\begin{figure}[ht!]
\begin{centering}
    \begin{subfigure}[b]{0.8\linewidth} 
        \includegraphics[width=\linewidth,height=0.5\textheight,keepaspectratio]{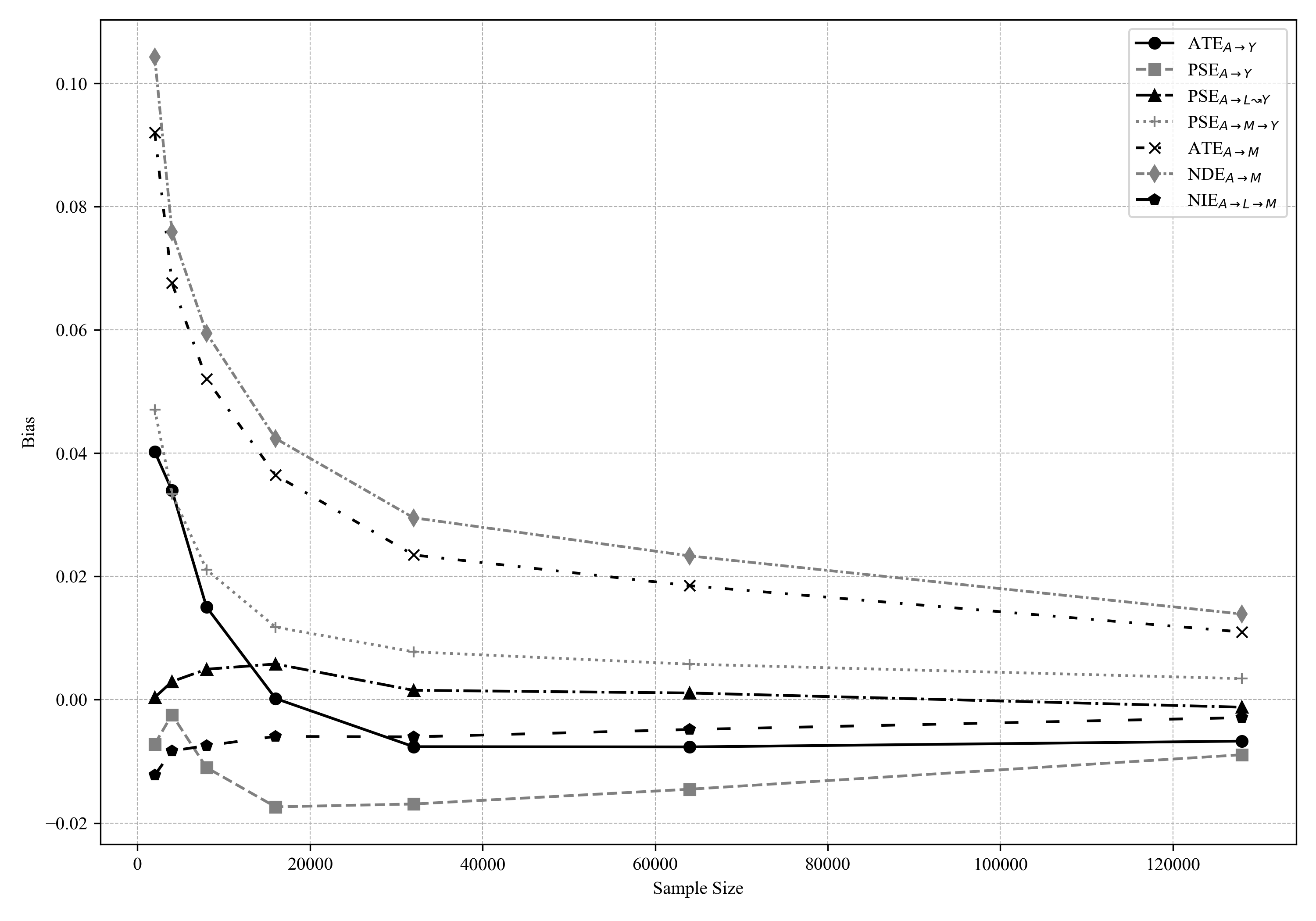}
        \caption{Bias of cGNF Effect Estimates}
        \label{fig:nonlinear_bias}
    \end{subfigure}
    \\[1ex] 
    \begin{subfigure}[b]{0.8\linewidth} 
        \includegraphics[width=\linewidth,height=0.5\textheight,keepaspectratio]{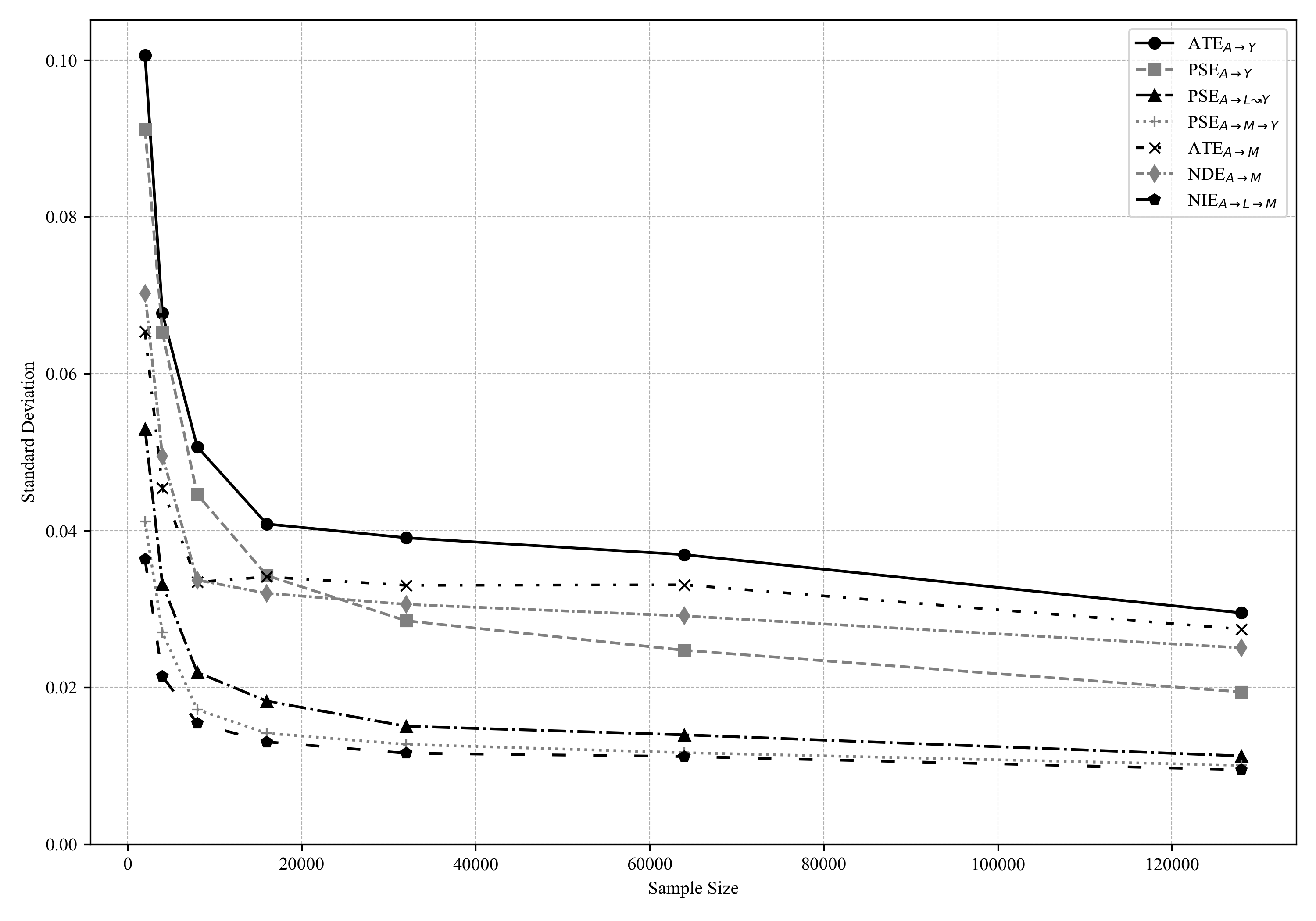}
        \caption{Standard deviation of cGNF Effect Estimates}
        \label{fig:nonlinear_sd}
    \end{subfigure}
    \caption{Performance of cGNFs in Monte Carlo Experiments Based on the Non-normal, Non-linear, and Non-additive Data Generating Process.}
    \label{fig:combined_nonlinear_MCEs}
\medskip{}
\par\end{centering}
Note: Each set of results is based on 400 simulated datasets. All cGNFs have an embedding network with five hidden layers, composed of 100, 90, 80, 70, and 60 nodes, and an integrand network also with five hidden layers, composed of 60, 50, 40, 30, and 20 nodes. The models are trained using a learning rate of 0.0001 and a batch size of 128. 
\end{figure}

Figure \ref{fig:combined_nonlinear_MCEs} summarizes results from the experiments involving a data-generating model with both continuous and discrete variables, non-linearity, heterogeneity, heteroscedasticity, and non-normality. With data simulated from this highly complex process, cGNF estimates initially exhibit nontrivial bias and high variance, particularly in smaller samples with fewer than 8,000 cases. However, as the sample size increases, both the bias and variance diminish rapidly. By the time the sample size reaches 16,000 to 32,000 cases, the bias and variance are small for most estimands. With further increases in sample size, both metrics continue to converge towards zero, resulting in nearly unbiased estimates with minimal variance in large samples. These experiments took approximately 2 days and 13 hours to complete, when parallelized across 10 nodes of a HPC with 40 CPUs each.

\subsection{Monte Carlo Experiments on Bootstrap Interval Coverage}\label{subsec:appendix_coverage}

To examine the performance of bootstrap confidence intervals for effect estimates from causal-Graphical Normalizing Flows (cGNFs), we conducted another Monte Carlo experiment. In this experiment, we simulated 100 datasets, each containing 8,000 cases. For each dataset, we constructed 90 percent confidence intervals for an average total effect using the 5th and 95th percentiles of a bootstrap distribution composed with 200 estimates. 

The datasets were generated from the following model:
\[
\begin{aligned}
C &\sim \text{Bernoulli}(0.6) \\
A &\sim \text{Bernoulli}(0.4 + 0.2C) \\
Y &\sim N(0.2A + 0.4C, 1). 
\end{aligned}
\]
With this model, we targeted the average total effect of $A$ on $Y$, denoted as $\mathbf{ATE}_{A \rightarrow Y} = \mathbf{E}[Y(a^*) - Y(a)]$, which equals 0.2 when contrasting $a^{*}=1$ with $a=0$. We chose a simple data-generating model and a relatively small sample size to ensure that our cGNF estimates for $\mathbf{ATE}_{A \rightarrow Y}$ would be essentially unbiased, while also keeping the experiment computationally tractable with the resources at our disposal. Computing 200 bootstrap estimates for all 100 simulated datasets necessitates training a total of 20,000 cGNFs--a task that could easily exceed the wall time limits on the HPC available to us, if we were to use a larger sample size and/or a more complex data-generating model.

For each simulated dataset, we selected 200 bootstrap samples. Then, for each bootstrap sample, we trained a cGNF using a batch size of 128, a learning rate of 0.0001, and a stopping criterion of no reduction in the loss for 50 consecutive epochs, evaluated on a validation sample comprising one-fifth of the data. The cGNF architecture included an embedding network with five hidden layers, containing 100, 90, 80, 70, and 60 nodes in succession, and an integrand network also composed of five hidden layers with 60, 50, 40, 30, and 20 nodes each. We computed the upper and lower limits of the confidence intervals using the 95th and 5th percentiles, respectively, of the bootstrap distribution for each simulated dataset.

\begin{landscape}
\begin{figure}[p]
\begin{centering}
    \includegraphics[width=\linewidth,height=\textheight,keepaspectratio]{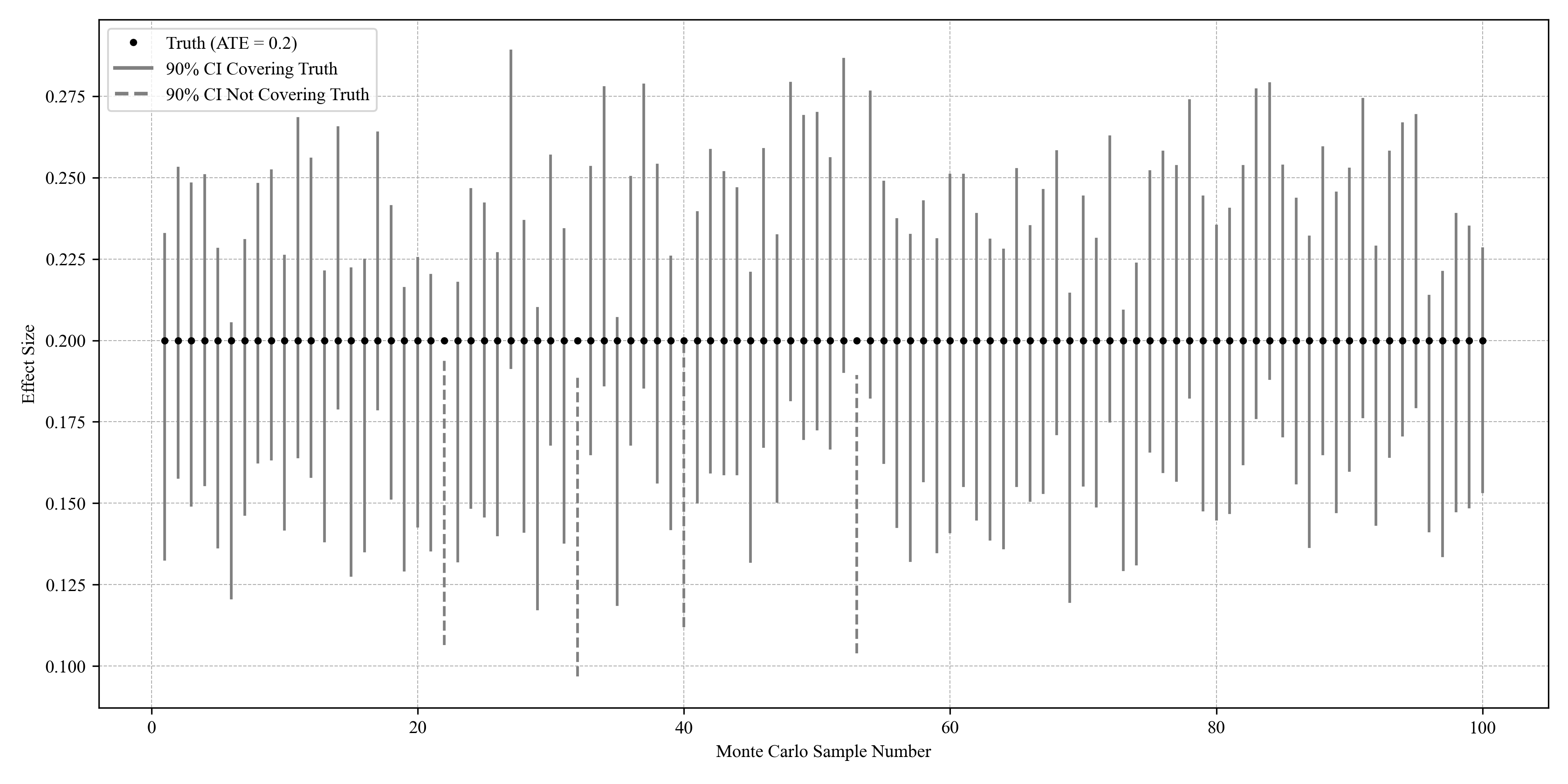}
    \caption{Coverage of Bootstrap Intervals for cGNF Effect Estimates in a Simple Monte Carlo Experiment}
    \label{fig:bootstrap_coverage}
\medskip{}
\par\end{centering}
Note: Results are based on 100 simulated datasets with 200 bootstrap samples per dataset. The figure displays 90 percent confidence intervals based on the 5th and 95th percentiles of the bootstrap distribution for each simulated dataset. All cGNFs have an embedding network with five hidden layers, composed of 100, 90, 80, 70, and 60 nodes, and an integrand network also with five hidden layers, composed of 60, 50, 40, 30, and 20 nodes. The models are trained using a learning rate of 0.0001 and a batch size of 128. 
\end{figure}
\end{landscape}

Figure \ref{fig:bootstrap_coverage} displays the results of this experiment. Specifically, it plots the 90 percent bootstrap intervals for the $\mathbf{ATE}_{A \rightarrow Y}$, computed across each of the 100 simulated datasets. In this figure, solid lines denote intervals that cover the true value of the target parameter, whereas dashed lines indicate intervals that fail to cover it. The figure shows that the 90 percent bootstrap intervals cover the true value of $\mathbf{ATE}_{A \rightarrow Y}$ in 96 of 100 simulated datasets. This result suggests that the bootstrap intervals achieve their nominal coverage rate but may be slightly conservative. The total computation time for this experiment was approximately 24 hours. Its replication files are available at \url{https://github.com/gtwodtke/deep_learning_with_DAGs/tree/main/MCEs}.
   
\subsection{Monte Carlo Experiments on Architecture and Hyper-parameters}\label{subsec:appendix_hyper}

In addition, we conducted another set of Monte Carlo experiments to investigate how the performance of cGNFs is influenced by variations in their architecture and hyper-parameter settings. In these experiments, we trained cGNFs with different architectures and hyper-parameters on 400 simulated datasets. In the first set of experiments, each dataset, consisting of 32,000 cases, was simulated from the normal, linear, and additive data-generating model described in Part \ref{subsec:appendix_bias_variance} of the Appendix. In the second set of experiments, each dataset was simulated from the data-generating model with both discrete and continuous variables, nonlinear relationships, effect heterogeneity, heteroscedasticity, and non-normal disturbances, also as outlined in Part \ref{subsec:appendix_bias_variance} of the Appendix.

\begin{figure}[ht!]
\begin{centering}
    \begin{subfigure}[b]{1\linewidth} 
        \includegraphics[width=\linewidth,height=0.6\textheight,keepaspectratio]{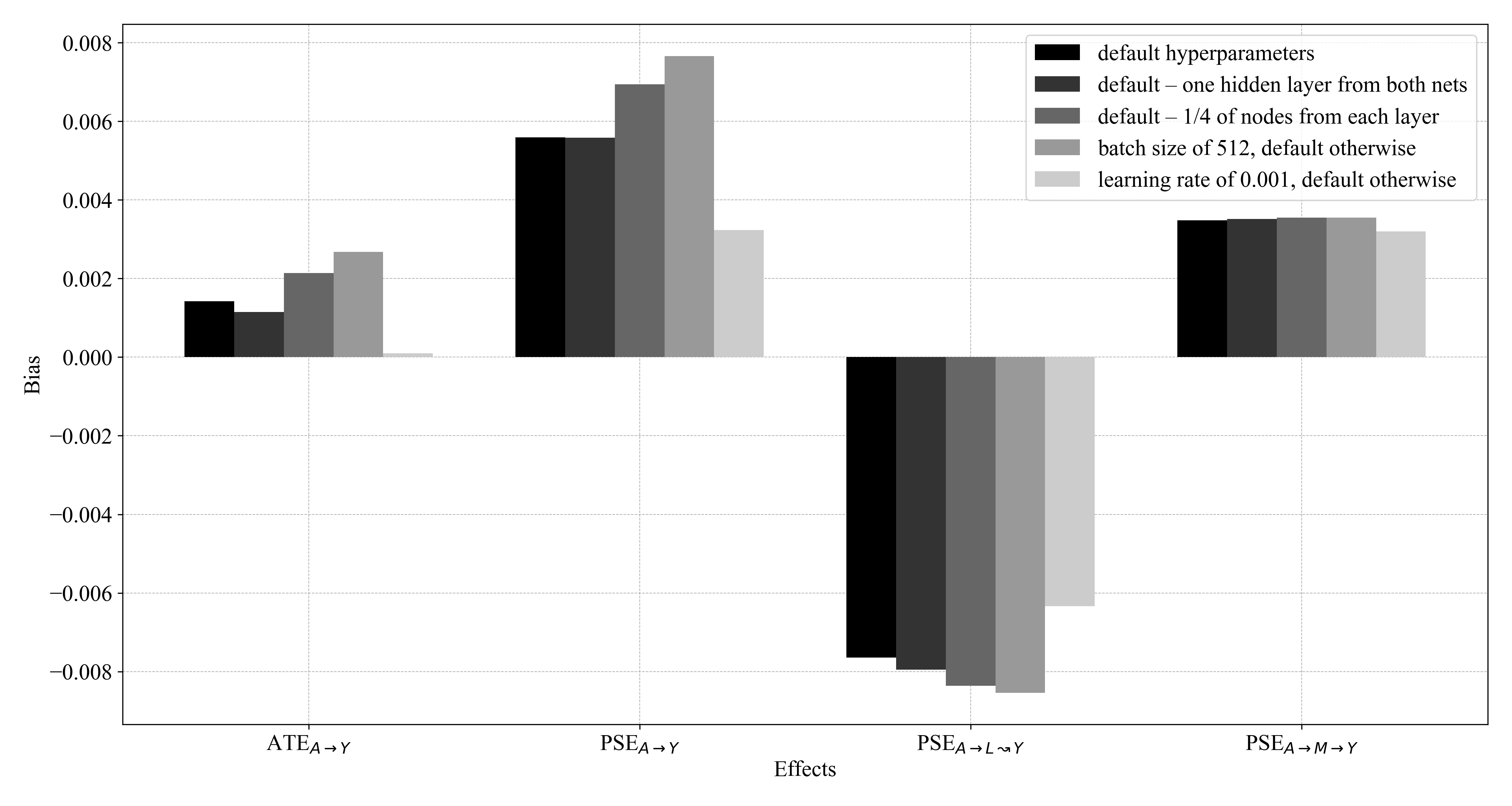}
        \caption{Bias of cGNF Effect Estimates}
        \label{fig:hyper_bias}
    \end{subfigure}
    \\[1ex] 
    \begin{subfigure}[b]{1\linewidth} 
        \includegraphics[width=\linewidth,height=0.6\textheight,keepaspectratio]{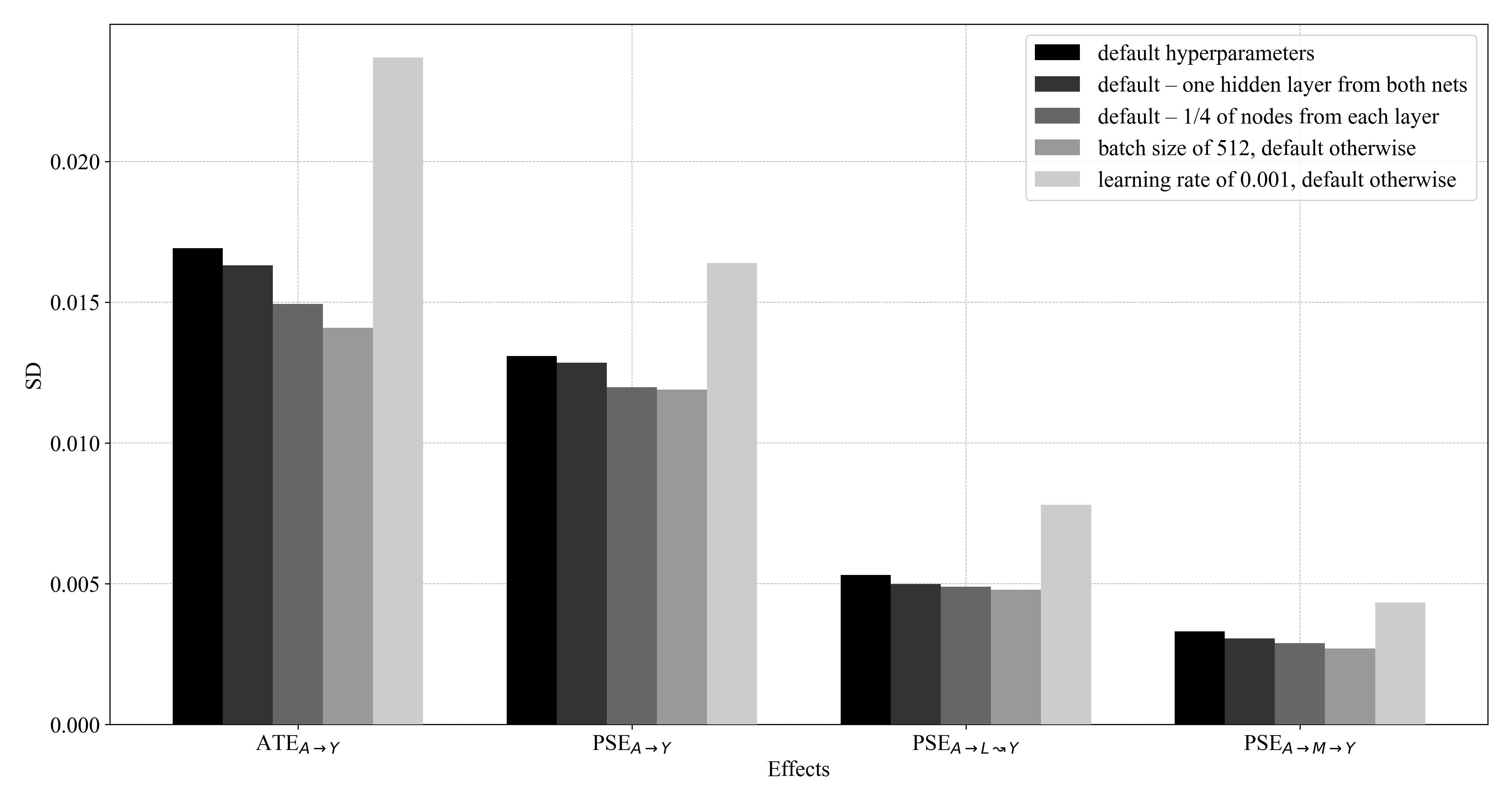}
        \caption{Standard deviation of cGNF Effect Estimates}
        \label{fig:hyper_sd}
    \end{subfigure}
    \caption{Performance of cGNFs across Different Architectures and Hyper-parameter Settings with a Normal, Linear, and Additive Data Generating Process.}
    \label{fig:combined_hyper_figure}
\medskip{}
\par\end{centering}
Note: Each set of results is based on 400 simulated datasets, each with a sample size of 32,000 generated from the linear and additive SEM with normal disturbances. The default hyper-parameters refer to a learning rate of 0.0001, a batch size of 128, an embedding network with five hidden layers, each composed of 100, 90, 80, 70, and 60 nodes, and an integrand network also with five hidden layers, composed of 60, 50, 40, 30, and 20 nodes each. 
\end{figure}

\begin{figure}[ht!]
\begin{centering}
    \begin{subfigure}[b]{1\linewidth} 
        \includegraphics[width=\linewidth,height=0.6\textheight,keepaspectratio]{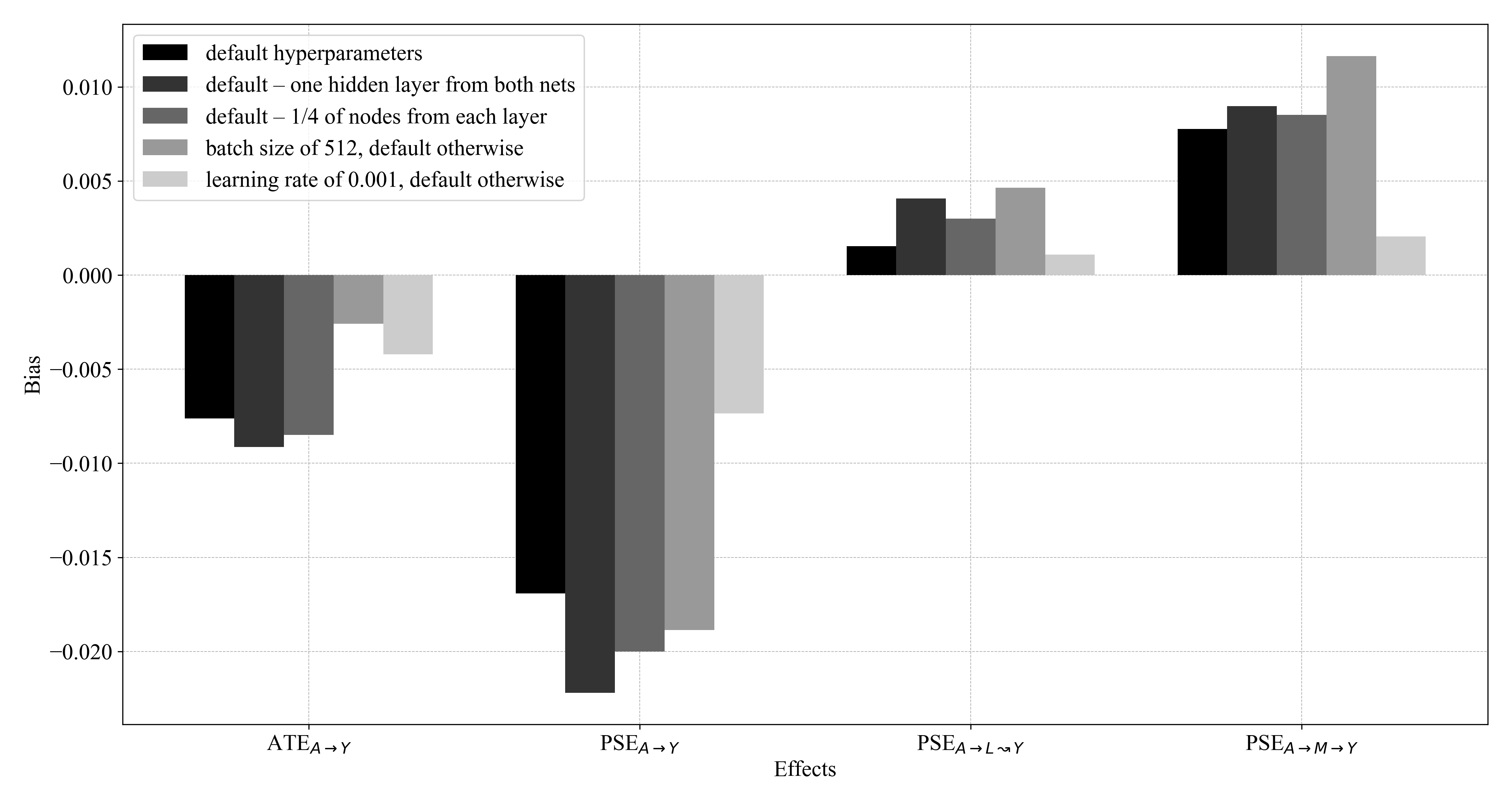}
        \caption{Bias of cGNF Effect Estimates}
        \label{fig:hyper_bias_nonlin}
    \end{subfigure}
    \\[1ex] 
    \begin{subfigure}[b]{1\linewidth} 
        \includegraphics[width=\linewidth,height=0.6\textheight,keepaspectratio]{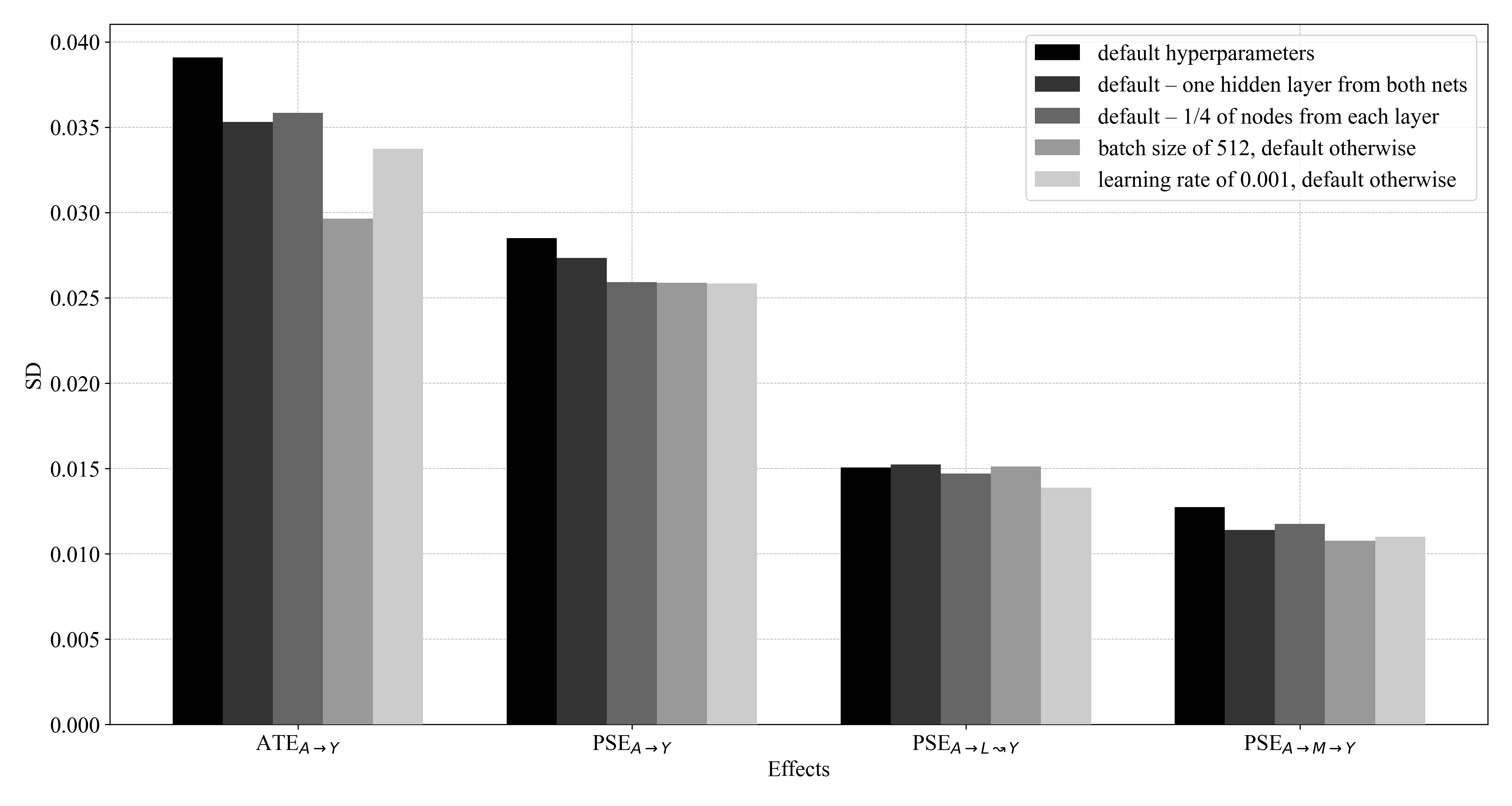}
        \caption{Standard deviation of cGNF Effect Estimates}
        \label{fig:hyper_sd_nonlin}
    \end{subfigure}
    \caption{Performance of cGNFs across Different Architectures and Hyper-parameter Settings with a Non-normal, Non-linear and Non-additive Data Generating Process.}
    \label{fig:combined_hyper_nonlin_figure}
\medskip{}
\par\end{centering}
Note: Each set of results is based on 400 simulated datasets, each with a sample size of 32,000 generated from the nonlinear and non-additive SEM. The default hyper-parameters refer to a learning rate of 0.0001, a batch size of 128, an embedding network with five hidden layers, each composed of 100, 90, 80, 70, and 60 nodes, and an integrand network also with five hidden layers, composed of 60, 50, 40, 30, and 20 nodes each. 
\end{figure}

The results from these two experiments are presented in Figures  \ref{fig:combined_hyper_figure} and \ref{fig:combined_hyper_nonlin_figure}, respectively. Both figures contain bar charts summarizing the bias and variance for selected cGNF estimates. Each bar corresponds to a cGNF trained with a specific architecture and hyper-parameter configuration. The ``default" bar represents a cGNF with the architecture, learning rate, and batch size outlined previously for the other simulation experiments. The bar labeled ``default $-$ one hidden layer" represents a cGNF with the same setup, minus the final hidden layer in both the embedding and integrand networks. Similarly, the bar labeled ``default $-$ 1/4 of nodes" refers to our default cGNF configuration with a $25\%$ reduction in the number of nodes in each hidden layer. The bar labeled ``batch size of 512" represents a cGNF with the default architecture and learning rate but an increased batch size of 512. Finally, the bar labeled ``learning rate of 0.001" refers to a cGNF with the default architecture and batch size, but a higher learning rate of 0.001. 

Overall, Figures \ref{fig:combined_hyper_figure} and \ref{fig:combined_hyper_nonlin_figure} suggest that cGNF estimates are fairly robust to modest variations in model architecture and hyper-parameter settings. Across the configurations tested, we observed levels of bias and variance that were low and stable for a variety of estimands. The only exception involved the variance of cGNF estimates when the learning rate was increased from 0.0001 to 0.001 in \ref{fig:combined_hyper_figure}, which led to a nontrivial increase in variance. Aside from this, the results suggest that cGNFs generally yield estimates with low bias and variance for a broad spectrum of sensible architectures and hyper-parameters, provided the sample size is sufficiently large. The wall time to complete each of these experiments was roughly 20 hours, when parallelized as above. Replication files are available at \url{https://github.com/gtwodtke/deep_learning_with_DAGs/tree/main/MCEs}.

\newpage

\end{document}